\pdfoutput=1
\PassOptionsToPackage{hyphens}{url}

\documentclass[dvipsnames]{article}

\usepackage[final]{slai}
\usepackage[utf8]{inputenc} % allow utf-8 input
\usepackage[T1]{fontenc}    % use 8-bit T1 fonts
\usepackage{url}            % simple URL typesetting

\usepackage{booktabs}       % professional-quality tables
\usepackage{amsfonts}       % blackboard math symbols
\usepackage{nicefrac}       % compact symbols for 1/2, etc.
\usepackage{multirow}
\usepackage{enumitem}
\usepackage{amssymb}
\usepackage{makecell}
\usepackage{listings}
\usepackage{subcaption}
\usepackage[labelfont=bf,tableposition=top]{caption} % bold caption labels

\usepackage{tabularx,array}
\usepackage{microtype}      % microtypography
\usepackage{xcolor}         % colors
\definecolor{slai}{HTML}{B34A78}
\usepackage{geometry}
\usepackage{mathpazo}
\usepackage[xcharter,bigdelims,vvarbb,noamssymbols,nosymbolsc]{newtxmath}
\usepackage[scaled=1.1]{zlmtt}
\usepackage{ragged2e}
\usepackage{amsmath}
\usepackage{bm}
\usepackage{latexsym}
\usepackage{graphicx}
\usepackage{float}
\usepackage[most,breakable]{tcolorbox}
\usepackage{etoolbox}
\usepackage{fancyhdr}
\usepackage{pgffor}
\usepackage{tikz}
\usepackage{bbm}
\usepackage{xspace}
\usepackage[english]{babel} % English hyphenation and typographic rules
\usepackage{hyphenat}
\usepackage{placeins}
\usetikzlibrary{arrows.meta}

\lstdefinestyle{benchmarkexample}{
  basicstyle=\ttfamily\footnotesize,
  columns=fullflexible,
  keepspaces=true,
  breaklines=true,
  breakatwhitespace=true,
  breakindent=0pt,
  showstringspaces=false,
  frame=single,
  rulecolor=\color{black!25},
  backgroundcolor=\color{black!2},
  framesep=7pt,
  xleftmargin=8pt,
  xrightmargin=8pt,
  linewidth=\linewidth,
  aboveskip=8pt,
  belowskip=12pt,
  numbers=none
}

\newcommand{\Rmnum}[1]{\expandafter\@slowromancap\romannumeral #1@}

\usepackage{chngcntr}
\usepackage[colorlinks, linkcolor=RoyalBlue, anchorcolor=BrickRed, citecolor=RoyalBlue, urlcolor=RoyalBlue]{hyperref}

\usepackage{cleveref}
\crefname{section}{\S}{\S\S}
\Crefname{section}{\S}{\S\S}

\newcommand{\model}{SLAI T-Rex-V2\xspace}

\newcommand\refsec[1]{Section~\hyperref[sec:#1]{\ref{sec:#1}}}
\newcommand\refsecs[2]{\hyperref[sec:#1]{\S\ref{sec:#1}:~\textsc{#1}}, \hyperref[sec:#2]{\S\ref{sec:#2}:~\textsc{#2}}}

\definecolor{academicblue}{RGB}{54, 95, 145}

\renewcommand{\headrulewidth}{0.5pt}
\renewcommand{\headrule}{\hbox to\headwidth{\color{black!35}\leaders\hrule height \headrulewidth\hfill}}
\usepackage{xparse}
\usepackage{colortbl}
\usepackage{needspace}
\newcolumntype{L}[1]{>{\raggedright\arraybackslash}p{#1}}
\newcolumntype{C}[1]{>{\centering\arraybackslash}p{#1}}
\newcolumntype{Y}{>{\raggedright\arraybackslash}X}

\pretocmd{\section}{\FloatBarrier}{}{}

\renewenvironment{abstract}{%
  \begin{tcolorbox}[
    enhanced,
    colback=slai!2!gray!2.5,
    colframe=slai!22!black!16,
    boxrule=0.55pt,
    arc=2.2mm,
    left=13pt,
    right=13pt,
    top=10.5pt,
    bottom=10.5pt
  ]
  \noindent{\normalsize\bfseries\color{slai!85!black} Abstract.}\hspace{0.55em}\fontsize{9.0}{11.6}\selectfont\ignorespaces
}{%
  \end{tcolorbox}
}

\title{%
\fontsize{13.2}{18.0}\selectfont
\model: Full-Stack Post-Training, RL Infrastructure,\\
and Agentic Modeling for Operations Research on Ascend SuperPOD%
}
\author{%
  \textbf{AI Training Platform Team, Shenzhen Loop Area Institute}%
}

\makeatletter
\newcommand\blfootnote[1]{%
  \begingroup
  \def\@makefntext##1{\noindent\fontsize{8.5}{11.2}\selectfont\color{black!70}##1}%
  \xdef\@thefnmark{}%
  \H@@footnotetext{#1}%
  \endgroup
}
\makeatother

\begin{document}
\maketitle
\blfootnote{For any questions or feedback regarding this technical report, please contact Dongfang Li (\mbox{\href{mailto:lidongfang@hit.edu.cn}{lidongfang@hit.edu.cn}}) and Baotian Hu (\mbox{\href{mailto:baotianhu@slai.edu.cn}{baotianhu@slai.edu.cn}}).}

\begin{abstract}
Solving practical operations research (OR) problems requires translating ambiguous natural-language descriptions into exact mathematical formulations and executable solver code. We present \model, an OR modeling system built on DeepSeek-V4-Flash and Huawei Ascend SuperPOD. Guided by failure modes surfaced through repeated base-model sampling, we construct a solver-verified data synthesis pipeline and post-train \model via cold-start SFT and multi-stage reinforcement learning. On our expert-curated \textsc{SLAI-OR-Bench} and \textsc{SLAI-OR-Bench-Hard} benchmarks, single-turn accuracy improves from 38.26\% to 50.86\% ($+5.82$ and $+6.59$ points over GPT-6.1-Sol and Claude Opus~5.5) and from 26.88\% to 35.89\%, slightly surpassing single-draw frontier baselines. Beyond single-turn generation, pairing the base model with a lightweight solver-in-the-loop agent harness lifts \textsc{SLAI-OR-Bench-Hard} accuracy to 40.86\%, reaching 49.46\% with reusable general OR skills and 55.48\% with benchmark-tailored priors. To sustain efficient and stable MoE reinforcement learning on Ascend, we further prevent training--inference numerical drift from triggering RL collapse, cut rollout token generation by 82.5\% through calibrated trajectory reuse, accelerate low-concurrency OR serving with native multi-token prediction, and scale agentic RL for NPU kernel optimization.
\end{abstract}

\vspace{6.5mm}
\begin{figure}[H]
  \centering
  \captionsetup{skip=7.5pt}
  \resizebox{\linewidth}{!}{% Two-layer overview for the SLAI T-Rex Part II report.
% Figure fragment: no document class, preamble, caption, or figure environment.
% Required in the host preamble:
%   \usepackage{tikz}
%   \usetikzlibrary{arrows.meta}
% Native size: approximately 152 mm x 83 mm.
% The host report's slai.sty selects Helvetica for \sffamily.
% All colors and drawing styles are scoped to this figure.
\begingroup
\definecolor{ovBerry}{HTML}{B34A78}
\definecolor{ovTeal}{HTML}{008B6A}
\definecolor{ovInk}{HTML}{132B53}
\definecolor{ovMuted}{HTML}{60748C}
\definecolor{ovPale}{HTML}{EAF6FA}
\definecolor{ovLine}{HTML}{C5E0E8}
\begin{tikzpicture}[
  x=1mm,y=1mm,
  line cap=round,line join=round,
  font=\sffamily\fontsize{8}{9.6}\selectfont,
  ovtitle/.style={text=ovInk,font=\sffamily\bfseries\fontsize{8.6}{10.2}\selectfont},
  ovnote/.style={text=ovMuted,font=\sffamily\fontsize{7}{8.4}\selectfont},
  ovstage/.style={
    draw=ovTeal!75,fill=white,line width=0.55pt,
    rounded corners=1.5pt,minimum height=9mm,text width=14.2mm,
    align=center,inner sep=0pt,outer sep=0pt,
    text=ovInk,font=\sffamily\bfseries\fontsize{7.1}{8.6}\selectfont
  },
  ovaction/.style={ovstage,minimum height=7.8mm,text width=17.5mm},
  ovflow/.style={draw=ovBerry,line width=0.8pt,-{Stealth[length=1.55mm,width=1.1mm]}},
  ovfeedback/.style={draw=ovTeal,line width=0.65pt,-{Stealth[length=1.35mm,width=1mm]}},
  ovsupport/.style={draw=ovMuted!75,line width=0.75pt,dash pattern=on 2.4pt off 2pt,
    -{Stealth[length=1.7mm,width=1.15mm]}},
  ovinfrahead/.style={text=ovInk,align=center,inner sep=0pt,
    font=\sffamily\bfseries\fontsize{7.5}{8.8}\selectfont},
  ovinfranote/.style={text=ovMuted,align=center,inner sep=0pt,
    font=\sffamily\fontsize{7}{8.5}\selectfont}
]
\path[use as bounding box] (-0.3,-0.3) rectangle (152.3,83.3);

% Layer 1: two separately evaluated routes, with no sequential link between them.
\path[draw=ovLine,fill=ovPale!32,line width=0.7pt,rounded corners=2.5mm]
  (0,39) rectangle (152,83);
\path[fill=ovBerry,rounded corners=2.5mm] (0,74) rectangle (152,83);
\path[fill=ovBerry] (0,74) rectangle (152,78);
\node[text=white,font=\sffamily\bfseries\fontsize{10}{12}\selectfont]
  at (76,78.5) {OR Learning and Agentic Execution};

% Left panel: solver-grounded parameter learning.
\path[draw=ovLine,fill=white,line width=0.55pt,rounded corners=2mm]
  (3,42) rectangle (75,71.5);
\node[ovtitle] at (39,67.2) {Solver-Grounded Post-Training};
\node[ovnote] at (39,62.5) {Diagnosis-guided, solver-verified supervision};
\node[ovstage] (ovdata) at (13.8,53.4) {Verified\\OR Data};
\node[ovstage] (ovsft) at (30.5,53.4) {Cold-Start\\SFT};
\node[ovstage] (ovchat) at (47.2,53.4) {Chat\\GRPO};
\node[ovstage] (ovthink) at (63.9,53.4) {Thinking\\GRPO};
\draw[ovflow] (ovdata.east) -- (ovsft.west);
\draw[ovflow] (ovsft.east) -- (ovchat.west);
\draw[ovflow] (ovchat.east) -- (ovthink.west);
\node[ovnote,font=\sffamily\fontsize{6.8}{8.2}\selectfont]
  at (39,45.4) {Initialization: DeepSeek-V4-Flash-0731 (Base)};

% Right panel: a compact runtime with event-driven OR skills.
\path[draw=ovLine,fill=white,line width=0.55pt,rounded corners=2mm]
  (77,42) rectangle (149,71.5);
\node[ovtitle] at (113,67.2) {Agentic OR Execution};
\node[ovnote] at (113,62.5) {Lightweight harness + modular OR skills};
\node[ovaction] (ovform) at (91,55.0) {Formulate};
\node[ovaction] (ovexec) at (113,55.0) {Execute};
\node[ovaction] (ovverify) at (135,55.0) {Verify};
\draw[ovflow] (ovform.east) -- (ovexec.west);
\draw[ovflow] (ovexec.east) -- (ovverify.west);
\draw[ovfeedback] (ovverify.south) -- (135,48.6) -- (91,48.6) -- (ovform.south);
\node[text=ovTeal,fill=white,inner xsep=1mm,inner ysep=0.25mm,
  font=\sffamily\fontsize{6.7}{8.1}\selectfont]
  at (113,48.6) {Feedback and revision};
\node[ovnote,font=\sffamily\fontsize{6.8}{8.2}\selectfont]
  at (113,45.2) {Base model with tools and runtime verification};

% The dashed relation denotes systems support and optimization studies.
\draw[ovsupport] (39,32.4) -- (39,38.4);
\draw[ovsupport] (113,32.4) -- (113,38.4);
\node[ovnote] at (76,35.5) {Systems support};

% Layer 2: the four infrastructure directions.
\path[draw=ovTeal!65,fill=ovPale!58,line width=0.7pt,rounded corners=2.5mm]
  (0,0) rectangle (152,32);
\node[anchor=west,text=ovTeal,font=\sffamily\bfseries\fontsize{8.3}{10}\selectfont]
  at (4,27.2) {INFRASTRUCTURE FOR EXECUTION-IN-THE-LOOP RL};
\foreach \x in {3,40,77,114}{
  \path[draw=ovLine,fill=white,line width=0.55pt,rounded corners=1.6mm]
    (\x,8) rectangle (\x+35,22.5);
}
\node[ovinfrahead] at (20.5,18.9) {Trajectory Reuse};
\node[ovinfranote] at (20.5,12.3) {Advantage recalibration\\and verified-sample reuse};

\node[ovinfrahead] at (57.5,18.5) {Training--Inference\\Consistency};
\node[ovinfranote] at (57.5,11.8) {Numerical diagnosis\\and alignment};

\node[ovinfrahead] at (94.5,18.9) {Native MTP};
\node[ovinfranote] at (94.5,12.3) {Draft prediction\\and speculative decoding};

\node[ovinfrahead] at (131.5,18.5) {Agentic Kernel\\Optimization};
\node[ovinfranote] at (131.5,11.8) {Intermediate feedback\\and asynchronous updates};

\draw[ovLine,line width=0.5pt] (4,6) -- (148,6);
\node[text=ovInk,font=\sffamily\fontsize{7}{8.4}\selectfont]
  at (76,3.1) {\textbf{Huawei Ascend SuperPOD}\quad\textbullet\quad Training, inference, and execution infrastructure};
\end{tikzpicture}
\endgroup}
  \caption{\linespread{1.14}\selectfont Two-layer overview of \model: the upper layer covers solver-grounded post-training and base-model agentic execution with OR skills; the lower layer summarizes four infrastructure directions on Huawei Ascend SuperPOD.}
  \label{fig:slai-overview}
\end{figure}
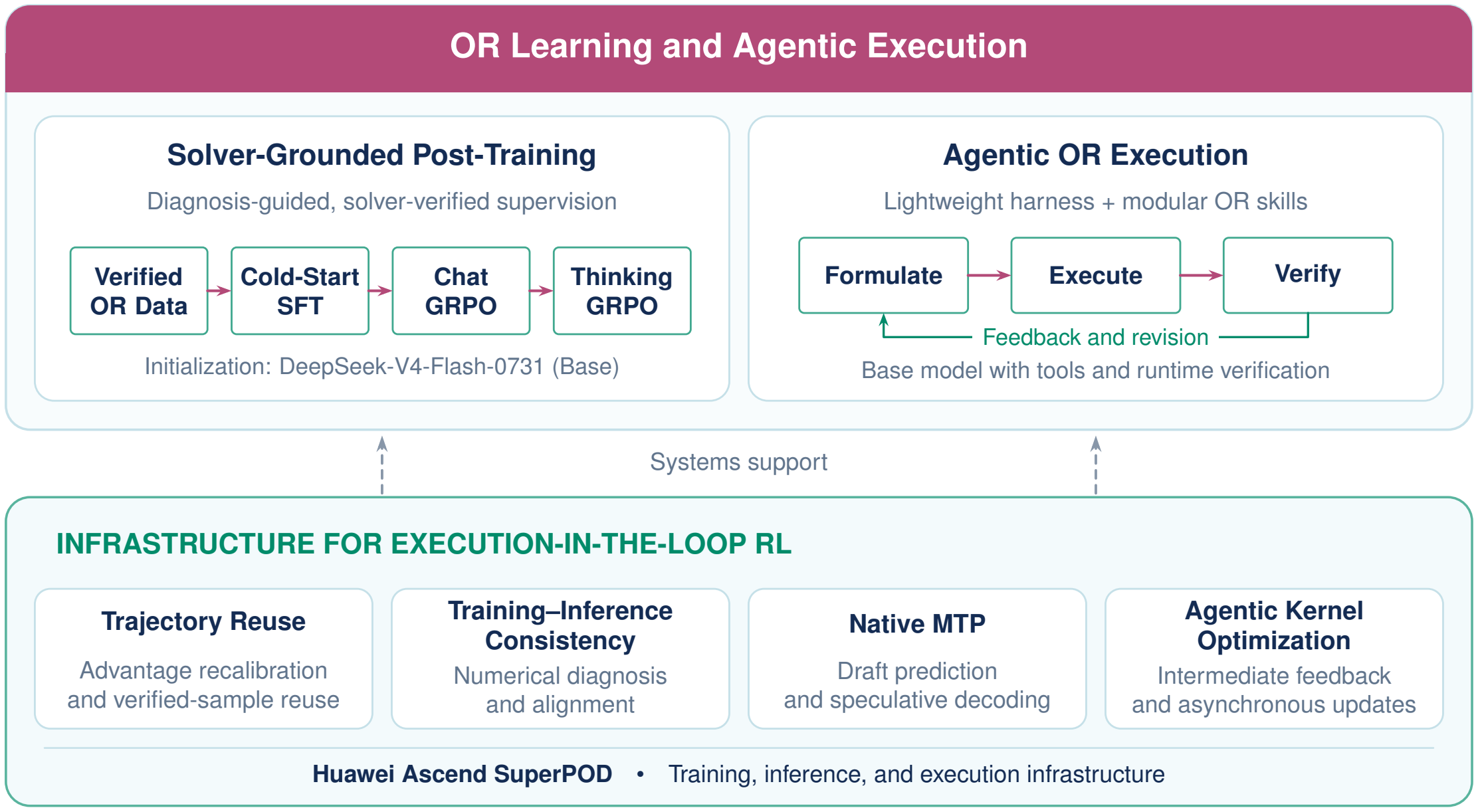

\newpage
{
  \thispagestyle{plain}
  \setcounter{tocdepth}{2}
  \setlength{\parskip}{1.2pt}
  \hypersetup{linkcolor=RoyalBlue, linktoc=page}
  \tableofcontents
}
\newpage

\section{Introduction}
\label{sec:introduction}

Supervised fine-tuning (SFT) and reinforcement learning (RL) with verifiable rewards have substantially improved the mathematical reasoning and coding abilities of large language models~\citep{shao2024deepseekmath,deepseekr1,dapo}. Operations research (OR), however, demands more than isolated mathematical derivation or code synthesis~\citep{optimus,wang2025orgeval}. Given a problem description in business or engineering language, a model must identify the decision variables, formulate the objective and operational constraints as a linear or mixed-integer linear program (LP/MILP), implement executable solver code (e.g., in GurobiPy or Pyomo), and recover the optimal objective value. A program that compiles and returns an ``optimal'' solver status can still be wrong if it omits a coupling constraint, mishandles units, or confuses individual demand paths with shared resource usage.

In Part~I of the SLAI T-Rex technical report~\citep{slai2026trexpart1}, we established a full-parameter post-training stack for the DeepSeek-V4 Mixture-of-Experts (MoE) family~\citep{deepseekv3} on Huawei CloudMatrix384 SuperPOD~\citep{zuo2025cloudmatrix384}, covering hybrid parallelism, memory management, AscendC kernel execution, and an initial solver-grounded continued pre-training (CPT) and SFT pipeline. In this Part~II report, we present \model, a full-stack OR reasoning and agentic platform centered on DeepSeek-V4-Flash. Rather than treating post-training, RL systems, and inference-time agency in isolation, we structure \model around two complementary capability routes---(i) \emph{solver-grounded parameter specialization} for single-turn OR modeling and (ii) \emph{verified test-time agentic execution} with modular OR skills---supported by four controlled systems and algorithmic studies addressing key bottlenecks of execution-in-the-loop RL on Ascend NPUs (Figure~\ref{fig:slai-overview}). Each component is evaluated against matched baselines isolating its specific mechanism, establishing both strong standalone performance and a clear path toward joint agentic trajectory internalization.

\paragraph{Capability Diagnosis, Data Synthesis, and OR Post-Training.}
We organize single-turn OR post-training around a closed loop from capability diagnosis to verified synthesis, sequential training, and paired transition analysis (Sections~\ref{sec:evaluation} and~\ref{sec:or-bench-post-training}). To evaluate models beyond largely saturated public datasets, we first curate \textsc{SLAI-OR-Bench} (131 tasks across 18 domains and 26 decision-task families) and its frontier-challenging subset \textsc{SLAI-OR-Bench-Hard} (93 tasks). Eight-sample evaluation of the base model (DeepSeek-V4-Flash-0731) on \textsc{SLAI-OR-Bench-Hard} distinguishes \emph{Stable}, \emph{Partial}, and \emph{Unsolved} problems: 53.76\% remain unsolved within eight draws, while the 19.36-point gap between pass@1 (26.88\%) and pass@8 (46.24\%) exposes substantial execution and formulation unreliability. We diagnose these gaps through a four-layer synthesis taxonomy---scenario, modeling technique, constraint families, and modeling pitfalls---and drive one-hop structural data synthesis verified via dual-interface (\texttt{Pyomo} and \texttt{gurobipy}) solver execution. Empirical base-model solve rates then route verified demonstrations and prompts across cold-start SFT, Chat GRPO, and Thinking GRPO, coupled with graded failure rewards and resource-limit outcome masking. On \textsc{SLAI-OR-Bench} and \textsc{SLAI-OR-Bench-Hard}, the resulting post-trained model raises single-turn accuracy from 38.26\% to \textbf{50.86\%} ($+5.82$ percentage points over GPT-6.1-Sol and $+6.59$ points over Claude Opus~5.5) and from 26.88\% to \textbf{35.89\%} (edging out GPT-6.1-Sol at 34.41\% and Claude Opus~5.5 at 33.33\%), while expanding Hard-set pass@8 coverage from 46.24\% to \textbf{51.61\%}.

\paragraph{Infrastructure for Execution-in-the-Loop RL on Ascend.}
Scaling solver-verified and profiler-verified RL for large MoE models on Ascend SuperPOD faces four recurring systems and optimization bottlenecks: expensive rollout generation and solver verification, numerical discrepancies between decoupled inference and training engines~\citep{precisionrl,zhong2026diagnosing,bai2026bitwise}, autoregressive decoding latency, and multi-hour NPU kernel profiling tails~\citep{wu2026ascendoptimizer,dai2026cudaagent}. Section~\ref{sec:or-rl-infrastructure} investigates targeted solutions to each bottleneck under controlled experimental protocols:
\begin{enumerate}[leftmargin=2em, itemsep=2pt, topsep=3pt]
  \item \textbf{Advantage Recalibration for Trajectory Reuse (Section~\ref{sec:v4-reuse}):} Because each OR rollout requires long reasoning traces and external solver execution, discarding verified trajectories after a single gradient step underutilizes supervision. Within a bounded same-batch loop ($R_{\max}=4$), we re-estimate each prompt's current-policy reference reward using tempered, length-normalized log-likelihood shifts on the learner backend, paired with a matching weighted actor objective and group eligibility gates. At a matched budget of 80 optimizer updates on 128 Ascend NPUs, calibrated reuse reduces newly generated trajectories by \textbf{75.0\%} and response tokens by \textbf{82.5\%} while matching or slightly exceeding fresh-rollout accuracy on \textsc{SLAI-OR-Bench-Hard}.
  \item \textbf{Training--Inference Consistency Optimization (Section~\ref{sec:train-infer-consistency}):} Numerical discrepancies between rollout engines (vLLM-Ascend) and training backends (MindSpeed-LLM/Megatron) bias on-policy updates and can flip discrete MoE routing. We develop a progressive diagnostic workflow, packaged as a reusable agent skill, that combines layer-wise token cosine scanning with recursive same-input replay to separate operator implementation bugs from downstream mathematical amplification. Across eight nodes (64 Ascend 910C NPUs) evaluated on 16 diagnostic prompt--response sequences, the workflow pinpoints and repairs a SwiGLU clipping discrepancy and a router correction-bias precision mismatch on DeepSeek-V4-Flash, preventing numerical-drift-induced RL collapse.
  \item \textbf{Speculative Decoding with Native MTP (Section~\ref{sdinsert:sec:speculative-decoding}):} Evaluating native multi-token prediction (MTP) across workloads and decoding regimes reveals pronounced variation in draft acceptance, output-length sensitivity, and serving speedup. After joint RL--MTP training exposes trade-offs between OR accuracy and draft acceptance, refining the private MTP module with the selected OR target frozen raises mean three-token agreement on the OR development panel from \textbf{48.44\% to 60.16\%}. In matched offline serving benchmarks, enabling saved native MTP modules delivers \textbf{$1.583\times$--$1.715\times$} OR token throughput relative to standard decoding at client concurrency one, while paired solver evaluation audits solution quality independently of raw token rate.
  \item \textbf{Decoupled Exploration for Long-Horizon Kernel Optimization (Section~\ref{sec:kernel-perf-opt}):} Iterative Ascend NPU kernel optimization suffers from heavy exploration costs and multi-hour episode tails caused by compilation, precision verification, and device profiling. Alongside benchmark diagnosis on \textsc{AscendGYM} and solver-guided tiling/buffering warm starts (\texttt{WindMatmulV2}), we convert each measured candidate along an \texttt{ops-math} optimization episode into a history-preserving training prefix fed into an asynchronous FIFO queue, augmented with an evidence-grounded shared lesson memory. Over 24 hours on Ascend, this increases completed parameter updates from \textbf{4 to 12 ($3\times$)}, raises candidate correctness pass rates from \textbf{39.4\% to 49.1\%}, and substantially accelerates the discovery of verified trajectories achieving $\geq 1.05\times$ and $\geq 1.10\times$ speedups.
\end{enumerate}

\paragraph{Lightweight Agent Harness and Modular OR Skills.}
While post-training internalizes single-turn formulation ability into model weights, complex OR tasks also benefit at test time from multi-turn data inspection, iterative execution, and semantic self-verification. General-purpose coding agents, however, do not uniformly improve OR performance: on \textsc{SLAI-OR-Bench-Hard}, Claude Code~\citep{anthropic2026claudecode} lifts Claude Opus~5.5 from 33.33\% to 38.71\%, whereas Codex~\citep{openai_codex_cli} slightly reduces GPT-6.1-Sol from 34.41\% to 33.33\%. In Section~\ref{sec:or-harness-skills}, we pair the base model with a minimalist PI-style ReAct~\citep{yao2023react} harness and eight modular OR skills (\texttt{scene\_cards}, \texttt{technique\_cards}, \texttt{numeric\_probes}, \texttt{field\_roles}, \texttt{code\_checks}, \texttt{modeling\_spec}, \texttt{gurobi\_landing}, and \texttt{referee}). Each skill couples declarative modeling guidance (\texttt{SKILL.md}) with deterministic Python runtime hooks that enforce formulation specifications before coding, probe numerical plausibility after solving, and audit constraint fidelity before submission. On \textsc{SLAI-OR-Bench-Hard}, the lightweight harness alone lifts base-model accuracy from \textbf{26.88\%} to \textbf{40.86\%} ($+13.98$ points); enabling the six benchmark-independent workflow skills raises accuracy to \textbf{49.46\%} ($+22.58$ points, surpassing external coding-agent baselines and single-turn base pass@8 without oracle selection), and adding the two benchmark-tailored prior cards reaches \textbf{55.48\%} ($+28.60$ points) as an attainable upper-bound probe.

\paragraph{Report Organization.}
Section~\ref{sec:evaluation} reviews public AI4OR benchmarks and introduces \textsc{SLAI-OR-Bench} and our evaluation protocol. Section~\ref{sec:or-bench-post-training} presents base-model reliability diagnosis, label-guided data synthesis, cold-start SFT and sequential Chat/Thinking GRPO, and paired scenario-level transition analysis. Section~\ref{sec:or-rl-infrastructure} details our four Ascend RL infrastructure studies across trajectory reuse (Section~\ref{sec:v4-reuse}), training--inference consistency (Section~\ref{sec:train-infer-consistency}), native MTP speculative decoding (Section~\ref{sdinsert:sec:speculative-decoding}), and NPU kernel optimization (Section~\ref{sec:kernel-perf-opt}). Section~\ref{sec:or-harness-skills} describes the lightweight OR harness, modular skill ablations, and our roadmap for internalizing skill-assisted trajectories into model parameters. Section~\ref{sec:conclusion} concludes the report, while Appendices~\ref{app:author-list}--\ref{app:sdv2} list the project contributors and provide dataset provenance, a worked \textsc{SLAI-OR-Bench-Hard} case study, detailed analyses of exploratory zero-advantage GRPO variants, and OR--MTP co-adaptation controls.

\section{SLAI-OR-Bench: Benchmark Curation and Diagnostic Protocol}

\label{sec:evaluation}

Evaluating AI for operations research (AI4OR) requires benchmarks
that test whether a model can translate a problem description
into a mathematical formulation and executable solver code.
This section reviews existing benchmarks, examines their
discriminative power for recent models, and introduces
\textsc{SLAI-OR-Bench}, an expert-curated collection designed to
complement existing coverage with application-specific
optimization tasks.

\subsection{Saturation Analysis of Public AI4OR Benchmarks}
\label{sec:evaluation-benchmarks}
Existing benchmarks evaluate optimization modeling through
natural-language problem descriptions, numerical instance data,
and reference formulations or solutions.
Our diagnostic survey includes
NL4OPT~\citep{ramamonjison2022nl4opt},
OptiBench~\citep{yang2025optibench},
Bench4Opt~\citep{wang2025orgeval},
ComplexOR~\citep{xiao2024chainofexperts},
IndustryOR~\citep{huang2025orlm},
MAMO-ComplexLP and MAMO-EasyLP~\citep{huang2025mathematical}, and NLP4LP~\citep{optimus}. We additionally include problems from the ICML 2024 Automated
Optimization competition, described by
Nguyen and Ko~\citep{nguyen2024icmlreport}.

These benchmarks provide complementary views of modeling
performance. However, high accuracy on a benchmark can reduce
its ability to distinguish stronger models, while a benchmark
that remains difficult may still cover only a limited selection
of application-specific requirements. Evaluation therefore
requires both sufficient difficulty and broad coverage of domain-specific
modeling structures.

\paragraph{Diagnostic performance of frontier models.}
Table~\ref{tab:frontier-benchmark-diagnostic} reports solver-based
accuracy for three external frontier models---GPT-6.1-Sol, Claude Opus~5.5, and Gemini-4-Argon---across nine public benchmarks to assess current ceiling effects. Because several original benchmark releases contain logically inconsistent or underspecified statements and erroneous ground-truth objectives, we adopt community-cleaned releases to minimize annotation noise; dataset versions and instance counts are documented in
Appendix~\ref{app:existing-benchmarks}.

\begin{table}[!htbp]
\centering
\small
\renewcommand{\arraystretch}{1.12}
\caption{Solver-based accuracy (\%) of external frontier models on cleaned public benchmark releases as a saturation diagnostic.
NL4OPT results use the same 202 instances for all three models under each model's recorded single-turn evaluation configuration.}
\label{tab:frontier-benchmark-diagnostic}
\begin{tabular*}{\linewidth}{@{\extracolsep{\fill}}lrccc@{}}
\toprule
\textbf{Benchmark} & $\bm{N}$
& \textbf{GPT-6.1-Sol}
& \textbf{Claude Opus 5.5}
& \textbf{Gemini-4-Argon} \\
\midrule
NL4OPT             & 202 & 83.66 & 88.61 & 83.17 \\
OptiBench          & 403 & 85.36 & 91.07 & 87.10 \\
ComplexOR          & 18  & 72.22 & 72.22 & 66.67 \\
ICML Competition   & 410 & 78.05 & 83.66 & 80.00 \\
IndustryOR         & 42  & 61.90 & 71.43 & 69.05 \\
MAMO-ComplexLP     & 111 & 63.06 & 86.49 & 62.16 \\
MAMO-EasyLP        & 545 & 98.53 & 99.45 & 98.90 \\
NLP4LP             & 178 & 93.82 & 94.38 & 92.70 \\
Bench4Opt          & 344 & 59.88 & 60.76 & 61.05 \\
\bottomrule
\end{tabular*}
\end{table}

Several widely used datasets (e.g., MAMO-EasyLP, NLP4LP, OptiBench, and NL4OPT) exceed 83\%--99\% accuracy on frontier models, leaving limited headroom, while smaller unsaturated sets (such as ComplexOR with $N=18$ or IndustryOR with $N=42$) offer narrow structural diversity, and public collections frequently overlap with seed or regression pools in open-source AI4OR pipelines.
To enable contamination-controlled evaluation on complex operational formulations, we construct \textsc{SLAI-OR-Bench} as an expert-curated test
collection strictly held out from all training, synthesis, and
development data in this work.

\subsection{Benchmark Construction, Taxonomy, and Hard-Subset Design}
\label{sec:benchmark_dataset}

\paragraph{Construction and provenance.}
We engaged domain and operations research experts to develop a dedicated internal
test collection. \textsc{SLAI-OR-Bench} contains 131 optimization
problems: 106 problems curated and annotated by these experts
and 25 problems adapted from Bench4Opt~\citep{wang2025orgeval}
with newly randomized numerical parameters.

All instances, their variants, and reference solver code are
excluded from this project's SFT and RL training data,
validation sets, and prompt examples.
In response to the OR-Bench community call for
contributions~\citep{zhu2026orbenchcall}, we submitted the
collection to the organizing committee and subsequently
revised it in response to their review feedback.

\paragraph{Instance representation.}
Each instance includes a problem statement, a reference
mathematical formulation, executable reference solver code,
and a reference optimal objective value and solution.
The 106 expert-curated instances are self-contained, with
numerical parameters embedded in their statements.
The 25 Bench4Opt-derived instances include separate JSON
data files. For these instances, the evaluated LLM receives only the problem
statement and a description of the data schema, without access
to the numerical data. The LLM must generate solver code that
loads the associated JSON file at execution time.
Reference formulations, reference code, objective values,
and solutions are withheld from the evaluated LLM.

\paragraph{Problem-family coverage.}
The collection spans 18 application-domain categories,
including production planning, transportation, healthcare,
energy, workforce scheduling, telecommunications, and finance.
It contains 24 LP and 107 MILP instances. Beyond its 18 application-domain labels, we organize the
131 instances into 26 decision-task families according to
their principal decision variables, objectives, and operational
constraints. Each instance is assigned to one primary family,
while application-specific variants are documented separately.
This classification distinguishes decision-task families from
both application domains and mathematical model classes
(LP and MILP). The 93-instance hard subset retains instances
from all 26 families.

The collection includes shared-channel network routing,
arc-service routing with replenishment, multi-agent
collision-free path planning, batching and reclassification,
dynamic intervention, safety-stock optimization, dynamic
pricing, and equilibrium-constrained strategic decisions,
in addition to production planning, assignment, and scheduling.
Table~\ref{tab:slai-task-family-counts} reports the distribution.

\begin{table}[!htbp]
\centering
\small
\renewcommand{\arraystretch}{1.06}
\caption{Decision-task families in the 131-instance \textsc{SLAI-OR-Bench}
collection and its 93-instance hard diagnostic subset (\textsc{SLAI-OR-Bench-Hard}). Each instance is counted
once.}
\label{tab:slai-task-family-counts}
\begin{tabularx}{\linewidth}{@{}Yrr@{}}
\toprule
\textbf{Decision-task family} & \textbf{Full} & \textbf{Hard} \\
\midrule
Network routing and shared-channel allocation & 3 & 3 \\
Network design and resilience & 2 & 1 \\
Facility location and coverage & 3 & 1 \\
Facility layout and interaction-cost assignment & 2 & 2 \\
Equipment operation and energy dispatch & 11 & 10 \\
Budgeted selection and resource allocation & 8 & 2 \\
Dynamic intervention and state control & 3 & 1 \\
Task and multistage workflow scheduling & 14 & 10 \\
Project and maintenance scheduling & 4 & 3 \\
Service-facility assignment and scheduling & 14 & 10 \\
Workforce scheduling and crew itineraries & 8 & 4 \\
Arc-service routing and replenishment & 5 & 2 \\
Pickup-and-delivery routing & 1 & 1 \\
Multi-agent collision-free path planning & 1 & 1 \\
Train timetabling and track-conflict scheduling & 3 & 3 \\
Tournament and event timetabling & 4 & 4 \\
Batching, sortation, and reclassification & 3 & 2 \\
Blending and mixture optimization & 11 & 7 \\
Production, extraction, and phased allocation & 9 & 7 \\
Inventory, replenishment, and safety stock & 3 & 3 \\
Transportation and time-dependent network flow & 7 & 6 \\
Multidimensional packing and spatial allocation & 7 & 5 \\
Discrete allocation under probabilistic risk & 1 & 1 \\
Game equilibrium and strategic decisions & 1 & 1 \\
Cutting and material utilization & 2 & 2 \\
Dynamic pricing with perishable inventory & 1 & 1 \\
\midrule
\textbf{Total} & \textbf{131} & \textbf{93} \\
\bottomrule
\end{tabularx}
\end{table}

The application-specific variants introduce distinct modeling
requirements within these families. For example, laboratory
sample assignment requires determining whether mixed racks
trigger re-batching; emergency transfers require tracking when
incoming resources become available for onward dispatch;
and dynamic pricing couples current prices with subsequent
demand and perishable inventory. These requirements provide
a more detailed characterization of benchmark coverage than
domain names or LP/MILP labels alone. While Table~\ref{tab:slai-task-family-counts} characterizes benchmark diversity at fine-grained OR problem-class resolution (26 decision-task families), Section~\ref{sec:ds-tags} consolidates related decision structures into a coarser 21-scenario synthesis taxonomy (19 of which occur in the 93-instance Hard subset) paired with 31 explicit mathematical formulation techniques to drive combinatorial training-data generation. Correct formulation also requires commonsense and domain
knowledge to interpret requirements expressed in business
language. Relevant distinctions include which quantities
represent indivisible units, how initial resource states affect
the first planning period, and whether an operational metric
counts individual tasks or shared resources. Such distinctions
can change the objective or feasible set even when the underlying
problem family is correctly recognized.
Appendix~\ref{app:representative_examples} presents an original
problem description, its reference formulation, and actual model
outputs illustrating the interpretation of shared-resource costs.

\paragraph{Hard-subset construction.}
We construct a 93-instance diagnostic subset, denoted
\textsc{SLAI-OR-Bench-Hard}, using DeepSeek-V4-Flash-0731, Claude Opus~5~\citep{anthropic2026opus5}, and GPT-5.6-Sol (evaluated at the time of benchmark freeze, prior to our final evaluations against Claude Opus~5.5 and GPT-6.1-Sol).
An instance is retained if at least one of these screening models is
scored incorrect, removing the 38 instances solved by all three while retaining 93 instances across all 18 application domains and all 26 decision-task families. Because filtering conditions on this initial screening pass, all primary single-turn comparisons report both the unconditioned 131-instance full collection and the 93-instance hard subset.

\subsection{Solver-Verified Evaluation and Multi-Sample Metrics}
\label{sec:evaluation-protocol}

\paragraph{Solver-based accuracy.}
The model generates solver code from the problem statement
and any associated numerical data. We execute the code and
compare the resulting objective value with the reference value.
For the curated collection, a numerical match satisfies
\[
\left|\widehat{z}-z^\star\right|
\leq 10^{-4}+10^{-6}\left|z^\star\right|,
\]
where $\widehat{z}$ is the generated model's objective value
and $z^\star$ is the reference value.
Applied after solver execution (distinct from the numerical solver's internal optimality-gap settings), this end-to-end metric penalizes formulation mismatches, code errors, and unfinished solves alike while evaluating objective agreement on each numerical instance.

\paragraph{Repeated evaluation and multi-sample coverage.}
For Base and \model (SFT+RL), we perform eight independent evaluation
runs on each benchmark and report mean solve accuracy across runs in aggregate comparisons. On \textsc{SLAI-OR-Bench-Hard}, we further analyze the $n=8$ responses per problem: for a problem with $c$ correct responses, we estimate
\[
\widehat{\mathrm{pass@}k}
=
1-\frac{\binom{n-c}{k}}{\binom{n}{k}},
\qquad
k\in\{1,2,4,6,8\},
\]
where $\binom{n-c}{k}=0$ when $n-c<k$, and average over the 93 problems so that $\mathrm{pass@}1$ measures single-attempt accuracy and $\mathrm{pass@}8$ measures eight-sample solve coverage. We also partition problems by observed eight-sample reliability into \emph{Stable} (8/8 correct), \emph{Partial} (1--7/8 correct), and \emph{Unsolved} (0/8 correct). Building on this benchmark and multi-sample protocol, Section~\ref{sec:or-bench-post-training} diagnoses base-model failure modes and develops our solver-verified synthesis and post-training pipeline.

% !TeX encoding = UTF-8
% Integrated chapter. The host document owns all counters.
% Required packages are loaded in main.tex.
\section{Solver-Grounded Post-Training for Operations Research}\label{sec:or-bench-post-training}

The training recipe in this section is \textbf{Base $\rightarrow$ cold-start SFT $\rightarrow$ GRPO}, where Base denotes DeepSeek-V4-Flash-0731 (initialized directly without the Part~I CPT checkpoint) and GRPO runs sequentially in Chat and Thinking modes. We first sample the Base model repeatedly and examine unsuccessful formulations to identify missing or unreliable OR modeling requirements. Training data are then constructed to broaden coverage across decision scenarios and formulation requirements, and post-training evaluation revisits which requirements become solvable or more reliable.

We characterize problem requirements through four label layers---scenario, modeling technique, constraint families, and modeling pitfalls (Section~\ref{sec:ds-tags})---while tracking empirical solution reliability through \emph{Stable}, \emph{Partial}, and \emph{Unsolved} sampling regimes. This section connects capability diagnosis to verified synthesis and sequential post-training, evaluates the resulting model via pass@$k$ and paired per-question transitions, and audits training--test non-overlap in Section~\ref{sec:ds-contamination}.

\subsection{Base-Model Reliability Regimes under Diagnostic Sampling}
\label{sec:base-capability-strata}

To illustrate the reliability gap that post-training must address, we first examine the Base model under repeated sampling on the 93-instance \textsc{SLAI-OR-Bench-Hard} subset (Section~\ref{sec:benchmark_dataset}), independently sampling eight solutions per question to analyze both aggregate coverage and per-question reliability.

\begin{figure}[!htbp]
  \centering
  \includegraphics[width=0.90\linewidth]{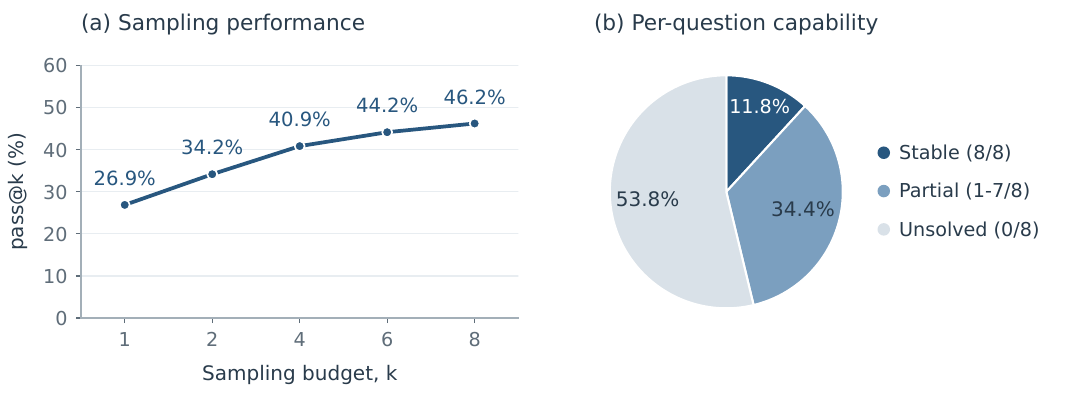}
  \caption{Base-model empirical solution reliability on \textsc{SLAI-OR-Bench-Hard} across 8 independent samples per question. (a) $\mathrm{pass}@k$ for $k \in \{1,2,4,6,8\}$. (b) Distribution of questions across Stable (8/8 solved), Partial (1--7/8 solved), and Unsolved (0/8 solved) regimes.}
  \label{fig:base-capability-analysis}
\end{figure}

We measure aggregate performance using pass@$k$ across $k\in\{1,2,4,6,8\}$. Figure~\ref{fig:base-capability-analysis}(a) shows pass@1 of 26.88\% and pass@8 of 46.24\%, a gain of 19.36 percentage points, indicating that the base model already possesses latent modeling knowledge on a substantial subset of problems where single-attempt generation fails.

To capture per-question reliability, we partition problems into three empirical eight-sample regimes, as shown in Figure~\ref{fig:base-capability-analysis}(b):
\begin{itemize}
  \setlength{\itemsep}{1pt}
  \setlength{\parskip}{0pt}
  \item \textbf{Stable (8/8 solved):} 11.83\% of questions (11/93).
  \item \textbf{Partial (1--7/8 solved):} 34.41\% of questions (32/93).
  \item \textbf{Unsolved (0/8 solved):} 53.76\% of questions (50/93).
\end{itemize}

Pairing these empirical reliability regimes with our four-layer structural labels pinpoints which constraint families and semantic pitfalls remain unexpressed or brittle. We apply the same empirical solve-rate principle to candidate training problems in Section~\ref{sec:ds-trajectories}: verified corrections and low-pass demonstrations seed SFT, partially solved prompts drive GRPO, and selected high-reliability records preserve mastered patterns.

\subsection{Taxonomy-Guided, Solver-Verified Data Synthesis}\label{sec:data-synthesis}

Repeated sampling and error analysis reveal formulation requirements that the base model handles unreliably. Unlike prior AI4OR synthesis pipelines that rely primarily on forward or bidirectional SFT data expansion~\citep{huang2025orlm,yang2025optibench,lu2025optmath} or process-reward preference learning~\citep{zhou2025steporlm,zhao2026sage}, our pipeline couples a structured four-layer OR taxonomy with one-hop operator-controlled generation, intermediate \texttt{FlatMILP} alignment checks, dual-interface solver execution, and empirical base-model solve-rate routing across SFT and sequential GRPO. The design objective is broad coverage of scenario--technique combinations across the OR taxonomy, using benchmark coverage for descriptive comparison rather than as the synthesis target.

\subsubsection{Capability Gaps and Four-Layer Label Taxonomy}\label{sec:ds-tags}

We characterize OR problems using four label layers: scenario, modeling technique, constraint families, and modeling pitfalls. Whereas Section~\ref{sec:benchmark_dataset} catalogs benchmark coverage across 26 fine-grained decision-task families, here we consolidate related problem structures into 21 synthesis \textbf{scenarios} (19 of which appear in \textsc{SLAI-OR-Bench-Hard}) paired with 31 \textbf{modeling techniques} specifying their mathematical representations; \textbf{constraint families} identify rules that the model must express; and \textbf{modeling pitfalls} capture operational semantics that are frequently misformulated. The original public-pool annotation assigns each problem one primary scenario and one primary modeling technique, with multiple constraint families and pitfalls allowed; for final training exports, all explicitly supported scenario--technique pairs in multi-requirement statements are recorded to track data coverage. Table~\ref{tab:ds-tags} summarizes the taxonomy and its role in synthesis.

\begin{table}[!htbp]
\centering\small
\setlength{\tabcolsep}{5pt}
\captionsetup{name=Table}
\caption{Label taxonomy linking capability gaps to synthesis design.}\label{tab:ds-tags}
\begin{tabularx}{\linewidth}{@{}L{0.22\linewidth}C{0.12\linewidth}L{0.30\linewidth}Y@{}}
\toprule
\textbf{Layer} & \textbf{Per problem} & \textbf{Inventory} & \textbf{Design role}\\
\midrule
Scenario & One primary & 21 & Decision task\\[3pt]
Modeling technique & One primary & 31 & Mathematical formulation\\[3pt]
Constraint families & Multiple & 50 & Required modeling rules\\[3pt]
Modeling pitfalls & Multiple & 195 distinct phrasings; 7 recurring & Error-prone semantics\\
\bottomrule
\end{tabularx}
\vspace{4pt}
\end{table}

For example, a berth-assignment problem may use an interval non-overlap formulation with time windows, resource eligibility, and a buffer after departure. The scenario and modeling technique define its decision structure, while constraint families and modeling pitfalls specify what the model must express. This turns capability gaps into concrete modeling requirements that guide subsequent problem modification and verification.

\subsubsection{Four-Stage Synthesis and Dual-Solver Verification}\label{sec:ds-pipeline}

\begin{figure}[!htbp]
\centering
\resizebox{0.90\linewidth}{!}{% Editable vector diagram: reference palette, two-panel publication layout.
% Requires TikZ and arrows.meta. Drawn at its final width of 160 mm.
\begingroup
\definecolor{spBerry}{HTML}{B44677}
\definecolor{spTeal}{HTML}{008B6A}
\definecolor{spInk}{HTML}{132B53}
\definecolor{spMuted}{HTML}{60748C}
\definecolor{spPale}{HTML}{EAF6FA}
\definecolor{spShadow}{HTML}{D2EBF3}
\definecolor{spLine}{HTML}{C5E0E8}
\begin{tikzpicture}[
 x=0.90mm,y=0.90mm,line cap=round,line join=round,
 font=\sffamily\fontsize{8.5}{10.2}\selectfont,
 flow/.style={draw=spBerry,line width=1.1pt,-{Stealth[length=2mm,width=1.5mm]}},
 fork/.style={draw=spBerry,line width=0.85pt,-{Stealth[length=1.4mm,width=1.1mm]}},
 title/.style={anchor=west,inner sep=0pt,text=spBerry,font=\sffamily\bfseries\fontsize{9.7}{11.5}\selectfont},
 body/.style={anchor=west,inner sep=0pt,text=spInk},
 note/.style={inner sep=0pt,text=spMuted,font=\sffamily\fontsize{8.1}{9.8}\selectfont},
 card/.style={draw=spTeal,fill=white,line width=0.75pt,rounded corners=6pt}
]
\path[use as bounding box] (0,-7) rectangle (160,99);

% Two purposeful regions replace the old stack of full-width bars.
\path[draw=spLine,fill=spPale!28,line width=0.5pt,rounded corners=4pt] (0,-6) rectangle (102,89);
\path[draw=spLine,fill=spPale!28,line width=0.5pt,rounded corners=4pt] (109,-6) rectangle (160,89);
\path[fill=spBerry,rounded corners=2pt] (0,91) rectangle (102,98);
\path[fill=spBerry,rounded corners=2pt] (109,91) rectangle (160,98);
\node[text=white,font=\sffamily\bfseries\fontsize{10.3}{12}\selectfont] at (51,94.5) {Label-guided data synthesis};
\node[text=white,font=\sffamily\bfseries\fontsize{9.5}{11}\selectfont] at (134.5,94.5) {Sampling and routing};

% A -> B -> C -> D follows one U-shaped path, ending beside the record store.
\foreach \x/\y/\letter/\name in {3/50/A/Parent Pool,3/0/B/Sampling,55/0/C/Generation,55/50/D/Gating}{
 \path[fill=spShadow,rounded corners=6pt] (\x+0.65,\y-0.7) rectangle (\x+44.65,\y+35.3);
 \path[card] (\x,\y) rectangle (\x+44,\y+36);
 \node[circle,fill=spBerry,text=white,minimum size=4.8mm,inner sep=0pt,font=\sffamily\bfseries\fontsize{9.2}{10.5}\selectfont] at (\x+5.2,\y+30.5) {\letter};
 \node[title] at (\x+9.5,\y+30.5) {\name};
 \draw[spTeal,line width=0.55pt] (\x+3,\y+25.6) -- (\x+41,\y+25.6);
 \foreach \dy in {19.5,13.6,7.7}{\fill[spTeal] (\x+4.1,\y+\dy) circle[radius=0.32mm];}
}
\node[body] at (10,69.5) {Seed problems};
\node[body] at (10,63.6) {Labels + metadata};
\node[body] at (10,57.7) {Fixed pool; no feedback};
\node[body] at (10,19.5) {Parent + mutation dims};
\node[body] at (10,13.6) {Structural operators};
\node[body] at (10,7.7) {Candidate plan};
\node[body] at (62,19.5) {LLM inference endpoint};
\node[body] at (62,13.6) {Semantic plan};
\node[body] at (62,7.7) {Problem $\to$ FlatMILP};
\node[body] at (62,69.5) {Code $\to$ COPT};
\node[body] at (62,63.6) {Returned objective (GT)};
\node[body] at (62,57.7) {Rules / neighbors / quota};
\draw[flow] (25,49.3) -- (25,36.7);
\draw[flow] (47.7,18) -- (54.3,18);
\draw[flow] (77,36.7) -- (77,49.3);

% Accepted records: a compact native vector cylinder.
\path[fill=spShadow] (114.65,76.3) -- (114.65,62.3)
 .. controls (114.65,58.3) and (155.65,58.3) .. (155.65,62.3)
 -- (155.65,76.3) .. controls (155.65,80.3) and (114.65,80.3) .. cycle;
\path[fill=white] (114,63) rectangle (155,77);
\path[draw=spTeal,fill=white,line width=0.8pt] (114,77) -- (114,63)
 .. controls (114,59) and (155,59) .. (155,63) -- (155,77);
\draw[draw=spTeal,fill=white,line width=0.8pt] (134.5,77) ellipse[x radius=20.5mm,y radius=3mm];
\node[text=spBerry,font=\sffamily\bfseries\fontsize{9.4}{11}\selectfont] at (134.5,71.5) {Accepted records};
\node[text=spInk,font=\sffamily\fontsize{8.3}{10}\selectfont] at (134.5,67) {Problem + GT + code};
\node[note] at (134.5,62.9) {Verification evidence};
\draw[flow] (99.7,68.8) -- (113.3,68.8);

% The shared base-model stage determines both nonexclusive training uses.
\path[fill=spShadow,rounded corners=5pt] (113.65,33.3) rectangle (156.65,54.3);
\path[card] (113,34) rectangle (156,55);
\node[text=spBerry,font=\sffamily\bfseries\fontsize{9.2}{11}\selectfont] at (134.5,49.7) {Base-model sampling};
\draw[spTeal,line width=0.5pt] (116,45.7) -- (153,45.7);
\node[text=spInk,font=\sffamily\fontsize{8.4}{10}\selectfont] at (134.5,41.9) {Run code / compare GT};
\node[text=spInk,font=\sffamily\fontsize{8.4}{10}\selectfont] at (134.5,37.7) {acc + verified fallback};
\draw[flow] (134.5,59.3) -- (134.5,55.3);

% Parallel output nodes share a side fork, not a decision diamond.
\foreach \y in {17,-3}{
 \path[fill=spShadow,rounded corners=4pt] (113.55,\y-0.6) rectangle (154.55,\y+12.4);
 \path[draw=spTeal,fill=white,line width=0.7pt,rounded corners=4pt] (113,\y) rectangle (154,\y+13);
}
\node[title,font=\sffamily\bfseries\fontsize{9.4}{11}\selectfont] at (115.5,25.3) {RL};
\node[body,font=\sffamily\fontsize{8.5}{10}\selectfont] at (125.5,25.3) {Problem + GT};
\node[note,anchor=west] at (115.5,20.8) {Selected verified prompts};
\node[title,font=\sffamily\bfseries\fontsize{9.4}{11}\selectfont] at (115.5,5.3) {SFT};
\node[body,font=\sffamily\fontsize{8.2}{10}\selectfont] at (125.5,5.3) {Problem + trajectory};
\node[note,anchor=west] at (115.5,0.8) {Verified / revised traces};
\draw[draw=spBerry,line width=0.85pt] (156.3,43.5) -- (157.8,43.5) -- (157.8,3.5);
\draw[fork] (157.8,23.5) -- (154.3,23.5);
\draw[fork] (157.8,3.5) -- (154.3,3.5);
\end{tikzpicture}
\endgroup}
\caption{Overview of the label-guided data synthesis (Blocks A--D) and trajectory-routing pipeline. Candidate generation incorporates stage-level checks and bounded revisions, after which base-model sampling and solver verification route accepted problems to SFT and RL (Section~\ref{sec:ds-trajectories}).}\label{fig:ds-pipeline}
\end{figure}
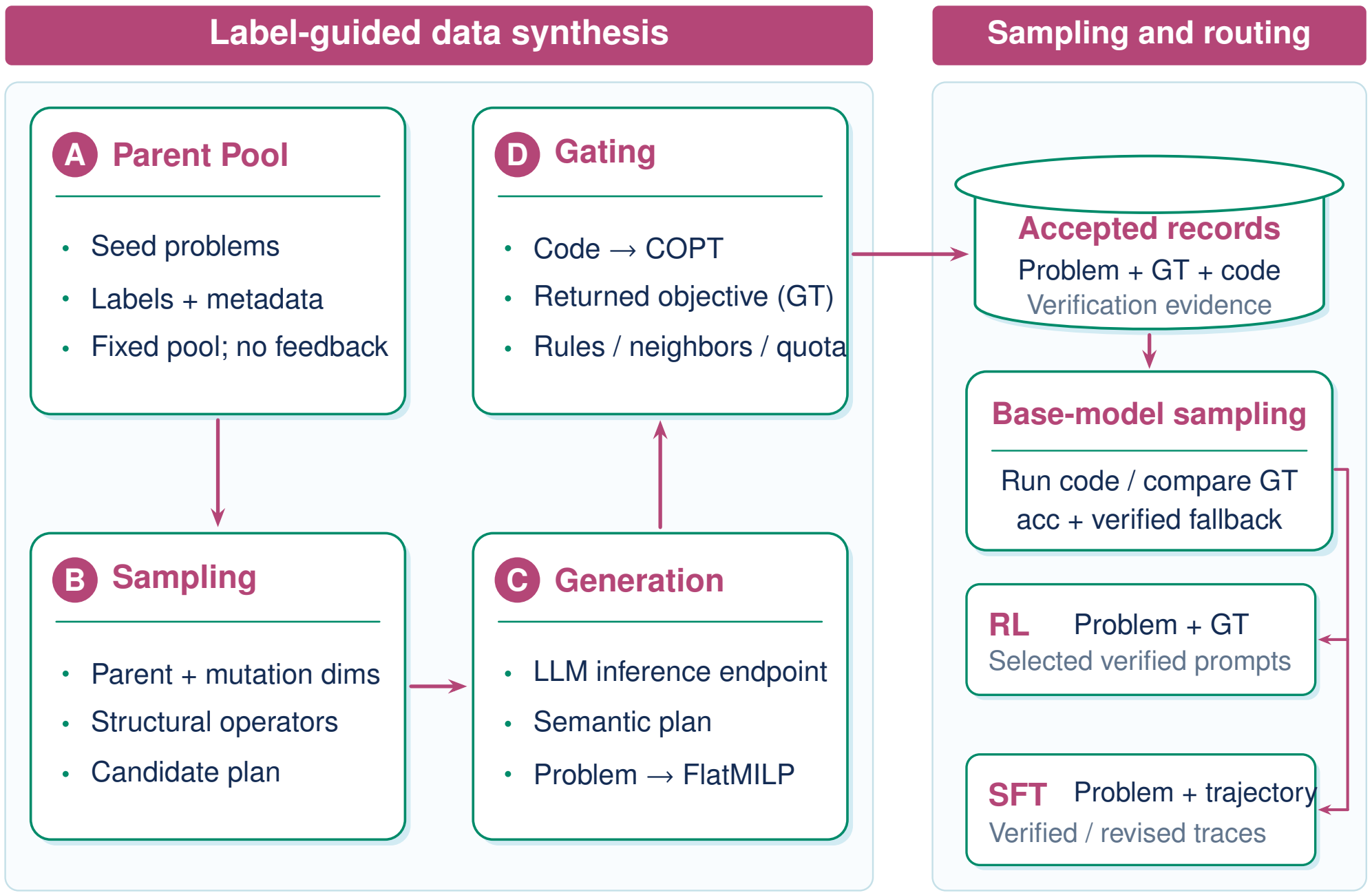

The synthesis pipeline comprises four blocks: A, the parent pool; B, sampling; C, generation; and D, gating. Figure~\ref{fig:ds-pipeline}, shown after the final training-data summary, summarizes the complete process. Candidates that pass the gates enter the accepted-record pool and proceed to base-model sampling and training-data routing in Section~\ref{sec:ds-trajectories}.

\textbf{A. Parent Pool.} A parent is a curated seed problem (drawn from labeled public optimization collections and internal reference templates, strictly excluding all \textsc{SLAI-OR-Bench} instances and their source variants) paired with its statement, provenance, and structural labels. Scenario and technique labels characterize the decision structure, while constraint families and pitfalls specify rules that can be modified or combined. The pool remains fixed during each run: accepted synthetic records are stored separately and are not fed back as new parents. This one-hop design limits the risk of recursive formulation drift and keeps similarity checks and per-parent quotas anchored to a traceable reference set. Sampling weights account for structural sparsity, parent use and remaining quotas, and opportunities to introduce new constraint families. Training--test checking is reported separately in Section~\ref{sec:ds-contamination}.

\textbf{B. Sampling.} The program first selects a parent problem, then samples variation dimensions and structural operators to form a candidate plan. The domain, entities, numerical values, and narrative are resampled each time. Structural, pitfall, and scale variations are selected according to prior probabilities, with at least one enabled. Sampling preserves the parent problem's decision structure and specifies what the new problem must change.

A structural operator specifies a concrete modification, such as introducing multiple periods, shared capacity, or fixed activation costs. Each operator declares the required constraint families and a minimum number of constraints. When structural variation is enabled, the constraint-family list must differ from that of the parent problem. Table~\ref{tab:ds-operators} groups the operators into five categories by the object of change. Generation and gating check compliance with these requirements.

\begin{table}[!htbp]
\centering\small
\setlength{\tabcolsep}{5pt}
\captionsetup{name=Table}
\caption{Structural operators and the modeling changes they require.}\label{tab:ds-operators}
\begin{tabularx}{\linewidth}{@{}L{0.14\linewidth}L{0.36\linewidth}Y@{}}
\toprule
\textbf{Group} & \textbf{Example changes} & \textbf{Required model changes}\\
\midrule
Time & Multiple periods; time windows & Links between states or resources across periods; time bounds\\[3pt]
Capacity & Shared capacity; bounded overflow & Aggregate resource constraints; overflow variables and upper bounds\\[3pt]
Cost & Fixed costs; conditional penalties & Activation--usage links; corresponding objective terms\\[3pt]
Logic & If--then rules; mutually exclusive choices & Conditional constraints; selection and cardinality limits\\[3pt]
Coupling & Shared resources; carryover across periods & Balance and linking constraints across decisions or periods\\
\bottomrule
\end{tabularx}
\end{table}

\textbf{C. Generation.} The candidate plan is submitted to an LLM inference endpoint to generate, in sequence, a structured semantic plan, a new problem statement, and a FlatMILP representation. The semantic plan specifies variable meanings, the objective, and constraint requirements; the problem statement supplies all data needed to solve the problem; and FlatMILP expresses these requirements as variables, an objective, and linear constraints. Each stage is checked against the candidate plan. Failure reasons are returned to the corresponding stage for a bounded number of revisions; candidates that still fail are rejected. Executable code is converted from FlatMILP, and the solver provides the reference objective.

\textbf{D. Gating.} Table~\ref{tab:ds-gates} lists eight admission checks. Failures in the plan, problem, or model checks are returned to the corresponding LLM generation stage for bounded revision; candidates are rejected when the revision limit is reached. Failures in the remaining checks do not trigger model revision. The code and solve checks share execution evidence.

\begin{table}[!htbp]
\centering\small
\setlength{\tabcolsep}{5pt}
\captionsetup{name=Table}
\caption{Gating criteria for accepting a synthesized candidate.}\label{tab:ds-gates}
\begin{tabularx}{\linewidth}{@{}L{0.16\linewidth}Y@{}}
\toprule
\textbf{Gate} & \textbf{Acceptance criterion}\\
\midrule
Plan & Preserve the decision structure specified by the candidate plan and change the domain as planned. The semantic plan must incorporate the specified constraint families, structural changes, and modeling pitfalls. If structural variation is enabled, the constraint-family list must differ from the parent problem's.\\[4pt]
Problem & The statement is self-contained and excludes excessively short text, placeholders, external files, attachments, URLs, implicit road networks, and ambiguous quantities. Data must be provided within the statement as prose or tables, as planned.\\[4pt]
Model & The objective direction and constraint families in FlatMILP correspond to the semantic plan. Added constraint families must be declared and present in the model; additions must stay within the declared scope. Each operator's minimum constraint count must be met.\\[4pt]
Scale & Only when scale variation is enabled, check that variable and constraint counts and the scale tier meet the configured requirements of the selected tier.\\[4pt]
Code & Convert the same FlatMILP into two code formats, check syntax, and record execution results for each. Retain code that passes the checks, together with its interface information.\\[4pt]
Solve & Obtain an objective value returned by COPT through the corresponding code path, record successful solving for that path, and save the reference objective as GT.\\[4pt]
Neighbors & Filter near neighbors by statement and structural similarity, using the thresholds below. Structural comparison is performed only when structural variation is enabled and excludes the parent problem itself.\\[4pt]
Quota & Once the number of accepted records from a parent problem reaches the configured limit, accept no further candidates from that parent.\\
\bottomrule
\end{tabularx}
\end{table}

Across four principal synthesis runs, deduplication by candidate ID yields 7,008 terminal candidates: 2,807 were accepted (40.05\%) and 4,201 were not retained (59.95\%). Table~\ref{tab:ds-gate-statistics} reports terminal rejection attribution across these runs (earlier production batches used Gurobi as the numerical backend before standardizing on the COPT dual-interface configuration below; finalized records retain the verified reference objectives from their respective solver runs). Because Code and Solve share execution logs, their rejections are combined as shares of terminal candidates.

\begin{table}[!htbp]
\centering\small
\caption{Terminal candidate rejection counts and shares across the four principal synthesis runs.}\label{tab:ds-gate-statistics}
\begin{tabular*}{\linewidth}{@{\extracolsep{\fill}}lrrr@{}}
\toprule
\textbf{Final rejection attribution} & \textbf{Count} & \textbf{All candidates} & \textbf{Not retained}\\
\midrule
Plan & 243 & 3.47\% & 5.78\%\\
Problem & 153 & 2.18\% & 3.64\%\\
Model & 155 & 2.21\% & 3.69\%\\
Scale & 126 & 1.80\% & 3.00\%\\
Code + Solve & 3,131 & 44.68\% & 74.53\%\\
Neighbors & 35 & 0.50\% & 0.83\%\\
Quota & 49 & 0.70\% & 1.17\%\\
Worker exception & 309 & 4.41\% & 7.36\%\\
\midrule
\textbf{Not retained} & \textbf{4,201} & \textbf{59.95\%} & \textbf{100.00\%}\\
\bottomrule
\end{tabular*}
\par\smallskip
\begin{minipage}{\linewidth}\footnotesize\raggedright
Denominators are 7,008 terminal candidates and 4,201 non-retained candidates. Accepted-pool counts track candidate problems, whereas final training exports count problem--interface records after supplementary curation.
\end{minipage}
\end{table}

\emph{Code and solver execution.} The two code formats converted from FlatMILP use the following execution paths:
\begin{itemize}
\setlength{\itemsep}{2pt}\setlength{\parskip}{0pt}
\item \textbf{Pyomo:} Call COPT through \texttt{copt\_direct}.
\item \textbf{gurobipy:} Export LP/MPS and pass it to COPT. gurobipy provides the modeling and export interface; COPT performs numerical optimization.
\end{itemize}
COPT serves as the common numerical backend for both interfaces and supplies each newly generated instance's reference objective (GT). Combining stage-level plan--statement--\texttt{FlatMILP} gates with dual-interface execution and subsequent independent statement-to-code reconstruction cross-validates both statement semantics and reference objective values.

\emph{Near-neighbor filtering.} For a candidate and problems in the seed pool, statement and structural similarities are computed as:
\begin{align*}
s_{\mathrm{text}} &= \max\!\left(J_{w,2},\;0.65J_w+0.35J_{c,5}\right),\\
s_{\mathrm{struct}} &= 0.72J_F+0.28\cos(\mathbf v_x,\mathbf v_y).
\end{align*}
Here, $J_{w,2}$ is word-bigram Jaccard similarity, $J_w$ is word-set Jaccard similarity, and $J_{c,5}$ is character-5-gram Jaccard similarity. $J_F$ is constraint-family Jaccard similarity, and $\mathbf v$ is a vector of scale features. A candidate is rejected if statement similarity reaches $0.72$ or structural similarity reaches $0.94$. Statement comparison uses the entire pool. Structural comparison excludes the parent problem and is performed only when structural variation is enabled.

\emph{Scope of checks.} Whether every constant is explicitly included in the statement, and whether all constraint families retained from the parent but not named by an operator are labeled, are recorded as warnings and do not independently cause rejection. Review, infeasibility-conflict repair, dual-reference checks, and weak-relaxation checks are optional and are not mandatory gates. Subsequent independent base-model sampling provides further verification.

\textbf{Outputs and Answer Verification.} We use the same objective-matching rule for independent synthesis verification, trajectory admission, RL reward scoring, and benchmark evaluation:
\begin{equation}
  \lvert \widehat{z}-z^\star\rvert \leq 10^{-4}+10^{-6}\lvert z^\star\rvert,
  \label{eq:objective-tolerance}
\end{equation}
where $\widehat{z}$ is the executed objective and $z^\star$ is the reference objective (distinct from the numerical solver's internal feasibility and optimality tolerances). Because Block~B resamples domains, entities, numerical parameters, and structural constraints, each synthesized candidate defines a brand-new optimization instance rather than reusing the parent seed's numerical answer. Its reference objective (GT) is derived from the newly generated \texttt{FlatMILP} via dual-format code compilation (Pyomo and gurobipy) and COPT numerical optimization in Block~D. To ensure that this solver-derived reference objective faithfully reflects the natural-language problem statement, stage-level gates first check plan--statement--\texttt{FlatMILP} alignment (Table~\ref{tab:ds-gates}), and Section~\ref{sec:ds-trajectories} subsequently performs independent text-to-code verification: the base model (or a teacher/expert fallback) must independently reconstruct solver code from the problem statement alone and match the COPT reference objective before the instance is admitted to SFT or RL training. Verified compiled code from Block~D can also be exported directly as code-answer SFT data when trajectory sampling is not required.

\subsubsection{Empirical Solve-Rate Sampling and Data Routing}\label{sec:ds-trajectories}

\paragraph*{Base-Model Sampling}
For each accepted problem--modeling-interface pair from Section~\ref{sec:ds-pipeline}, the base model independently samples multiple solution responses. Each attempt receives only the natural-language problem statement (without access to the semantic plan, \texttt{FlatMILP}, or reference objective) and generates a reasoning trace and executable modeling code. The code is executed, its returned objective is compared against the COPT reference objective from Block~D using Eq.~\eqref{eq:objective-tolerance}, and responses that pass verification are retained. This step simultaneously measures the problem's empirical solve rate, $\mathrm{acc}$ (the fraction of sampled responses whose objectives match the reference), and cross-verifies that the problem statement and reference objective are mutually consistent.

\paragraph*{SFT and RL Routing}
SFT uses problem statements paired with verified reasoning and modeling code; RL uses statements and reference objectives, with responses generated by the current policy. Table~\ref{tab:ds-routing}(a) gives the principal use of each empirical Base $\mathrm{acc}$ group (evaluated under its source-pool sampling budget), and part (b) reports the final training exports after supplementary revision and verification.

For problems with Base $\mathrm{acc}=0$, we apply Grok-4.6-assisted correction followed by human revision, admitting corrected problem statements and trajectories only after execution matches the reference objective under Eq.~\eqref{eq:objective-tolerance}. This verified group (which retains its initial $\mathrm{acc}=0$ stratum label) contributes 405 SFT problem--interface records, of which 313 and 263 also enter Chat RL and Thinking RL, respectively.

\begin{table}[!htbp]
\centering\small
\setlength{\tabcolsep}{5pt}
\caption{Training-data routing and final composition.}\label{tab:ds-routing}
\textbf{(a) Principal training use}\par\smallskip
\begin{tabularx}{\linewidth}{@{}L{0.21\linewidth}L{0.18\linewidth}Y@{}}
\toprule
\textbf{Base-model acc} & \textbf{Training use} & \textbf{Supervision / verification}\\
\midrule
$\mathrm{acc}=0$ & LLM / human correction & Admit corrected problems and trajectories only after execution and reference-objective verification.\\[3pt]
$0<\mathrm{acc}\leq0.20$ & SFT + RL & Correct sampled trajectories demonstrate solution methods; verified prompts also support online RL.\\[3pt]
$0.20<\mathrm{acc}<1$ & Primarily RL & Statements and reference objectives support improved sampling reliability; selected correct demonstrations may enter SFT.\\[3pt]
$\mathrm{acc}=1$ & Selectively retained & Retain selected verified records to preserve already reliable solution patterns.\\
\bottomrule
\end{tabularx}
\par\medskip
\textbf{(b) Final problem--interface records}\par\smallskip
\begin{tabularx}{\linewidth}{@{}Yrrr@{}}
\toprule
\textbf{Base acc / record group} & \textbf{SFT} & \textbf{Chat RL} & \textbf{Thinking RL}\\
\midrule
$\mathrm{acc}=0$ & 405 (28.72\%) & 313 (13.47\%) & 263 (20.64\%)\\
$0<\mathrm{acc}\leq0.20$ & 418 (29.65\%) & 371 (15.96\%) & 289 (22.68\%)\\
$0.20<\mathrm{acc}<1$ & 149 (10.57\%) & 1,343 (57.79\%) & 690 (54.16\%)\\
$\mathrm{acc}=1$ & 438 (31.06\%) & 297 (12.78\%) & 32 (2.51\%)\\
\midrule
\textbf{Total} & \textbf{1,410} & \textbf{2,324} & \textbf{1,274}\\
\bottomrule
\end{tabularx}
\par\smallskip
\begin{minipage}{\linewidth}\footnotesize\raggedright
Percentages use each export's total. The Base $\mathrm{acc}=0$ row contains the verified correction group. Exports overlap and are not disjoint counts of questions; rounded percentages may not sum exactly to 100\%.
\end{minipage}
\end{table}

\paragraph*{Chat and Thinking RL Data}
Chat GRPO uses a broader pool of 2,324 records to consolidate accessible formulation patterns efficiently, after which Thinking GRPO focuses extended reasoning on a 1,274-record subset contained within the Chat pool (SFT shares 1,068 problem--interface pairs with Chat RL and 656 with Thinking RL), connecting data routing directly to our sequential training recipe in Section~\ref{sec:post-training}.

\begin{figure}[!htbp]
\centering
\includegraphics[width=\linewidth]{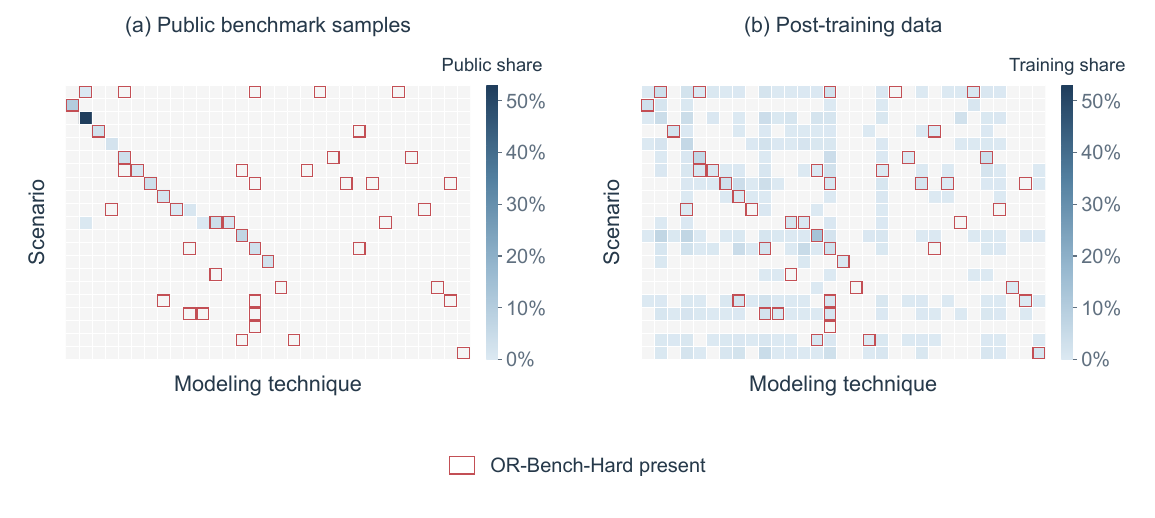}
\caption{Scenario--technique coverage across data sources. (a) Original labeled public benchmark samples (primary labels); (b) final post-training records (including multi-pair statements, so cell shares overlap). Blue shading gives the proportion of each panel's records involving a cell on a common linear scale; light gray denotes zero observed records; red outlines indicate \textsc{SLAI-OR-Bench-Hard} cells.}\label{fig:ds-coverage}
\end{figure}

\paragraph*{Coverage of the Final Training Data}
Figure~\ref{fig:ds-coverage} compares the original public-pool distribution with the modeling requirements represented in the final training exports. Synthesis broadens coverage across the shared OR taxonomy---spanning production and extraction, allocation, routing, facility location, inventory, transfers, computing-task placement, communication routing, and project maintenance, paired with fixed charges, network flows, inventory balances, time-indexed decisions, assignments, and resource-constrained scheduling. Multiple explicitly supported requirements may coexist in a statement, so a training record can occupy several cells.

Remaining blank cells reflect finite synthesis budgets and quality-gating exclusions across the $21 \times 31$ matrix; Section~\ref{sec:evaluation-scenario-transfer} evaluates how this training distribution translates into post-training reliability gains across scenarios.

\subsubsection{Lexical and Structural Non-Overlap Audit}\label{sec:ds-contamination}
The near-neighbor gate in Block~D (\emph{Near-neighbor filtering}) screens generated candidates against the seed pool. After supplementary correction and verification, we also audit \emph{every finalized training record} against all 393 distinct text versions of the 131 \textsc{SLAI-OR-Bench} instances. This final-export audit covers all 1,410 SFT, 2,324 Chat-RL, and 1,274 Thinking-RL records (5,008 records and 1,685 distinct statements after interface-template removal).

Normalized exact matching, number-masked template matching, and text-neighbor screening at similarity $\geq0.72$ each flag \textbf{0 records (0\%) in every export}, confirming zero direct duplicates or above-threshold lexical neighbors between the finalized training files and \textsc{SLAI-OR-Bench}.

At the taxonomy level, 36 of the 45 distinct Hard scenario--technique pairs appear in training (with 786 SFT, 1,161 Chat-RL, and 705 Thinking-RL records sharing at least one scenario--technique pair with the test set), reflecting shared mathematical formulation families rather than instance overlap. A finer source-level check over retained constraint-family profiles for 60 historical candidates and 84 benchmark-matching sources (5,040 pairs) linked to 72 final records (27 SFT, 30 Chat-RL, and 15 Thinking-RL) bounds the Block~D structural score via $0.72J_F\leq s_{\mathrm{struct}}\leq0.72J_F+0.28$ (where archived scale vectors $\mathbf{v}$ were omitted). For 70 of the 72 records, even this conservative upper bound remains strictly below the $0.94$ threshold against all 84 benchmark sources; for the remaining two SFT records, the interval straddles $0.94$ solely due to missing scale vectors, while both pass all exact, number-masked template, and lexical near-neighbor ($<0.72$) checks above.

\subsection{Cold-Start SFT and Failure-Tiered Multi-Stage GRPO}\label{sec:post-training}

Post-training follows Base $\rightarrow$ cold-start SFT $\rightarrow$ GRPO. SFT supplies verified formulation demonstrations, followed by Chat GRPO on the broader prompt pool and Thinking GRPO on its selected subset. Failure reward tiers distinguish usable execution outcomes, while outcome masking excludes responses whose correctness cannot be judged reliably.

\subsubsection{Cold-Start Supervised Fine-Tuning}\label{sec:pt-cold-start}

Cold-start SFT demonstrates constraint formulations and modeling structures, aiming to raise the probability of sampling useful trajectories during subsequent GRPO. It uses 1,410 verified problem--interface demonstrations, including 405 from the verified Base $\mathrm{acc}=0$ correction group. Training initializes directly from Base and runs for 25 updates with learning rate $1.25\times10^{-6}$, global batch size 64, micro batch size 1, and maximum sequence length 65,536 tokens. We use two warmup updates and cosine decay to a minimum learning rate of $1.25\times10^{-8}$. The training schedule consumes 1,600 sample indices, covering all 1,410 records with 190 repeated indices. The checkpoint comparisons in Section~\ref{sec:evaluation-sampling-migration} examine the resulting SFT-only and combined outcomes.

\subsubsection{Failure-Tiered Rewards, Outcome Masking, and Sequential GRPO}\label{sec:pt-reward-mask}

GRPO estimates relative advantages across multiple sampled responses to the same prompt. Rather than training a separate neural reward model, we execute the generated solver code and compare its returned objective value against the reference objective using Eq.~\eqref{eq:objective-tolerance}, the same absolute-plus-relative tolerance used in data verification and evaluation. As summarized in Table~\ref{tab:pt-reward-mask}, responses that yield a definitive execution outcome receive tiered scores based on correctness and failure mode, whereas responses affected by external resource or interface limits are masked out of advantage estimation.

\begingroup
\small
\setlength{\tabcolsep}{4.5pt}
\renewcommand{\arraystretch}{1.16}
\begin{table}[!htbp]
\caption{GRPO reward tiers and masking rules.}\label{tab:pt-reward-mask}
\centering\small
\begin{tabularx}{\linewidth}{@{}L{0.44\linewidth}C{0.08\linewidth}C{0.15\linewidth}Y@{}}
\toprule
\textbf{Outcome} & \textbf{Score} & \textbf{Advantage} & \textbf{Purpose} \\
\midrule
Objective matches the reference under Eq.~\eqref{eq:objective-tolerance} & $1$ & Included & Objective-level correctness \\[3pt]
A numeric solution is returned, but the objective does not match & $0$ & Included & Prefer executable errors to execution failures \\[3pt]
Infeasible or empty model; no code can be extracted & $-0.1$ & Included & Rank comparable failures \\[3pt]
Syntax error, unfinished code, JSON literals used as names, or overlong variable names & $-0.15$ & Included & Penalize avoidable code errors \\[3pt]
More than 2,000 variables or constraints, or a license size limit; HiGHS retry also fails & $0^{*}$ & Masked & Exclude resource-limited outcomes \\[3pt]
Solver runtime exceeds 120 seconds & $0^{*}$ & Masked & Do not infer correctness from a timeout \\[3pt]
Response reaches the output length limit & $0^{*}$ & Masked & Do not treat truncation as a modeling error \\[3pt]
HiGHS does not support the nonlinear or higher-order expression & $0^{*}$ & Masked & Avoid bias toward supported formulations \\
\bottomrule
\end{tabularx}
\par\smallskip
\begin{minipage}{\linewidth}\footnotesize\raggedright
$^{*}$ Zero is a placeholder only; masked responses are excluded from group statistics and advantage estimation. Data verification, reward scoring, and \textsc{SLAI-OR-Bench} evaluation share the absolute-plus-relative tolerance in Eq.~\eqref{eq:objective-tolerance}.
\end{minipage}
\end{table}
\endgroup

\paragraph{Failure Reward Tiers.}
When every valid response in a prompt group fails, binary rewards assign zero to all trajectories and yield no centered task-reward contrast. Failure tiers ($0$, $-0.1$, and $-0.15$) distinguish an executable but numerically incorrect formulation from infeasible or unextractable code and from basic syntax failures. Whenever multiple failure severities appear in an all-wrong group, subtracting the valid-response mean preserves positive and negative relative advantages toward executable formulations (groups whose valid responses all land in the same tier yield zero task-reward advantage, while the separate reference KL term remains active).

\paragraph{Outcome Masking.}
Assigning a fixed penalty to resource-limited or truncated rollouts would distort group statistics by penalizing valid large-scale formulations. Models that exceed the local Gurobi license size limit (2,000 variables or constraints) are automatically retried with HiGHS under a 60-second limit and a 20,000-size cap. If HiGHS returns a numeric solution, the response is scored under the first two tiers of Table~\ref{tab:pt-reward-mask}; otherwise, the trajectory is masked. Solver timeouts ($>120$\,s), expressions unsupported by HiGHS, and responses truncated at the token limit are likewise excluded from group advantage computation.

Together, Failure Reward Tiers and Outcome Masking separate gradable modeling errors from ungradable resource or interface limits. Group-mean advantage centering is computed solely over valid unmasked responses (Eq.~\eqref{eq:grpo-variants-baseline}); if all responses in a group are masked ($G_g=0$), the group contributes zero task-reward gradient.

\begin{figure}[!htbp]
\centering
\includegraphics[width=0.95\linewidth]{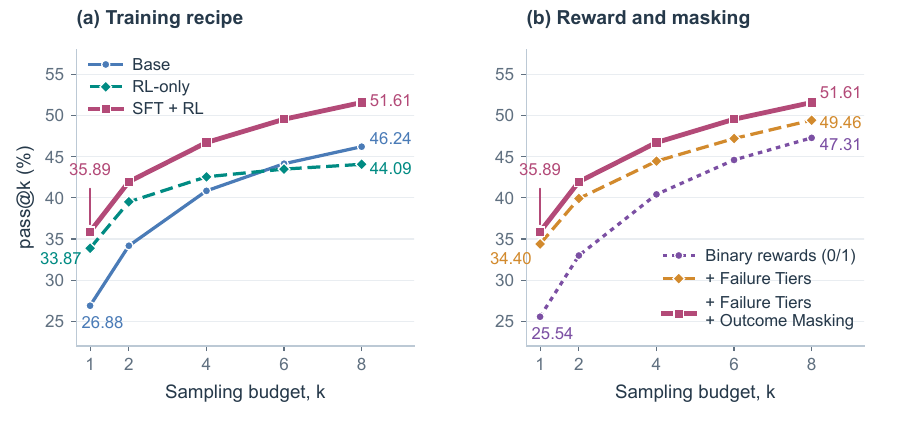}
\caption{Sampling performance on \textsc{SLAI-OR-Bench-Hard} at $k\in\{1,2,4,6,8\}$. (a) Base, RL-only, and SFT+RL. (b) Ablation evidence for the GRPO reward and masking design: binary rewards (0/1), Failure Tiers without masking, and Failure Tiers with Outcome Masking. The rose curves in both panels use the same main SFT+RL measurements and lead at every plotted budget.}\label{fig:pt-initialization-comparison}
\end{figure}

\Needspace{6\baselineskip}
\paragraph*{Ablation Evidence for the GRPO Reward and Masking Design}
Figure~\ref{fig:pt-initialization-comparison}(b) provides empirical support for the reward and masking design above. Failure Tiers give below-mean failures negative relative advantages, supplying a task-reward update that suppresses those responses even in an all-wrong group when failure scores differ. Outcome Masking restricts this feedback to gradable outcomes by excluding resource-limited, timed-out, or truncated responses from the group statistics.

The ablation compares binary rewards (0/1), Failure Tiers without masking, and the complete design used by the main SFT+RL model. Failure Tiers raise pass@1 from 25.54\% to 34.40\% and pass@8 from 47.31\% to 49.46\%. Adding Outcome Masking further raises these scores to 35.89\% and 51.61\%, gains of 1.49 and 2.15 percentage points over tiers alone. The full design leads at all five evaluated sampling budgets, exceeding binary rewards by 10.35 points at pass@1 and 4.30 points at pass@8. These results support combining graded failure feedback with the exclusion of ungradable outcomes in the main GRPO recipe, improving both single-attempt reliability and sampled solve coverage.

\begin{figure}[!htbp]
\centering
\includegraphics[width=0.95\linewidth]{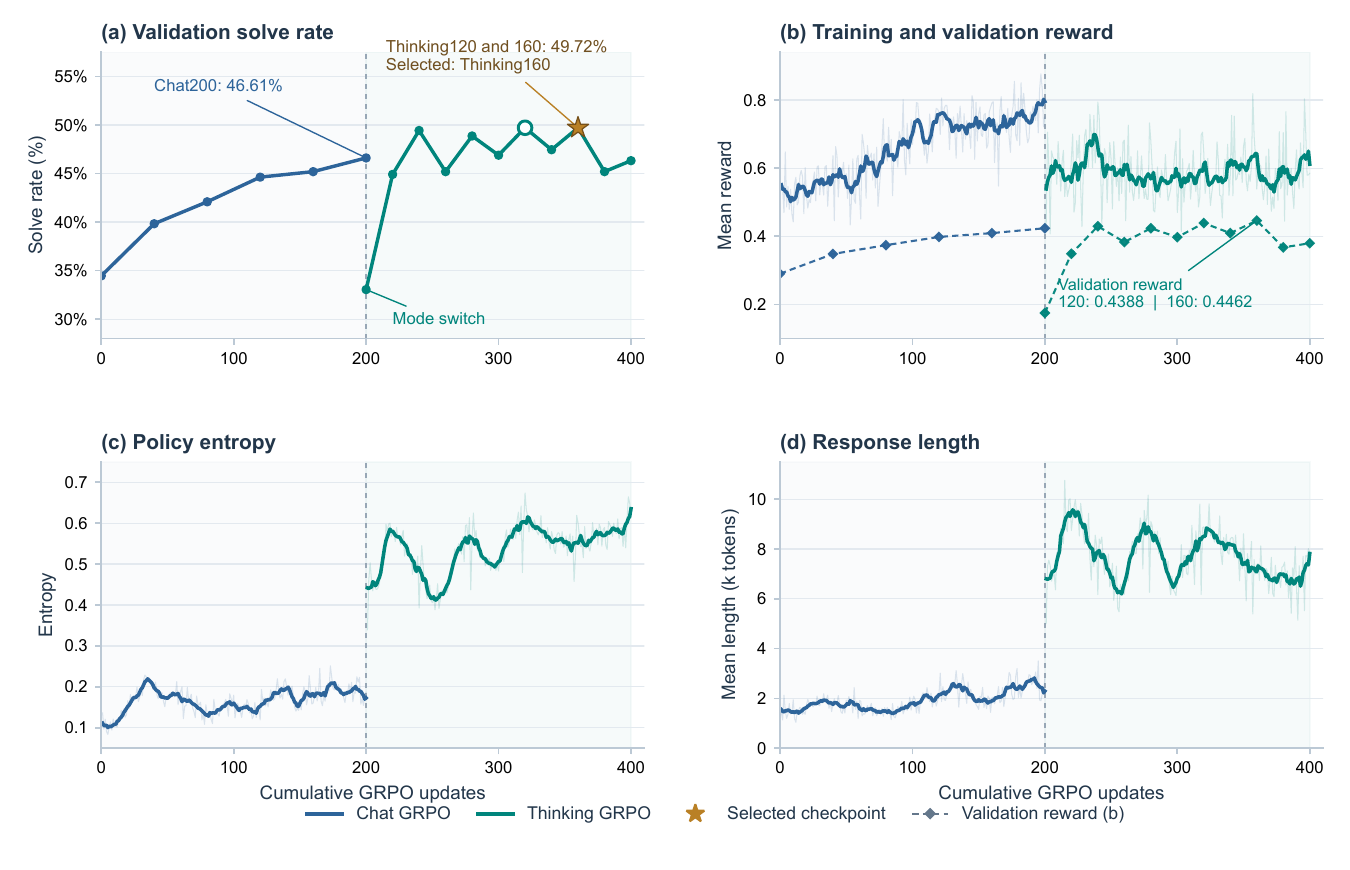}
\caption{Sequential Chat and Thinking GRPO training dynamics. (a) Validation solve rate; (b) training and validation reward; (c) policy entropy; (d) response length. The horizontal axis counts cumulative GRPO updates. Faint lines show raw training values and solid lines show stage-specific seven-step moving averages; validation solve rates are unsmoothed. Both stages use the same validation records, but generation templates and scoring paths change at the boundary, so the boundary evaluations are shown separately. The star marks Thinking160 (cumulative update 360), used for final test evaluation.}\label{fig:pt-baseline}
\end{figure}

Chat GRPO initializes from the 25-update cold-start SFT checkpoint, and Thinking GRPO initializes from Chat200. Thus both stages in Figure~\ref{fig:pt-baseline} belong to the SFT-initialized main run. Each stage runs for 200 updates, with 32 prompts per update and a nominal average of eight responses per prompt. They use learning rate $10^{-6}$, a separate reference-policy KL loss with coefficient $0.001$, prompt and response limits of 16,384 tokens each, rollout temperature 1, and top-$p$ 1. Validation uses 177 held-out seed problems from the curated parent pool (strictly disjoint from \textsc{SLAI-OR-Bench}), represented through both Pyomo and gurobipy interfaces (354 records), with greedy generation and one response per record.

\paragraph*{Training Outcomes}
Chat GRPO increases validation solve rate from 34.46\% at the SFT checkpoint to 46.61\% at Chat200. Switching to the Thinking generation template and scoring path re-evaluates the inherited Chat200 weights at 33.05\%. The solve rate rapidly recovers to 44.92\% at Thinking20 and 49.44\% at Thinking40, reaches 49.72\% at both local steps 120 and 160, and ends at 46.33\% at step 200.

\paragraph*{Training-Process Interpretation}
The boundary dip reflects initial format and length adaptation upon switching to the Thinking template and scoring path. Among the 354 validation responses at Thinking0, 98 (27.68\%) lack the closing thinking marker and 69 (19.49\%) have an empty thinking body, causing rejection prior to solver-code scoring; re-encoding the 98 unclosed responses with the run's tokenizer places all of them at the 16,384-token output limit. By Thinking40, empty thinking bodies disappear, unclosed responses drop to 42 (11.86\%), and solve rate rebounds to 49.44\%; by Thinking160, unclosed responses fall further to 26 (7.34\%).

\paragraph*{Selection of Thinking160}
Final test evaluation uses Thinking160 (cumulative update 360: 200 Chat updates plus 160 Thinking updates). Checkpoints 120 and 160 tie at the peak validation solve rate (49.72\%, ahead of 46.33\% at step 200), with step 160 exhibiting higher mean validation reward (0.4462 versus 0.4388) and a lower oversized-model failure rate (11.02\% versus 12.15\%).

\paragraph*{Comparison with RL without Cold-Start SFT}
Figure~\ref{fig:pt-initialization-comparison}(a) compares Base, RL-only, and the main SFT+RL recipe on the Hard set at $k\in\{1,2,4,6,8\}$. RL-only improves pass@1 from Base's 26.88\% to 33.87\%, whereas its pass@8 reaches 44.09\%, below Base's 46.24\%. SFT+RL leads at every plotted budget, reaching 35.89\% at pass@1 and 51.61\% at pass@8 ($+2.02$ and $+7.52$ percentage points over RL-only, albeit under distinct training schedules), improving both single-attempt precision and multi-sample solve coverage.

\subsubsection{Zero-Advantage Interventions in Group-Relative RL}
\label{sec:pt-variants}

Under group-mean advantage centering,
\begin{equation}
 A_i=r_i-b_g,\qquad
 b_g=\bar r_g=\frac{1}{G_g}\sum_{j\in I_g}r_j,
 \label{eq:grpo-variants-baseline}
\end{equation}
where $I_g$ denotes the $G_g=|I_g|$ unmasked responses to prompt $x_g$, both all-correct groups and uniform-score all-wrong groups yield zero centered task-reward advantages (though the separate reference KL penalty remains active). While Failure Reward Tiers (Table~\ref{tab:pt-reward-mask}) recover relative gradients whenever failure severities differ within a group, uniform groups still either expend rollout budget on already reliable prompts or provide no positive target on persistently unsolved prompts.

To test whether explicit zero-advantage interventions improve generalization, we compare standard group-mean GRPO (\textbf{Baseline}) against three targeted mechanisms in a controlled 200-update Chat-GRPO study initialized directly from Base (without cold-start SFT; 32 prompts and 256 nominal responses per update, evaluated every ten updates via greedy decoding on the 177-problem, 354-record seed validation set):
\begin{itemize}[leftmargin=1.8em, itemsep=2pt, topsep=3pt]
  \item \textbf{Downsample} reduces the nominal rollout allocation from 8 to 2 on prompts that were all-correct at their previous visit, redistributing saved slots round-robin across remaining prompts (up to 16 per prompt)~\citep{dapo,zheng2025greso,li2025knapsackrl}.
  \item \textbf{Replay} maintains a library of verified same-prompt successes and replaces one failed trajectory in each all-wrong group with a stored correct trajectory (re-scored at PPO ratio 1 without off-policy importance correction)~\citep{yan2025luffy,zhang2025rlep,v4_zhan2026exgrpo}.
  \item \textbf{Virtual Reward} replaces the group-mean baseline $b_g$ with a positive constant $c=0.1$ whenever $\max_{j\in I_g} r_j \leq 0$, rendering every failure advantage strictly negative while preserving relative failure-tier spacing~\citep{zhu2025nsr}.
\end{itemize}

\begin{table}[!htbp]
 \centering\small
 \setlength{\tabcolsep}{5pt}
 \renewcommand{\arraystretch}{1.15}
 \caption{Summary of Base-initialized Chat-GRPO variants for zero-advantage groups. Validation columns report greedy solve rates (\%) on the 354-record seed panel; parentheses denote the best checkpoint's update. Training reward, entropy, and response length are means over updates 191--200.}
 \label{tab:grpo-variants-results}
 \begin{tabularx}{\linewidth}{@{}Yrrrrrr@{}}
 \toprule
 & \multicolumn{3}{c}{\textbf{Validation (\%)}} & \textbf{Training} & & \textbf{Length}\\
 \cmidrule(lr){2-4}
 \textbf{Experiment} & \textbf{Update 0} & \textbf{Best (update)} & \textbf{Update 200} & \textbf{reward} & \textbf{Entropy} & \textbf{(tokens)}\\
 \midrule
 Baseline & 40.11 & \textbf{50.00} (150) & 48.02 & 0.783 & 0.044 & 2,729\\
 Downsample & 34.46 & 49.15 (120) & \textbf{48.87} & 0.701 & 0.057 & 2,129\\
 Replay & 37.29 & 49.72 (150) & 42.66 & 0.719 & 0.058 & 3,118\\
 Virtual Reward & 33.90 & 47.18 (80) & 44.35 & 0.775 & 0.132 & 6,247\\
 \bottomrule
 \end{tabularx}
\end{table}

As summarized in Table~\ref{tab:grpo-variants-results}, each intervention produces its intended training-side shift: Downsample directs 90.9\% of active rollout slots to non-easy prompts and achieves the highest final-step accuracy (48.87\%); Replay lifts autonomous rollout success on hard training prompts in the final quarter of training from 4.4\% to 37.8\%; and Virtual Reward accelerates early training reward and validation gains around updates 41--90 before inducing late-stage entropy and length inflation ($2.3\times$ Baseline length). Nevertheless, none of the three variants surpasses standard group-mean \textbf{Baseline}'s peak validation solve rate (\textbf{50.00\%}), as localized training-prompt gains, uncorrected off-policy replay ratios, and unannealed negative baselines do not translate cleanly into held-out greedy accuracy (note also that Update-0 greedy scores span $33.90\%$--$40.11\%$ across identical initial weights due to non-deterministic parallel reductions and discrete Top-6 MoE routing boundary flips in distributed rollout workers prior to the alignment repairs in Section~\ref{sec:train-infer-consistency}). Moreover, while Base-initialized Chat-GRPO \textbf{Baseline} reaches a slightly higher greedy score on this in-distribution seed panel ($50.00\%$ peak, $48.02\%$ final) than SFT-initialized Chat200 ($46.61\%$) because it sharpens Base's native seed formatting without off-policy distribution shift, cold-start SFT proves critical for expanding multi-sample structural coverage on held-out \textsc{SLAI-OR-Bench-Hard} once paired with sequential Chat/Thinking GRPO (Section~\ref{sec:evaluation-sampling-migration}). We therefore retain standard group-mean advantage centering paired with Failure Reward Tiers and Outcome Masking in our main SFT+RL recipe, and provide full mathematical formulations, mechanism diagrams, learning curves, and diagnostic analyses in Appendix~\ref{app:grpo-variants}.

\subsection{Empirical Gains and Scenario-Level Capability Transitions}
\label{sec:post-training-evaluation}

\subsubsection{Single-Turn Benchmark Accuracy}\label{sec:evaluation-scores}

We compare DeepSeek-V4-Flash-0731 (Base), our post-trained \model (SFT+RL) model, and two frontier closed-source baselines, Claude Opus~5.5 and GPT-6.1-Sol (Gemini-4-Argon was evaluated only in the initial public-benchmark saturation survey in Table~\ref{tab:frontier-benchmark-diagnostic}). Table~\ref{tab:benchmark-accuracy} reports single-turn solve accuracy on the full \textsc{SLAI-OR-Bench} collection (131 tasks) and \textsc{SLAI-OR-Bench-Hard} (93 tasks).

\begin{table}[!htbp]
\caption{Single-turn solver-based accuracy on \textsc{SLAI-OR-Bench} (\%). Base and \model (SFT+RL) report mean accuracy across eight independent evaluation runs; closed-source baselines reflect recorded single-turn benchmark evaluations.}\label{tab:benchmark-accuracy}
\centering\small
\setlength{\tabcolsep}{6.5pt}
\renewcommand{\arraystretch}{1.2}
\begin{tabularx}{\linewidth}{@{}L{0.30\linewidth}*{4}{>{\centering\arraybackslash}X}@{}}
\toprule
\textbf{Benchmark} & \textbf{Base} & \textcolor{slai}{\makecell{\textbf{\model}\\\textbf{(SFT+RL)}}} & \makecell{\textbf{Claude Opus}\\\textbf{5.5}} & \textbf{GPT-6.1-Sol} \\
\midrule
\textsc{SLAI-OR-Bench} (131) & $38.26$ & $\mathbf{50.86}$ & $44.27$ & $45.04$ \\
\textsc{SLAI-OR-Bench-Hard} (93) & $26.88$ & $\mathbf{35.89}$ & $33.33$ & $34.41$ \\
\bottomrule
\end{tabularx}
\end{table}

Post-training in \model raises accuracy from 38.26\% to 50.86\% on the full 131-task benchmark ($+12.60$ percentage points over Base, and $+5.82$ and $+6.59$ points ahead of GPT-6.1-Sol and Claude Opus~5.5) and from 26.88\% to 35.89\% on the 93-task hard subset ($+9.01$ points over Base, slightly surpassing GPT-6.1-Sol at 34.41\% and Claude Opus~5.5 at 33.33\% under single-draw evaluation).

\subsubsection{Multi-Sample Coverage and Reliability Migration}\label{sec:evaluation-sampling-migration}

\begin{figure}[!htbp]
\centering
\includegraphics[width=0.95\linewidth]{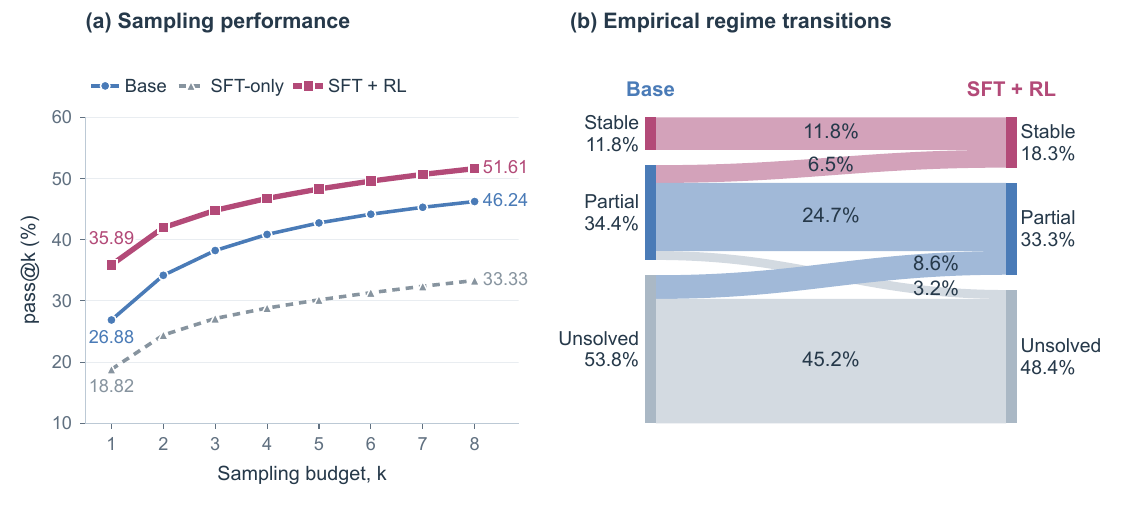}
\caption{Sampling performance along Base $\rightarrow$ cold-start SFT $\rightarrow$ GRPO on \textsc{SLAI-OR-Bench-Hard}. (a) Base, SFT-only, and SFT+RL checkpoints; SFT+RL leads for every evaluated $k$. (b) Paired Base-to-SFT+RL transitions among Stable (8/8), Partial (1--7/8), and Unsolved (0/8), using eight responses per problem.}\label{fig:sampling-and-migration}
\end{figure}

Figure~\ref{fig:sampling-and-migration}(a) tracks multi-sample performance across the three checkpoints of the Base $\rightarrow$ cold-start SFT $\rightarrow$ GRPO recipe: Base, SFT-only, and the final SFT+RL model. Base and SFT+RL use eight per-question draws (supporting paired reliability transitions in Panel~(b)), whereas SFT-only is summarized by aggregate pass@$k$ statistics; Figure~\ref{fig:pt-initialization-comparison}(a) provides the companion comparison with RL-only.

The final SFT+RL checkpoint rises from 35.89\% at pass@1 to 51.61\% at pass@8, outperforming Base by 5.38--9.01 percentage points and SFT-only by 17.07--18.28 points across $k=1,\ldots,8$. Notably, 25 updates of standalone cold-start SFT temporarily lower Hard-set pass@1 from 26.88\% to 18.82\% and pass@8 from 46.24\% to 33.33\%, yet initializing two-stage GRPO from this SFT checkpoint achieves higher multi-sample coverage than our standalone RL-only run from Base (51.61\% versus 44.09\% at pass@8; Figure~\ref{fig:pt-initialization-comparison}(a)). While RL-only follows a separate single-run schedule rather than a step-matched two-stage ablation, this ``dip-then-lift'' trajectory along the main pipeline reflects complementary stage dynamics: brief off-policy SFT on dual-interface demonstrations---nearly 29\% of which originate from corrected Base $\mathrm{acc}=0$ tasks---perturbs the base model's default output distribution and narrows raw sampling diversity prior to online adaptation, depressing immediate zero-shot pass@$k$, while seeding explicit formulation structures that subsequent solver-verified Chat and Thinking GRPO recalibrate and amplify.

In the paired Base-to-SFT+RL analysis, all initially Stable problems remain Stable. Six problems move from Partial to Stable (6.45\% of Hard), eight move from Unsolved to Partial (8.60\%), and three regress from Partial to Unsolved (3.23\%), with no direct Unsolved-to-Stable jump, yielding a net solve-coverage gain of $+5.38$ percentage points (Figure~\ref{fig:sampling-and-migration}(b)).

\subsubsection{Scenario and Pitfall Analysis of Reliability Transitions}\label{sec:evaluation-scenario-transfer}

\begin{figure}[!htbp]
\centering
\includegraphics[width=0.95\linewidth]{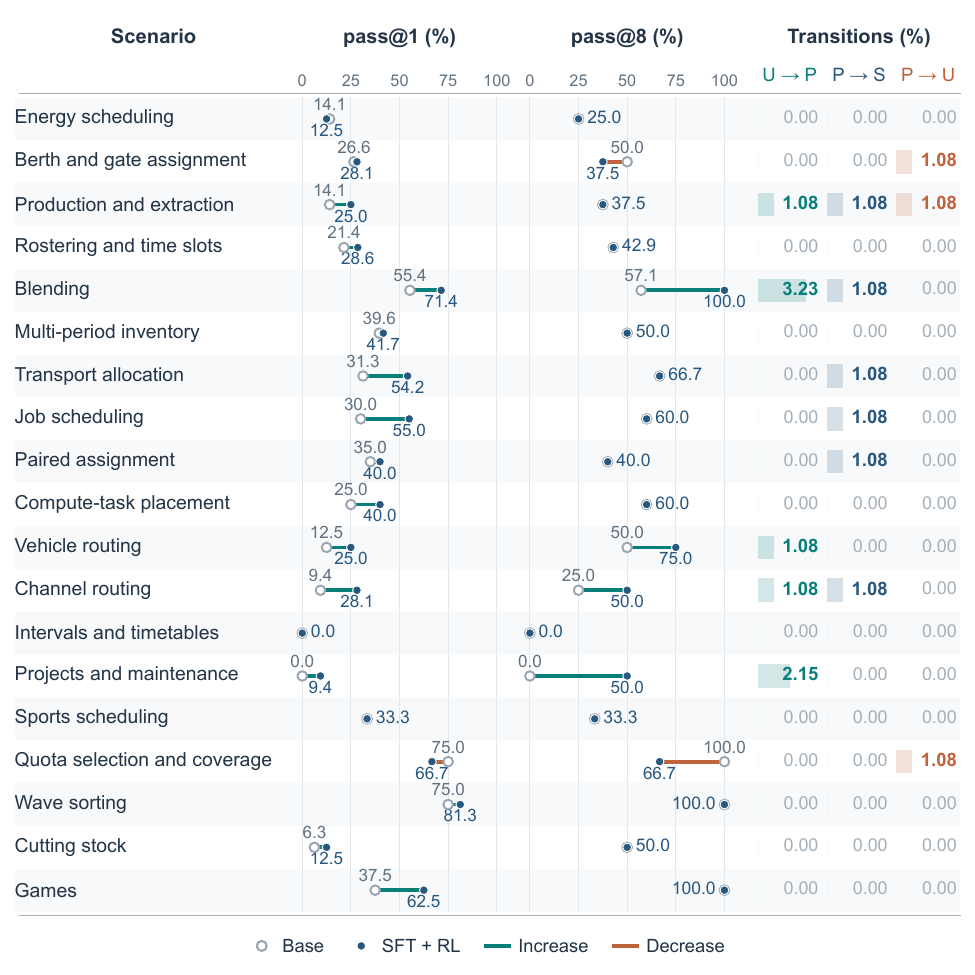}
\caption{Paired scenario-level performance and reliability transitions from Base to SFT+RL across the 19 synthesis scenarios represented in \textsc{SLAI-OR-Bench-Hard}.}\label{fig:scenario-capability-transfer}
\end{figure}

Figure~\ref{fig:scenario-capability-transfer} decomposes these paired transitions across the 19 synthesis scenarios represented in \textsc{SLAI-OR-Bench-Hard}. Open and filled points denote Base and SFT+RL within-scenario pass rates, while right-hand columns report transition shares normalized over all 93 Hard instances.

Blending contributes 3.23\% of all Hard problems to Unsolved-to-Partial migration, projects and maintenance contribute 2.15\%, and vehicle routing, channel routing, and production and extraction each contribute 1.08\%. Their associated pitfalls involve aggregate product-quality constraints; resource eligibility, concurrency, precedence, and per-period budgets; waiting actions and space--time conflicts; route-wide channel consistency and shared capacity; and aggregate energy and environmental limits. These transitions connect our synthesis taxonomy directly to formulation requirements that post-training unlocks, while the absence of direct Unsolved-to-Stable jumps underscores the difficulty of mastering newly acquired structures in a single training cycle.

Partial-to-Stable consolidation occurs in blending, production and extraction, transport allocation, job scheduling, paired assignment, and channel routing; in transport allocation and job scheduling, higher pass@1 at unchanged pass@8 reflects sharpened single-attempt reliability within existing coverage. Localized regressions remain in berth and gate assignment, quota selection/coverage, and one production-and-extraction instance, indicating that capability gains across coupled combinatorial constraints remain uneven. Scaling this solver-in-the-loop GRPO pipeline reliably and cost-effectively across large MoE models motivates the Ascend systems and algorithmic optimizations examined next in Section~\ref{sec:or-rl-infrastructure}.

\section{Systems and Algorithmic Infrastructure for Execution-in-the-Loop RL}
\markboth{\thesection.\ Infrastructure for Execution-in-the-Loop RL}{\thesection.\ Infrastructure for Execution-in-the-Loop RL}
\label{sec:or-rl-infrastructure}

Training \model with reinforcement learning (RL) requires coupling long-horizon reasoning and external solver or profiler execution with full-parameter updates of a large Mixture-of-Experts (MoE) model~\citep{deepseekr1,optimus,deepseekv3}. Building on our foundational training stack for the DeepSeek-V4 family on Huawei CloudMatrix384 SuperPOD~\citep{zuo2025cloudmatrix384,slai2026trexpart1}---whose compute nodes equip Huawei Ascend~910C / Ascend~A3 (\texttt{ascend910\_93}) dual-die packages, so an eight-card node exposes either 8 physical cards or 16 logical NPU dies depending on the parallelism layout---this section presents four controlled systems and algorithmic studies across the execution-in-the-loop RL pipeline. Rather than retroactively modifying the frozen post-training run of Section~\ref{sec:or-bench-post-training}, each subsection isolates a specific bottleneck against a matched baseline under its corresponding workload and checkpoint lineage to establish validated components for next-generation unified training:

\begin{enumerate}[leftmargin=2em, itemsep=2pt, topsep=3pt]
  \item \textbf{High Rollout and Verification Cost (Section~\ref{sec:v4-reuse}):} Solving difficult OR problems requires generating multi-thousand-token reasoning traces and executing external solvers, while uniform-outcome groups under GRPO~\citep{shao2024deepseekmath,dapo} yield zero task-reward gradients. Discarding verified trajectories after a single gradient step leaves substantial supervision unused, motivating bounded same-batch trajectory reuse with recalibrated advantages.
  \item \textbf{Training--Inference Numerical Mismatch (Section~\ref{sec:train-infer-consistency}):} Decoupled RL pipelines~\citep{hybridflow} generate rollouts in an inference engine (vLLM-Ascend) and compute gradients in a training engine (MindSpeed-LLM/Megatron). Differences in parallel reduction orders, precision formats, and NPU kernel implementations cause the two engines to assign different token probabilities under identical weights~\citep{precisionrl,zhong2026diagnosing}, flipping discrete MoE routing decisions and destabilizing optimization.
  \item \textbf{Autoregressive Decoding Latency (Section~\ref{sdinsert:sec:speculative-decoding}):} Long OR reasoning and code traces make decoding throughput a central deployment and rollout concern. We characterize workload-dependent draft acceptance in native multi-token prediction (MTP), study auxiliary-gradient routing during joint OR--MTP RL, and refine the private MTP module with the selected OR target frozen, auditing both offline serving throughput and solver-verified accuracy.
  \item \textbf{Long-Tail Latency in NPU Kernel Optimization (Section~\ref{sec:kernel-perf-opt}):} Optimizing underlying Ascend NPU operators~\citep{flashattention} via agentic RL requires repeated code edits, compilation, correctness verification, and hardware profiling~\citep{wu2026ascendoptimizer,dai2026cudaagent}. Waiting for multi-hour episodes to finish causes severe learner starvation, motivating solver-guided parameter warm starts and asynchronous candidate-prefix RL with shared experimental memory.
\end{enumerate}

% Include this file under the infrastructure section as a subsection. Required packages:
% amsmath, amssymb, booktabs, array, graphicx, natbib, placeins.
% Place this file in sections/; keep figures/ at the report root.
% Keep the existing custom.bib and citation keys; no new entries are required.
% Revised 2026-10-30 from the supplied infrastructure-subsection manuscript.
% Frozen update-80 values, embedded to avoid an external data/results dependency.
\def\VFourBaseRounds{80}
\def\VFourBaseTrajectories{20,480}
\def\VFourBaseTokensM{57.55}
\def\VFourBaseUses{10,520}
\def\VFourBaseSolved{21}
\def\VFourBaseAcc{22.58}
\def\VFourBaseMeanLength{2,810}
\def\VFourReuseRounds{20}
\def\VFourReuseTrajectories{5,120}
\def\VFourReuseTokensM{10.08}
\def\VFourReuseUses{9,838}
\def\VFourReuseSolved{24}
\def\VFourReuseAcc{25.81}
\def\VFourReuseMeanLength{1,968}
\def\VFourTokenSaving{82.5}

\subsection{Advantage Recalibration for Same-Batch Trajectory Reuse}
\label{sec:v4-reuse}

\begin{figure}[!htbp]
  \centering
  \includegraphics[width=0.84\linewidth]{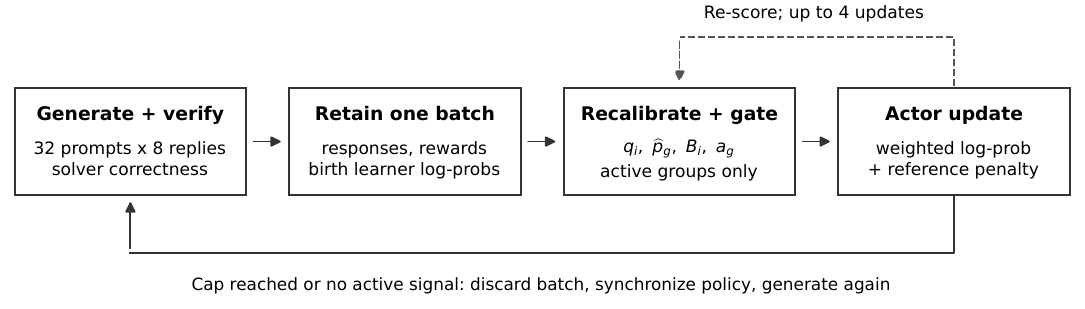}
  \caption{Workflow of bounded same-batch trajectory reuse ($R_{\max}=4$). Stored responses and their birth-policy log-probabilities are retained across up to four updates; before each reuse step, the current learner re-scores active groups to recalibrate advantages and evaluate group eligibility gates without re-running generation or the solver.}
  \label{fig:v4-reuse-workflow}
\end{figure}

Improving the throughput of rollout generation addresses only one part of the
cost of reinforcement learning (RL). In operations research (OR), each new
response can require a long reasoning trace, solver-facing code, and execution
to obtain a correctness signal. Discarding this response after a single update
may leave useful supervision unexploited. Prior work such as ReMix and ExGRPO
has explored off-policy data reuse and experience management for language-model
RL \citep{v4_liang2025remix,v4_zhan2026exgrpo}. We focus on extracting the
remaining learning value of already generated and verified trajectories, while
avoiding repeated optimization against an outdated advantage. Our design couples
current-reference advantage recalibration, a matching weighted actor objective,
and group-level reuse gates. It is implemented as a bounded same-batch loop in
DeepSeek-V4-Flash training (Figure~\ref{fig:v4-reuse-workflow}), without a persistent
replay buffer or additional rollout-based calibration stage.

\subsubsection{Reinterpreting Historical Outcomes under the Current Policy}
\label{sec:v4-reference}

Let $x_g$ be a prompt, $y_i$ a stored response, and $r_i\in\{0,1\}$ its
solver-based correctness reward. Although a response and its verified outcome
are fixed, its advantage depends on the policy used as the comparison reference.
For a sampling policy $\pi_u^{s}$ after $u$ optimizer updates, the conceptual
reference and unstandardized advantage are
\begin{equation}
 V_u(x_g)=\mathbb{E}_{Y\sim\pi_u^{s}(\cdot\mid x_g)}[r(x_g,Y)],
 \qquad A^*_{i,u}=r_i-V_u(x_g).
 \label{eq:v4-population}
\end{equation}
Consequently, $A^*_{i,u}-A^*_{i,v}=-[V_u(x_g)-V_v(x_g)]$. A successful answer
becomes less exceptional as the policy improves on its prompt. An unsuccessful
answer can instead acquire a more negative advantage. Thus, neither reuse count
nor a universal decay factor describes the evolution of every training signal.
The reference must be allowed to move in either direction.

We estimate a current-relative reference using the responses already available
for each prompt. Let $I_g$ contain the responses with valid training masks,
$G_g=|I_g|$, $m_{i,t}$ be the fixed birth token mask, and
$L_i=\sum_t m_{i,t}$. At the beginning of each reuse update, a frozen snapshot
$\bar\pi_u$ of the current learner scores the stored responses. Relative to
the learner snapshot $\pi_b$ taken when the batch was generated, define
\begin{align}
 z_{i,u}&=\frac{1}{L_i}\sum_t m_{i,t}
 \left[\log\bar\pi_u(y_{i,t}\mid h_{i,t})
       -\log\pi_b(y_{i,t}\mid h_{i,t})\right], \nonumber\\
 q_{i,u}&=\frac{\exp(z_{i,u}/\tau)}{\sum_{j\in I_g}\exp(z_{j,u}/\tau)},
 \qquad \widehat p_{g,u}=\sum_{i\in I_g}q_{i,u}r_i .
 \label{eq:v4-weights}
\end{align}
Here $h_{i,t}=(x_g,y_{i,<t})$ is the token context and $\tau=0.03$ controls
the sensitivity of relative weights. Because both $\bar\pi_u$ and $\pi_b$ are evaluated on the learner backend,
any numerical discrepancy between the rollout and training engines is cancelled out of $z_{i,u}$.
The tempered, length-normalized weights $q_{i,u}$ capture relative likelihood shifts over the observed candidate set $I_g$ rather than full-trajectory importance sampling ratios~\citep[Chapter~9]{v4_owen2013mc}, and $\widehat p_{g,u}$ serves as a finite-support proxy for the current expected reward on prompt $x_g$.

We retain the birth-group sample standard deviation
$\sigma_{g,0}=[\sum_{i\in I_g}(r_i-\bar r_g)^2/(G_g-1)]^{1/2}$,
where $\bar r_g=G_g^{-1}\sum_{i\in I_g}r_i$, and form
\begin{equation}
 \widehat A_{i,u}=\frac{r_i-\widehat p_{g,u}}{\sigma_{g,0}+\epsilon},
 \qquad B_{i,u}=G_g q_{i,u}\widehat A_{i,u},
 \qquad \epsilon=10^{-6}.
 \label{eq:v4-coefficient}
\end{equation}
Using the same weights $q_{i,u}$ for both reference estimation and the actor coefficient ensures zero-sum group weights,
$\sum_{i\in I_g} B_{i,u}=0$. At the initial update ($u=0$), $z_{i,0}=0$ and $B_{i,0}$ reduces to the standard GRPO group-normalized advantage~\citep{shao2024deepseekmath}.
Freezing the denominator at the birth scale $\sigma_{g,0}$ prevents residual noise from being amplified as within-group reward variance shrinks across reuse steps~\citep{v4_liu2025r1zero}.
For binary rewards $r_i\in\{0,1\}$, the mean absolute actor coefficient satisfies
\begin{equation}
 \frac{1}{G_g}\sum_{i\in I_g}|B_{i,u}|
 =\frac{2\widehat p_{g,u}(1-\widehat p_{g,u})}{\sigma_{g,0}+\epsilon},
 \label{eq:v4-contrast}
\end{equation}
which smoothly diminishes as the candidate probability mass concentrates on either all-pass ($\widehat p_{g,u}\to 1$) or all-fail ($\widehat p_{g,u}\to 0$) outcomes.

\subsubsection{A Matching Actor Objective and Bounded Reuse}
\label{sec:v4-objective}

Let $a_{g,u}\in\{0,1\}$ indicate whether group $g$ is eligible for the next
update, and let $N$ be the nominal number of response slots in the batch. We
optimize the sequence-mean, token-mean objective
\begin{equation}
 \mathcal{L}_{\mathrm{reuse}}(\theta;u)
 =-\frac{1}{N}\sum_g\sum_{i\in I_g}
 a_{g,u}\,\operatorname{sg}(B_{i,u})
 \frac{1}{L_i}\sum_t m_{i,t}\log\pi_\theta(y_{i,t}\mid h_{i,t})
 +\beta\mathcal{K}_{\mathrm{ref}}(\theta;u),
 \label{eq:v4-loss}
\end{equation}
where $\operatorname{sg}$ stops gradients through the estimated coefficients and
$\beta=10^{-3}$. The reference-policy penalty uses the same group and token
masks and the same nominal denominator. Specifically,
\begin{equation}
 \mathcal{K}_{\mathrm{ref}}
 =\frac{1}{N}\sum_g\sum_{i\in I_g}\frac{a_{g,u}}{L_i}
 \sum_t m_{i,t}\kappa(d_{i,t}),\qquad
 d_{i,t}=\log\pi_{\mathrm{ref}}(y_{i,t}\mid h_{i,t})
          -\log\pi_\theta(y_{i,t}\mid h_{i,t}),
 \label{eq:v4-kl}
\end{equation}
with $\pi_{\mathrm{ref}}$ denoting the frozen initial reference policy. We use the clamped Schulman low-variance KL estimator~\citep{v4_schulman2020kl,shao2024deepseekmath},
$\kappa(d)=\operatorname{clip}(e^{\tilde d}-\tilde d-1,-10,10)$ with
$\tilde d=\operatorname{clip}(d,-20,20)$, evaluated over retained responses.

The reward term in Equation~\eqref{eq:v4-loss} directly optimizes weighted
log-probabilities without a token-level PPO clipping ratio~\citep{v4_schulman2017ppo}.
With the batch and gates fixed, its gradient evaluated at $\theta=\bar\theta_u$ equals
$-\tau N^{-1}\sum_g a_{g,u}G_g
(\sigma_{g,0}+\epsilon)^{-1}\nabla_\theta\widehat p_{g,u}$,
connecting the detached coefficient $B_{i,u}$ to local ascent on the candidate-weighted reward proxy $\widehat p_{g,u}$ while retaining gradient-norm clipping at $1.0$ and the reference KL penalty.

Group eligibility gates decouple advantage estimation from the decision to continue updating on stored responses. Alongside $\widehat p_{g,u}$, the gate monitors common likelihood drift $\delta_{g,u}=|G_g^{-1}\sum_{i\in I_g}z_{i,u}|$ and the normalized effective sample size (ESS) diagnostic~\citep{v4_martino2017ess} computed from untempered weights $\widetilde q_{i,u}$:
\begin{equation}
 \widetilde q_{i,u}=\frac{e^{z_{i,u}}}{\sum_{j\in I_g}e^{z_{j,u}}},
 \qquad e_{g,u}=\frac{1}{G_g\sum_{i\in I_g}\widetilde q_{i,u}^{\,2}}.
 \label{eq:v4-gate-stats}
\end{equation}
Here $e_{g,u}\in[1/G_g, 1]$ measures weight concentration across the group, while $\delta_{g,u}$ detects uniform likelihood shifts that within-group softmax normalization would otherwise cancel out. Table~\ref{tab:v4-gates} lists the gating rules.

\begin{table}[!htbp]
\centering\small
\caption{Group eligibility rules, evaluated in the listed order. Any \emph{Remove}, \emph{Suspend}, or \emph{Retire} action sets $a_{g,u}=0$ for the remainder of the current four-step batch without blacklisting future generations of the same prompt (labels distinguish structural filtering, drift/low-yield suspension, and high-yield saturation).}
\label{tab:v4-gates}
\begin{tabular}{@{}p{0.38\linewidth}p{0.56\linewidth}@{}}
\toprule
Condition & Action and interpretation \\
\midrule
$G_g<2$ & Remove: insufficient valid responses for a comparison. \\
All valid rewards are identical & Remove: no within-group reward contrast. \\
$\delta_{g,u}>0.03$ or $e_{g,u}<0.5$ & Suspend: the historical support is unreliable. \\
$\widehat p_{g,u}\geq0.95$ on two consecutive checks & Retire: predicted high-success saturation in the candidate set. \\
$\widehat p_{g,u}\leq0.05$ on two consecutive checks & Suspend: predicted low-success saturation, not mastery. \\
Otherwise & Train using the recalibrated coefficients. \\
\bottomrule
\end{tabular}
\end{table}

Each generated batch supports at most $R_{\max}=4$ optimizer updates. Before
each additional update, the learner re-scores the original responses, recomputes
the coefficients, and applies the gate. Removed groups contribute neither reward
loss nor reference penalty; their slots are not refilled and the denominator
$N$ does not shrink. If every reward coefficient is zero, the loop stops without
an optimizer step. After termination, the batch is discarded, the rollout policy
is synchronized, and a new batch is generated. No repeated solver calls, targeted
refresh queue, or online calibration rollouts are required within this loop.
Re-scoring and additional optimization still incur compute costs.

\subsubsection{DeepSeek-V4-Flash OR Training Setup}
\label{sec:v4-setup}

To evaluate generation efficiency under a shared initialization and
correctness-only rewards, we compare a fresh-only baseline with calibrated
dynamic reuse starting directly from the same
DeepSeek-V4-Flash-0731 BF16 initialization on a veRL-based full-parameter
training stack \citep{hybridflow} across
$8\times16=128$ Ascend NPUs. The OR training file contains 557 prompts, and the
fixed \texttt{or\_bench\_hard93} (\textsc{SLAI-OR-Bench-Hard}) evaluation file contains 93 problems. There are no exact duplicate prompt
records across these two files. Each rollout round samples 32 prompts and
generates eight responses per prompt, giving $N=256$ trajectories. Training
sampling uses temperature $1.0$ and top-$p$ $1.0$; evaluation uses one greedy
response per problem. Prompt and response limits are 4,096 and 12,288 tokens.
Both runs use seed 42, learning rate $10^{-6}$, Adam with $(0.9,0.999)$ momentum
parameters, weight decay $0.01$, and the reference penalty above.

The fresh-only baseline uses GRPO-style group-standardized correctness advantages
\citep{shao2024deepseekmath} and a PPO-style clipped actor
\citep{v4_schulman2017ppo}, configured with ratio range $[0.8,1.2]$ and
dual-clip factor $3$; each batch
supports at most one optimizer update. Whereas naive multi-epoch PPO reuses static birth-group advantages against shifting policy likelihoods, our method jointly introduces learner-side reference recalibration ($q_{i,u}$), the matching weighted log-probability objective, and ESS/drift group eligibility gates across up to $R_{\max}=4$ updates per batch. This controlled comparison evaluates the integrated reuse design against standard single-update fresh-rollout GRPO under identical binary solver rewards (without length shaping); responses lacking a usable verifier outcome are masked from gradient updates while remaining counted in generation cost.

\subsubsection{Generation Efficiency under Matched Training Updates}
\label{sec:v4-results}

We compare the fresh-only baseline and calibrated reuse at a matched budget of
80 actual optimizer updates, corresponding to 80 fresh rollout rounds ($80 \times 1$) for the baseline and 20 rollout rounds ($20 \times 4$) for calibrated reuse, evaluated on the 93-problem \textsc{SLAI-OR-Bench-Hard} set.

\begin{figure}[!htbp]
  \centering
  \includegraphics[width=0.92\linewidth]{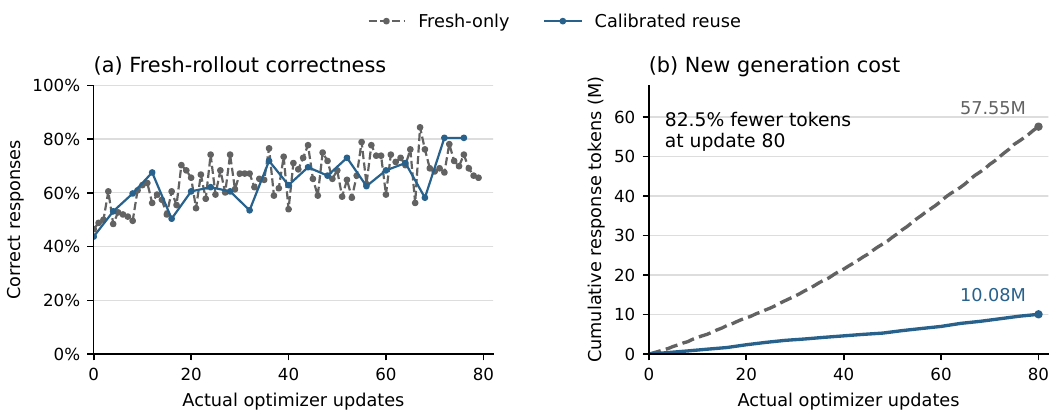}
  \caption{Training quality and generation cost under a matched budget of 80
  actual optimizer updates. (a) Unsmoothed correctness on fresh training
  rollouts, located at the number of updates completed before generation
  (80 observed batches for fresh-only training and 20 for calibrated reuse).
  (b) Cumulative new training-response tokens, counted at completed batch-update
  boundaries. Lines connect recorded observations; fixed-set evaluation is
  reported separately in Table~\ref{tab:v4-main}.}
  \label{fig:v4-quality-cost}
\end{figure}

\begin{table}[!htbp]
\centering\small
\renewcommand{\arraystretch}{1.12}
\caption{Generation cost and fixed-set evaluation at 80 actual optimizer updates.}
\label{tab:v4-main}
\begin{tabular*}{\linewidth}{@{\extracolsep{\fill}}lrr@{}}
\toprule
\textbf{Metric} & \textbf{Fresh-only} & \textbf{Calibrated reuse} \\
\midrule
Actual optimizer updates & 80 & 80 \\
Fresh rollout rounds & \VFourBaseRounds & \VFourReuseRounds \\
New trajectories & \VFourBaseTrajectories & \VFourReuseTrajectories \\
New response tokens (million) & \VFourBaseTokensM & \VFourReuseTokensM \\
Solved evaluation problems & \VFourBaseSolved/93 & \VFourReuseSolved/93 \\
Evaluation accuracy (\%) & \VFourBaseAcc & \VFourReuseAcc \\
\bottomrule
\end{tabular*}
\par\smallskip
\begin{minipage}{\linewidth}
\footnotesize\raggedright
Both runs initialize from Base (seed 42) and use single-draw greedy evaluation on \textsc{SLAI-OR-Bench-Hard} (retaining unscored attempts in the 93-problem denominator). Generation cost counts all sampled training responses, excluding prompt prefill, learner re-scoring forward passes, and backward updates.
\end{minipage}
\end{table}

As shown in Figure~\ref{fig:v4-quality-cost} and Table~\ref{tab:v4-main}, under
the same 80-update budget, calibrated reuse cuts newly generated
trajectories (and external solver verification calls) by \textbf{75.0\%}, from \VFourBaseTrajectories{} to
\VFourReuseTrajectories{}, and reduces newly generated response tokens by
\textbf{\VFourTokenSaving\%}, from \VFourBaseTokensM\,M to
\VFourReuseTokensM\,M. On \textsc{SLAI-OR-Bench-Hard}, single-draw greedy evaluation solves
\VFourReuseSolved/93 problems (\VFourReuseAcc\%) versus
\VFourBaseSolved/93 (\VFourBaseAcc\%) for fresh-only training, confirming that aggressive rollout reduction preserves early-stage learning progress at matched update counts (note that this standalone Chat-mode protocol uses a stricter 12,288-token cap and single greedy draw, so scores compare the two matched 80-update runs rather than the eight-run 16K/64K evaluation in Section~\ref{sec:or-bench-post-training}).

All 20 reuse batches complete four updates, yielding \VFourReuseUses{} nonzero-coefficient trajectory uses after group eligibility gating and zero-contrast filtering (versus \VFourBaseUses{} for fresh-only). The \VFourTokenSaving\% token saving reflects both the $4\times$ reduction in sampled trajectories and a shorter observed mean response length (\VFourReuseMeanLength{} versus \VFourBaseMeanLength{} tokens). Although each reuse step incurs an extra forward scoring pass on the learner backend, learner prefill over stored tokens is parallelizable across sequence positions and avoids both sequential autoregressive decoding and external MIP solver execution---the two dominant latency contributors in OR RL rollouts; benchmarking end-to-end wall-clock step times and uncalibrated multi-epoch PPO baselines over longer horizons is left to future systems evaluation.

\subsection{Progressive Diagnosis and Alignment of Training--Inference Drift}
\label{sec:train-infer-consistency}

Large language model reinforcement learning remains difficult to stabilize in practice: policies can deteriorate rapidly, leading to declining rewards or training collapse.
Diagnosing these failures is challenging because optimization settings, reward design, and numerical differences in the underlying system can interact.
In particular, differences in operator implementations, numerical precision, and floating-point reduction order can cause training and inference engines to assign different token probabilities even under identical model weights and inputs, a phenomenon known as training--inference mismatch (TIM)~\citep{he2025nondeterminism,bai2026bitwise}.
Controlled experiments have shown that such numerical discrepancies can independently trigger training collapse, making their effects important to identify when investigating training stability~\citep{zhong2026diagnosing}.

In our system, vLLM-Ascend generates rollouts, while MindSpeed-LLM/Megatron performs policy updates.
Debugging, rollbacks, and reruns following training failures increase the overall cost of effective training~\citep{bai2026bitwise}.
For operations research tasks that require extended reasoning and solver verification, these additional costs can further constrain the efficiency of capability exploration.
However, final outputs and training curves alone are insufficient to identify the computational source of a discrepancy or determine an appropriate fix.
A diagnostic workflow is therefore needed to trace where discrepancies originate, analyze how they propagate and amplify, and verify the effects of corrective measures, supporting stable training on Ascend SuperPOD.

To address this need, we conducted a series of training--inference consistency analyses and repairs, and distilled the resulting experience into a diagnostic workflow based on progressive localization and controlled validation.
The workflow first compares intermediate states across model layers to identify where discrepancies increase substantially.
It then uses same-input replay to progressively inspect internal modules and operators, tracing the sources of these differences.
Confirmed implementation errors and precision-configuration issues are repaired, followed by renewed full-model comparisons.
For amplified numerical perturbations, controlled experiments test the proposed explanations.
By connecting discrepancy detection, causal investigation, and result verification, this workflow provides a repeatable procedure for diagnosing training--inference inconsistency in complex MoE models.
To facilitate reuse, we further package the workflow as a Skill that guides coding agents through the diagnostic steps.

\subsubsection{Layer-Wise Cosine Scanning and Recursive Module Replay}

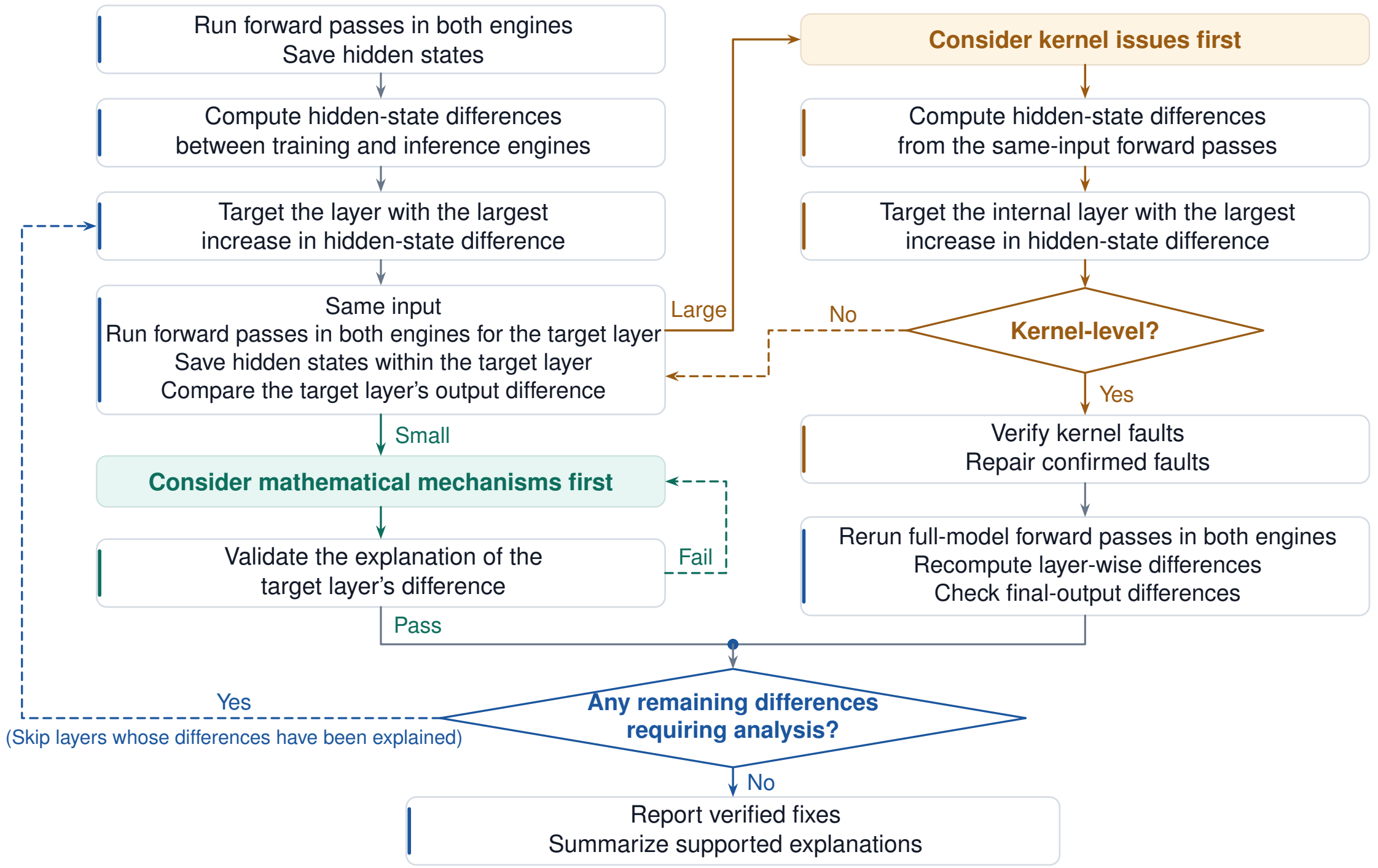
\begin{figure}[!htbp]
  \centering
  \resizebox{\linewidth}{!}{% Progressive diagnosis and alignment workflow for training--inference consistency (Section 4.2).
% Native size: 159 mm x 102.2 mm (~1.0 \linewidth, so \resizebox{\linewidth}{!}{...} scales ~1:1).
% Preserves 100% of the wording, visual hierarchy, color palette, and control-flow topology
% of figures/train_infer_consistency_flow_v2.pdf with full-size readable typography.
\begingroup
\definecolor{tiBlue}{HTML}{1B569E}
\definecolor{tiAmber}{HTML}{9A5B13}
\definecolor{tiAmberBg}{HTML}{FDF4E3}
\definecolor{tiTeal}{HTML}{0E6E5E}
\definecolor{tiTealBg}{HTML}{E7F6F2}
\definecolor{tiBorder}{HTML}{D5DCE4}
\definecolor{tiSlate}{HTML}{687789}
\definecolor{tiInk}{HTML}{111827}
\begin{tikzpicture}[
  x=1mm,y=1mm,
  line cap=round,line join=round,
  font=\sffamily\fontsize{8.0}{9.8}\selectfont,
  arrSlate/.style={draw=tiSlate,line width=0.65pt,-{Stealth[length=1.6mm,width=1.15mm]}},
  arrBlue/.style={draw=tiBlue,line width=0.68pt,-{Stealth[length=1.6mm,width=1.15mm]}},
  arrBlueDash/.style={draw=tiBlue,line width=0.68pt,dash pattern=on 2.6pt off 1.9pt,-{Stealth[length=1.6mm,width=1.15mm]}},
  arrAmber/.style={draw=tiAmber,line width=0.72pt,-{Stealth[length=1.6mm,width=1.15mm]}},
  arrAmberDash/.style={draw=tiAmber,line width=0.68pt,dash pattern=on 2.6pt off 1.9pt,-{Stealth[length=1.6mm,width=1.15mm]}},
  arrTeal/.style={draw=tiTeal,line width=0.68pt,-{Stealth[length=1.6mm,width=1.15mm]}},
  arrTealDash/.style={draw=tiTeal,line width=0.68pt,dash pattern=on 2.6pt off 1.9pt,-{Stealth[length=1.6mm,width=1.15mm]}},
  lblBlue/.style={text=tiBlue,font=\sffamily\fontsize{7.4}{8.8}\selectfont,inner sep=1pt},
  lblAmber/.style={text=tiAmber,font=\sffamily\fontsize{7.4}{8.8}\selectfont,inner sep=1pt},
  lblTeal/.style={text=tiTeal,font=\sffamily\fontsize{7.4}{8.8}\selectfont,inner sep=1pt}
]
\path[use as bounding box] (0.5,-3.3) rectangle (160.3,99.0);

% Helper macro idea implemented directly for crisp left accent bars on rounded white cards:
% Left column: x in [10, 77] (width 67mm, center 43.5)
% Right column: x in [93, 160] (width 67mm, center 126.5)

% L1: Run forward passes in both engines / Save hidden states
\path[draw=tiBorder,fill=white,line width=0.55pt,rounded corners=1.2mm] (10,90.5) rectangle (77,98.5);
\draw[draw=tiBlue,line width=1.15pt] (10.35,91.9) -- (10.35,97.1);
\node[text=tiInk,align=center] at (43.8,94.5) {Run forward passes in both engines\\Save hidden states};

% L2: Compute hidden-state differences between training and inference engines
\path[draw=tiBorder,fill=white,line width=0.55pt,rounded corners=1.2mm] (10,79.5) rectangle (77,87.5);
\draw[draw=tiBlue,line width=1.15pt] (10.35,80.9) -- (10.35,86.1);
\node[text=tiInk,align=center] at (43.8,83.5) {Compute hidden-state differences\\between training and inference engines};

% L3: Target the layer with the largest increase in hidden-state difference
\path[draw=tiBorder,fill=white,line width=0.55pt,rounded corners=1.2mm] (10,68.5) rectangle (77,76.5);
\draw[draw=tiBlue,line width=1.15pt] (10.35,69.9) -- (10.35,75.1);
\node[text=tiInk,align=center] at (43.8,72.5) {Target the layer with the largest\\increase in hidden-state difference};

% L4: Same input (4 lines)
\path[draw=tiBorder,fill=white,line width=0.55pt,rounded corners=1.2mm] (10,50.3) rectangle (77,65.5);
\draw[draw=tiBlue,line width=1.15pt] (10.35,51.9) -- (10.35,63.9);
\node[text=tiInk,align=center,font=\sffamily\fontsize{7.7}{9.4}\selectfont] at (43.8,57.9)
  {Same input\\Run forward passes in both engines for the target layer\\Save hidden states within the target layer\\Compare the target layer's output difference};

% L5: Consider mathematical mechanisms first (teal banner)
\path[fill=tiTealBg,draw=tiTeal!22,line width=0.45pt,rounded corners=1.2mm] (10,39.4) rectangle (77,45.4);
\node[text=tiTeal,font=\sffamily\bfseries\fontsize{8.0}{9.6}\selectfont,align=center] at (43.5,42.4)
  {Consider mathematical mechanisms first};

% L6: Validate the explanation of the target layer's difference
\path[draw=tiBorder,fill=white,line width=0.55pt,rounded corners=1.2mm] (10,27.6) rectangle (77,35.6);
\draw[draw=tiTeal,line width=1.15pt] (10.35,29.0) -- (10.35,34.2);
\node[text=tiInk,align=center] at (43.8,31.6)
  {Validate the explanation of the\\target layer's difference};

% R1: Consider kernel issues first (amber banner)
\path[fill=tiAmberBg,draw=tiAmber!22,line width=0.45pt,rounded corners=1.2mm] (93,91.5) rectangle (160,97.5);
\node[text=tiAmber,font=\sffamily\bfseries\fontsize{8.0}{9.6}\selectfont,align=center] at (126.5,94.5)
  {Consider kernel issues first};

% R2: Compute hidden-state differences from the same-input forward passes
\path[draw=tiBorder,fill=white,line width=0.55pt,rounded corners=1.2mm] (93,79.5) rectangle (160,87.5);
\draw[draw=tiAmber,line width=1.15pt] (93.35,80.9) -- (93.35,86.1);
\node[text=tiInk,align=center] at (126.8,83.5)
  {Compute hidden-state differences\\from the same-input forward passes};

% R3: Target the internal layer with the largest increase in hidden-state difference
\path[draw=tiBorder,fill=white,line width=0.55pt,rounded corners=1.2mm] (93,68.5) rectangle (160,76.5);
\draw[draw=tiAmber,line width=1.15pt] (93.35,69.9) -- (93.35,75.1);
\node[text=tiInk,align=center] at (126.8,72.5)
  {Target the internal layer with the largest\\increase in hidden-state difference};

% R4: Kernel-level? (amber diamond, center (126.5, 60.2))
\path[draw=tiAmber,fill=white,line width=0.68pt]
  (126.5,65.2) -- (147.5,60.2) -- (126.5,55.2) -- (105.5,60.2) -- cycle;
\node[text=tiAmber,font=\sffamily\bfseries\fontsize{8.0}{9.6}\selectfont] at (126.5,60.2) {Kernel-level?};

% R5: Verify kernel faults / Repair confirmed faults
\path[draw=tiBorder,fill=white,line width=0.55pt,rounded corners=1.2mm] (93,42.2) rectangle (160,50.2);
\draw[draw=tiAmber,line width=1.15pt] (93.35,43.6) -- (93.35,48.8);
\node[text=tiInk,align=center] at (126.8,46.2)
  {Verify kernel faults\\Repair confirmed faults};

% R6: Rerun full-model forward passes in both engines (3 lines)
\path[draw=tiBorder,fill=white,line width=0.55pt,rounded corners=1.2mm] (93,26.6) rectangle (160,38.2);
\draw[draw=tiBlue,line width=1.15pt] (93.35,28.2) -- (93.35,36.6);
\node[text=tiInk,align=center,font=\sffamily\fontsize{7.8}{9.5}\selectfont] at (126.8,32.4)
  {Rerun full-model forward passes in both engines\\Recompute layer-wise differences\\Check final-output differences};

% Bottom Diamond: Any remaining differences requiring analysis? (center (85, 14.5))
\path[draw=tiBlue,fill=white,line width=0.68pt]
  (85,20.3) -- (119.5,14.5) -- (85,8.7) -- (50.5,14.5) -- cycle;
\node[text=tiBlue,font=\sffamily\bfseries\fontsize{7.8}{9.4}\selectfont,align=center] at (85,14.5)
  {Any remaining differences\\requiring analysis?};

% Final Box: Report verified fixes / Summarize supported explanations
\path[draw=tiBorder,fill=white,line width=0.55pt,rounded corners=1.2mm] (46.5,-2.8) rectangle (123.5,5.2);
\draw[draw=tiBlue,line width=1.15pt] (46.85,-1.4) -- (46.85,3.8);
\node[text=tiInk,align=center] at (85.3,1.2)
  {Report verified fixes\\Summarize supported explanations};

% Left column vertical arrows
\draw[arrSlate] (43.5,90.5) -- (43.5,87.5);
\draw[arrSlate] (43.5,79.5) -- (43.5,76.5);
\draw[arrSlate] (43.5,68.5) -- (43.5,65.5);
\draw[arrTeal] (43.5,50.3) -- node[lblTeal,right=1.2mm] {Small} (43.5,45.4);
\draw[arrTeal] (43.5,39.4) -- (43.5,35.6);

% Fail dashed loop (L6.east -> L5.east)
\draw[arrTealDash] (77,31.6) -- node[lblTeal,above=0.5mm] {Fail} (84.2,31.6) -- (84.2,42.4) -- (77,42.4);

% Large solid orthogonal arrow (L4 upper-east -> R1.west)
\draw[arrAmber] (77,60.2) -- node[lblAmber,above=0.5mm] {Large} (85.0,60.2) -- (85.0,94.5) -- (93,94.5);

% No dashed orthogonal arrow (R4.west -> L4 lower-east, non-crossing!)
\draw[arrAmberDash] (105.5,60.2) -- node[lblAmber,above=0.5mm,pos=0.46] {No} (89.2,60.2) -- (89.2,54.8) -- (77,54.8);

% Right column vertical arrows
\draw[arrAmber] (126.5,91.5) -- (126.5,87.5);
\draw[arrAmber] (126.5,79.5) -- (126.5,76.5);
\draw[arrAmber] (126.5,68.5) -- (126.5,65.2);
\draw[arrAmber] (126.5,55.2) -- node[lblAmber,right=1.2mm] {Yes} (126.5,50.2);
\draw[arrSlate] (126.5,42.2) -- (126.5,38.2);

% Convergence into bottom diamond
\draw[draw=tiSlate,line width=0.65pt] (43.5,27.6) -- (43.5,23.2) -- (126.5,23.2) -- (126.5,26.6);
\node[lblTeal,anchor=west] at (44.6,25.5) {Pass};
\fill[tiBlue] (85,23.2) circle (0.65mm);
\draw[arrSlate] (85,23.2) -- (85,20.3);

% Outer Yes dashed loop (bottom diamond west -> L3.west)
\draw[arrBlueDash] (50.5,14.5) -- (1.2,14.5) -- (1.2,72.5) -- (10,72.5);
\node[lblBlue,anchor=south] at (26.2,15.1) {Yes};
\node[lblBlue,font=\sffamily\fontsize{6.6}{8.0}\selectfont,anchor=north] at (26.2,13.6)
  {(Skip layers whose differences have been explained)};

% Final No arrow
\draw[arrBlue] (85,8.7) -- node[lblBlue,right=1.2mm] {No} (85,5.2);

\end{tikzpicture}%
\endgroup}
  \caption{Progressive diagnosis and alignment of training--inference computations. Layer-wise hidden-state comparisons identify the layer with the largest increase in discrepancy and guide finer inspection down to kernel-level computations. Same-input replay directs the analysis toward local alignment or a mathematical explanation of difference amplification. Full-model comparisons validate alignment changes and guide further analysis.}
  \label{fig:train-infer-diagnosis-flow}
\end{figure}

The core idea is to trace training--inference differences from model layers down to their internal computations. Both engines first run with the same model weights and saved token sequences, with hidden states recorded at each layer. Comparing these hidden states identifies the layer where the discrepancy grows most. That layer is then examined in greater detail, progressively narrowing the search until a kernel-level inconsistency is located or the mathematical mechanism behind the amplification is explained. Figure~\ref{fig:train-infer-diagnosis-flow} summarizes the process.

Cosine similarity is the main metric for selecting where to inspect the two engines. Let $x_T$ and $x_I$ be their activation matrices at the same point in the model. Rows $x_{T,i}$ and $x_{I,i}$ are the full output vectors for the same valid token $i$. The metric is the mean token cosine $\operatorname{Cos}(x_T,x_I)=\frac{1}{|V|}\sum_{i\in V}(x_{T,i}^{\mathsf T}x_{I,i})/(\lVert x_{T,i}\rVert_2\lVert x_{I,i}\rVert_2)$, where $V$ is the set of valid tokens. The output difference is defined as $\operatorname{Diff}(x_T,x_I)=1-\operatorname{Cos}(x_T,x_I)$. This measure captures differences in output direction and provides a practical basis for locating layers that amplify discrepancies. Layers are ranked by the increase in Diff from input to output. A cosine close to one does not establish agreement in magnitude, so the largest absolute element difference, $\lVert x_T-x_I\rVert_{\infty}:=\max_{i,j}|x_{T,i,j}-x_{I,i,j}|$, provides an additional check when needed.

Same-input replay tests whether the training--inference difference originates within the target layer. The replay feeds the exact layer input captured from the training engine (and, symmetrically, from the inference engine) into both engines' implementations of that target layer, saving intermediate hidden states within the layer. A large output difference under identical layer inputs directs the search toward finer-level inconsistencies within the target layer. The search then drills down recursively inside the target layer: hidden-state differences from the same-input forward passes are computed, the internal component with the largest discrepancy increase is selected, and the inspection checks whether a kernel has been reached. If the selected component is still a composite sub-module, it becomes the new target for finer-grained same-input forward passes and internal hidden-state comparison. Once a kernel is reached, its computational rules, configuration, and precision are checked across the engines to confirm the source of the discrepancy and then targeted alignment changes are applied. Full-model forward passes are then rerun in both engines to recompute layer-wise and final-output differences.

A small difference under identical inputs shifts the analysis toward mathematical sensitivity to upstream differences. Candidate mechanisms include changes in MoE expert selection and amplification associated with the multi-head latent attention (MLA) output projection matrices $W_{oa}$ and $W_{ob}$~\citep{deepseekv3}. Finer-grained comparisons locate the amplification. Controlled replay across higher-level modules then tests whether the proposed mechanism explains the observed amplification. After an alignment change is verified or an explanation is validated, the workflow checks whether any layer discrepancy still requires explanation. If so, the remaining layer with the largest discrepancy increase becomes the next target. Otherwise, the workflow ends with a report of the alignment changes, validation results, and mathematical explanations.

The complete workflow is packaged as a reusable Skill that enables coding agents to automate the diagnostic process. The Skill is publicly available at \url{https://github.com/SLAI-AITP/rl-train-infer-diagnosis-tool}.

\subsubsection{Operator Defect Isolation and Alignment on Ascend}

We evaluate both forward-pass numerical agreement and end-to-end RL training stability using DeepSeek-V4-Flash-BF16 across eight nodes, each equipped with eight Huawei Ascend 910C NPUs (64 NPUs in total). The training backend uses MindSpeed-LLM/Megatron with TP4/EP64/PP2 and sequence parallelism, while the rollout backend uses vLLM-Ascend with TP8/EP8/PP1 and full-context prefill. Forward-pass diagnostics run on a fixed panel of 16 prompt--response sequences drawn from 16 distinct prompts.

\begin{figure}[!htbp]
  \centering
  \includegraphics[width=0.49\linewidth]{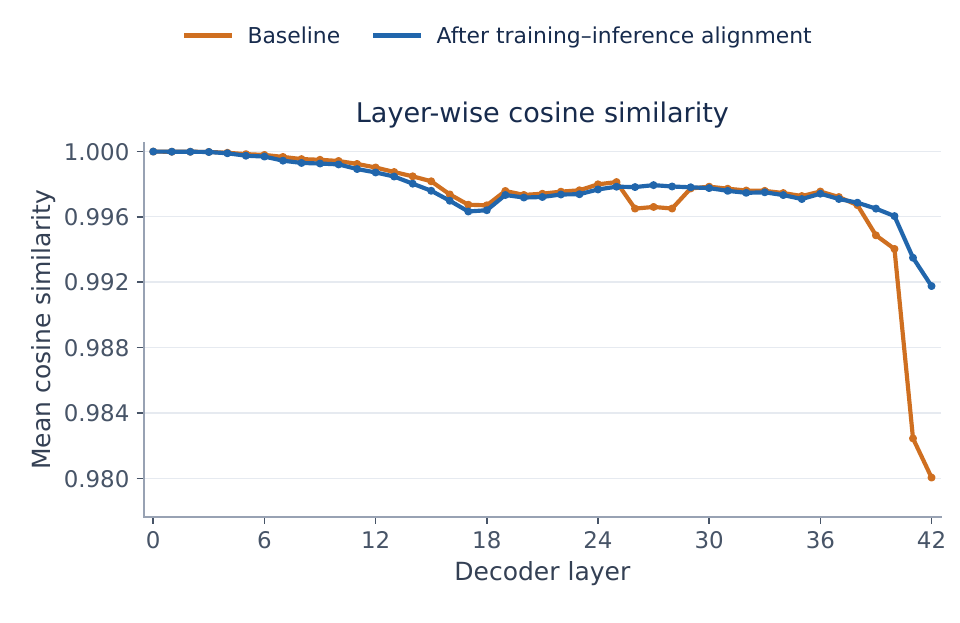}\hfill
  \includegraphics[width=0.49\linewidth]{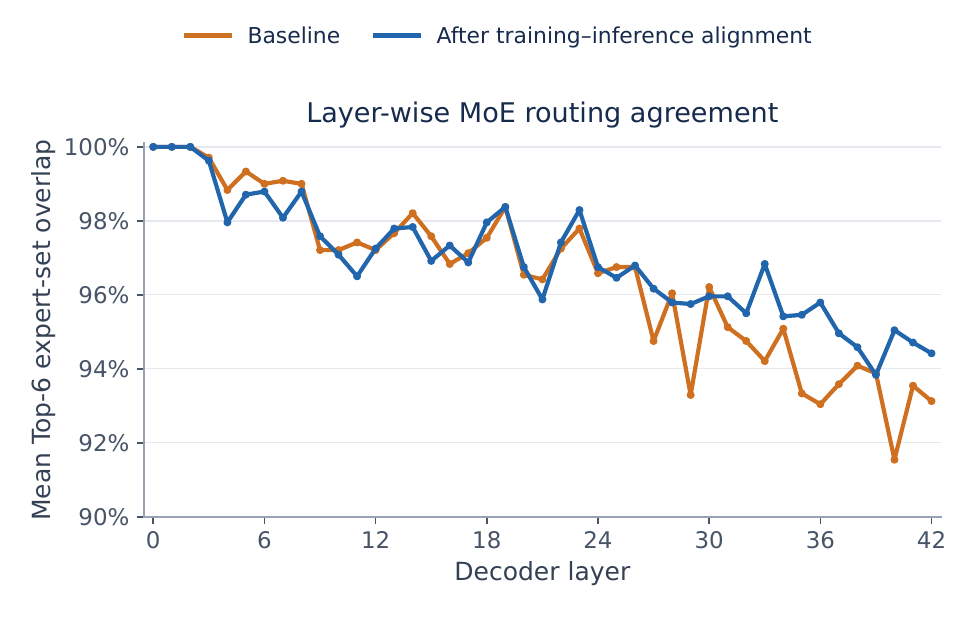}
  \caption{Layer-wise training--inference cosine similarity (left) and MoE routing agreement (right) on the fixed baseline Top-400 high-discrepancy token positions before and after alignment. Routing agreement measures the mean intersection size of the two Top-6 expert sets divided by six.}
  \label{fig:train-infer-layer-comparison}
\end{figure}

Applying the progressive localization workflow pinpoints two concrete implementation discrepancies between the engines: (i) inconsistent application of the configured SwiGLU activation clipping limit along the MoE expert feed-forward path, and (ii) a precision mismatch in the data type used to load the MoE router correction bias. In a 43-layer sparse MoE architecture, even slight pre-router numerical perturbations near Top-6 expert selection boundaries flip discrete expert assignments and compound across depth. Aligning both computations substantially improves forward agreement across all decoder layers (L0--L42). Figure~\ref{fig:train-infer-layer-comparison} compares layer-wise agreement before and after alignment on the 400 token positions exhibiting the largest baseline training--inference log-probability discrepancies: at the final decoder layer (L42), mean token cosine rises from $0.980049$ to $0.991762$, mean Top-6 expert-set overlap rises from $93.13\%$ to $94.42\%$, and mean absolute log-probability discrepancy drops by $34.1\%$ (from $0.604612$ to $0.398707$), with residual differences attributable to distinct tensor/expert parallel reduction orders.

\begin{figure}[!htbp]
  \centering
  \includegraphics[width=0.96\linewidth]{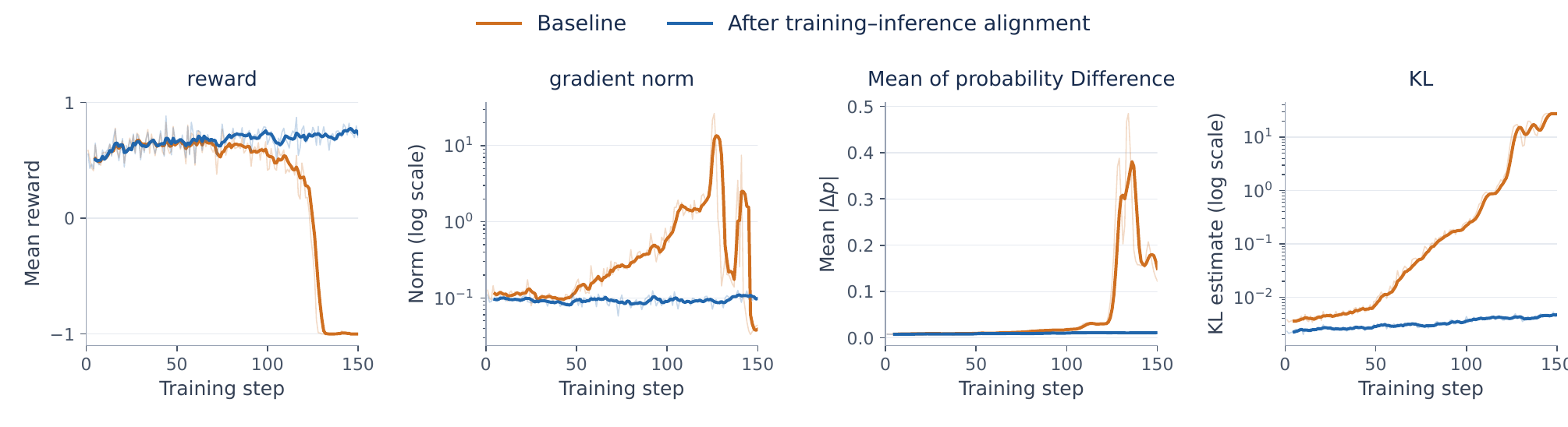}
  \caption{Reward, gradient norm, mean absolute token-probability difference, and the rollout-to-training KL estimate for the baseline and candidate aligned runs, from left to right. Thin lines show raw values; thick lines show trailing five-step means. Gradient norm and KL use logarithmic axes.}
  \label{fig:train-infer-bug-comparison}
\end{figure}

End-to-end RL training over 150 updates confirms the downstream stability impact of these repairs (Figure~\ref{fig:train-infer-bug-comparison}). In the unaligned baseline, rising token-probability discrepancies and rollout-to-learner KL spikes precede abrupt gradient-norm explosion and reward collapse around step 130. After aligning the SwiGLU clipping and router-bias precision paths, token-probability differences and KL divergence remain bounded, gradient norms stay flat, and reward improves steadily, preventing numerical-drift-induced RL collapse across the 150-update horizon.

\FloatBarrier
% Speculative-decoding section; numbering and packages are supplied by the parent report.
\subsection{Native Multi-Token Prediction and Decoupled OR--MTP Alignment}
\label{sec:sdv2}
\label{sec:mtp-alignment-deployment}
\label{sdinsert:sec:speculative-decoding}

Operations-research (OR) solutions contain mathematical formulations and executable solver code, making generation efficiency a practical concern. Speculative decoding reduces sequential target-model evaluations by verifying several draft tokens in one pass~\citep{leviathan2023}. Multi-token prediction (MTP) extends next-token training to several future tokens~\citep{gloeckle2024}. Native MTP generates drafts from target features, shared embeddings and the output head, together with parameters specific to the MTP module. Its serving benefit depends on draft acceptance and the computational cost of drafting and verification.

OR reinforcement learning (RL) updates the target policy to improve formulation correctness, whereas auxiliary MTP cross-entropy trains multi-step continuation prediction through shared representations and weights. Related work studies speculative decoding for RL rollouts~\citep{iso2026}, joint MTP--RL coefficient calibration~\citep{occ2026}, and confidence-scheduled drafting~\citep{dspark2026}. Here we examine OR--MTP co-adaptation within the SLAI T-Rex stack~\citep{slai2026trexpart1} across three stages: workload-dependent draft acceptance, auxiliary-gradient routing during joint RL, and private MTP refinement on a frozen OR target.

\subsubsection{Workload-Dependent Draft Acceptance}
\label{sec:sdv2-motivation}
\label{sec:mtp-workload-dependence}

\begin{figure}[!htbp]
\centering
\includegraphics[width=0.95\linewidth]{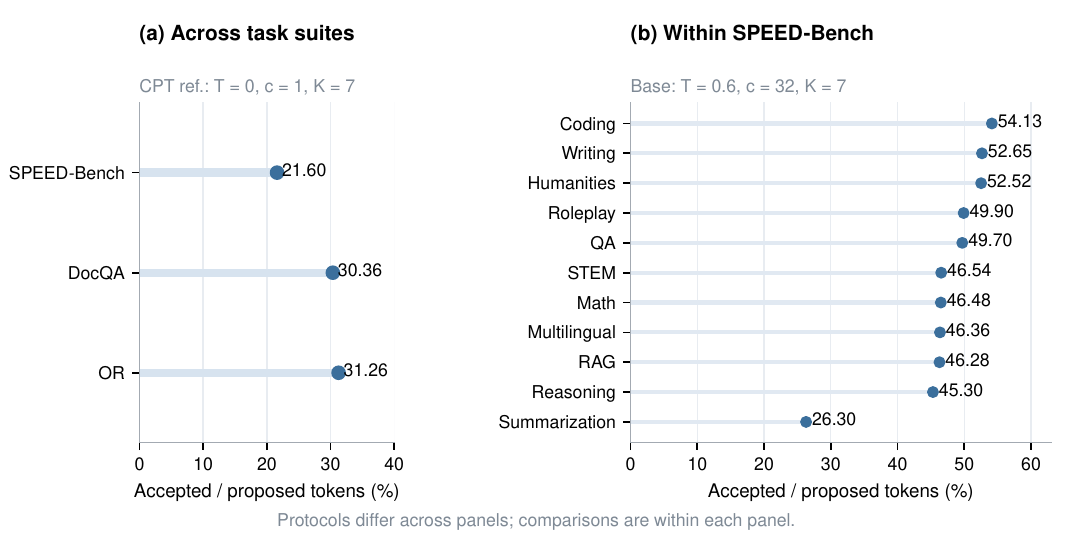}
\caption{Native-MTP acceptance across workloads. Panel (a) pools accepted and proposed token counts on the historical CPT reference with greedy decoding and concurrency one. Panel (b) compares SPEED-Bench categories on the base checkpoint at temperature 0.6 and concurrency 32. Both panels use seven proposals under panel-specific checkpoints and decoding protocols.}
\label{fig:sdv2-workload}
\end{figure}

We evaluate native MTP across SPEED-Bench, NPU-Bench, OR, and long-context document question answering (DocQA) under controlled configurations within each benchmark panel (Figure~\ref{fig:sdv2-workload}). On the historical CPT reference checkpoint, greedy decoding with seven draft proposals at concurrency one yields pooled acceptance rates of 21.60\% on SPEED-Bench (896 requests), 30.36\% on DocQA (192 requests), and 31.26\% on OR (256 requests). On the base checkpoint at temperature 0.6, concurrency 32, and seven proposals (80 prompts per category), acceptance across the eleven SPEED-Bench categories spans more than a two-fold range, from 26.30\% on summarization to 54.13\% on coding.

Substantial cross-workload variation motivates evaluating speculative decoding directly on target OR workloads rather than extrapolating from generic coding benchmarks. Moreover, draft acceptance alone does not determine end-to-end serving efficiency: under stochastic sampling on open-ended SPEED-Bench categories, trajectory divergence and stopping-condition sensitivity cause MTP runs to emit $4.21\times$--$11.36\times$ as many completion tokens as standard decoding. Because longer completions can inflate raw token-per-second (TPS) rates without shortening user-perceived latency, we jointly evaluate solver-verified OR accuracy, token throughput, and end-to-end request latency below.

\subsubsection{Joint Optimization of the OR Policy and Native MTP}
\label{sec:sdv2-training}

To study representation sharing during RL, we run controlled 200-update joint-training experiments initialized from our native-MTP CPT lineage~\citep{slai2026trexpart1} (distinct from the 0731 SFT--RL lineage in Section~\ref{sec:or-bench-post-training}). Each update samples eight responses across 64 prompts from a 4,322-record self-contained OR code dataset with binary solver rewards, group-relative clipped policy optimization, and a $10^{-3}$ reference KL penalty (rollouts use standard target decoding so training dynamics isolate gradient-routing effects).

\begin{figure}[!htbp]
\centering
\includegraphics[width=0.95\linewidth]{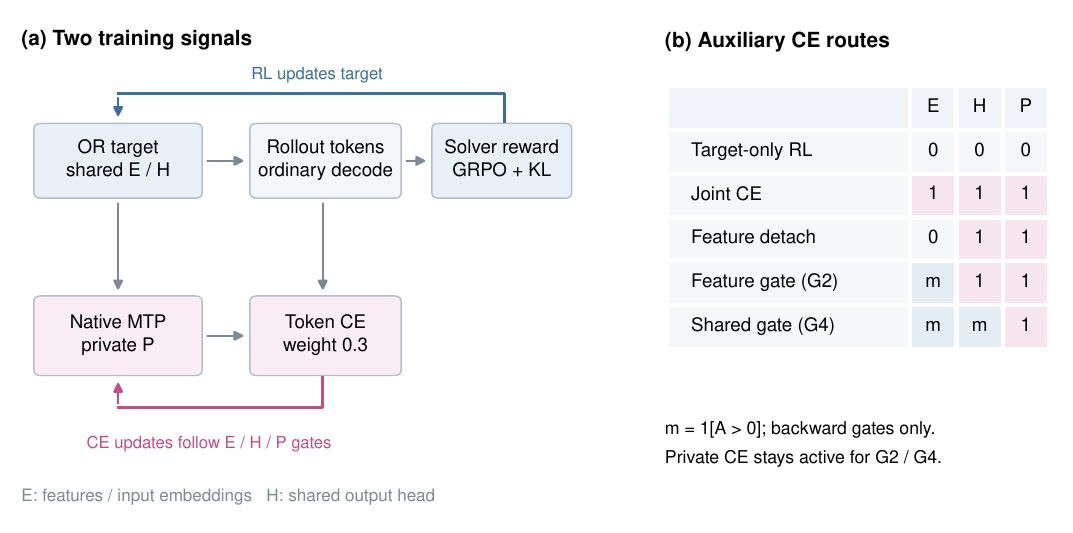}
\caption{Joint OR--MTP optimization and auxiliary-gradient routing. Solver rewards update the target through the RL objective, while backward gates $(a_E,a_H,a_P)$ selectively route teacher-forced MTP cross-entropy gradients through shared and private parameter blocks.}
\label{fig:sdv2-training}
\end{figure}

Let $\ell_i^{\mathrm{MTP}}$ denote the native one-step MTP cross-entropy averaged over valid shifted positions of response $i$. We control auxiliary gradients into target features and input embeddings ($E$), the shared output head ($H$), and private MTP parameters ($P$):
\begin{equation}
g_b=\nabla_b\mathcal L_{\mathrm{RL}}
 +\frac{0.3}{B}\sum_{i=1}^{B}a_{b,i}\nabla_b \ell_i^{\mathrm{MTP}},
\qquad b\in\{E,H,P\},\qquad \nabla_P\mathcal L_{\mathrm{RL}}=0,
\label{eq:sdv2-routes}
\end{equation}
where $B$ is the batch response count and $a_{b,i}$ gates auxiliary supervision into parameter block $b$ (Figure~\ref{fig:sdv2-training}). Target-only RL uses $(0,0,0)$; Joint CE uses $(1,1,1)$; Feature Detachment uses $(0,1,1)$; and the positive-advantage gated variants Feature Gating (denoted G2) and Shared Gating (denoted G4) use $(m_i,1,1)$ and $(m_i,m_i,1)$ with $m_i=\mathbf 1[A_i>0]$, retaining private MTP supervision on all valid responses. Because all-correct binary groups yield $A_i=0$ (accounting for $\approx 97.1\%$ of the $\approx 90\%$ uniform groups in G2/G4), positive-advantage gating leaves shared auxiliary supervision sparse ($6.98\%/6.63\%$ of responses).

\subsubsection{OR Quality and Draft Acceptance after Joint Training}
\label{sec:sdv2-quality}
\label{sec:mtp-quality}

\begin{figure}[!htbp]
\centering
\includegraphics[width=0.95\linewidth]{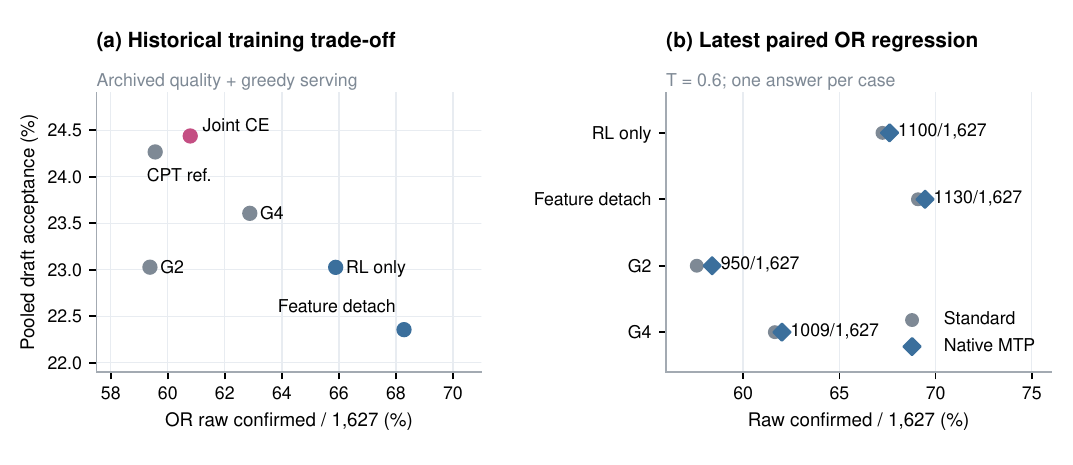}
\caption{OR quality and draft acceptance. Panel (a) plots raw OR accuracy against pooled greedy acceptance across archived endpoints. Panel (b) compares standard (STD) and native-MTP decoding across the four step-200 targets on the 1,627-case regression suite.}
\label{fig:sdv2-outcomes}
\end{figure}

Archived joint-training endpoints suggest an empirical tension between OR task accuracy and draft acceptance (Figure~\ref{fig:sdv2-outcomes}a): full Joint CE slightly raises pooled greedy acceptance relative to target-only RL (24.44\% versus 23.03\%) alongside lower raw OR accuracy (60.79\% versus 65.89\%), whereas Feature Detachment reaches 68.29\% OR accuracy with 22.36\% acceptance; because these archived runs also differ in recipe and seed provenance (Appendix~\ref{app:sdv2}), we treat them as motivating context and conduct a controlled four-target comparison next.

To audit decoding quality systematically, we evaluate the four step-200 checkpoints (target-only RL, Feature Detachment, G2, and G4) on a 1,627-case OR regression suite (289 NL4Opt, 600 OptiBench, 344 Bench4Opt-Feasible, and 394 structural LP cases) at temperature 0.6, top-$p=1$, seed 42, and a 9,600-token cap. Denoting the Feature Detachment target as \emph{Trunk} and its saved MTP module as $\phi_0$, we also include a frozen-target post-RL private-MTP refinement condition (denoted H3; Section~\ref{sec:sdv2-refinement}) and report both raw code extraction and a format-cleaned code extraction pass (denoted H1). Table~\ref{tab:sdv2-quality} and Figure~\ref{fig:sdv2-outcomes}b summarize all nine conditions.

\begin{table}[!htbp]
\centering\small
\renewcommand{\arraystretch}{1.10}
\caption{OR correctness on the 1,627-case regression pool under primary raw extraction and format-cleaned extraction (H1). Each condition has one unresolved license case retained in the denominator. H3 refines only the private MTP module on the frozen Feature Detachment (Trunk) target.}
\label{tab:sdv2-quality}
\begin{tabular*}{\linewidth}{@{\extracolsep{\fill}}llrrr@{}}
\toprule
\textbf{Target} & \textbf{Decoding} & \textbf{Confirmed / 1,627} & \textbf{Score bounds (\%)} & \textbf{H1 confirmed} \\
\midrule
Target-only RL & STD & 1094 & 67.24--67.30 & 1120 \\
Target-only RL & MTP & 1100 & 67.61--67.67 & 1127 \\
Feature detachment & STD & 1124 & 69.08--69.15 & 1124 \\
Feature detachment & MTP & 1130 & 69.45--69.51 & 1130 \\
Feature gating (G2) & STD & 937 & 57.59--57.65 & 1027 \\
Feature gating (G2) & MTP & 950 & 58.39--58.45 & 1052 \\
Shared gating (G4) & STD & 1003 & 61.65--61.71 & 1046 \\
Shared gating (G4) & MTP & 1009 & 62.02--62.08 & 1054 \\
\midrule
Feature detach + H3 & MTP & \textbf{1138} & \textbf{69.94--70.01} & \textbf{1138} \\
\bottomrule
\end{tabular*}

\end{table}

Feature Detachment (\emph{Trunk}) achieves the strongest raw accuracy among the four targets (69.08\%--69.15\% STD), while G2 and G4 exhibit higher sensitivity to output formatting (partially recovered under H1 cleanup). Enabling the saved native MTP module shifts raw confirmed counts by $+6,+6,+13,+6$ across the four targets, with all paired 95\% cluster confidence intervals spanning zero (Table~\ref{tab:sdv2-paired}), indicating no statistically detectable degradation from speculative verification.

\subsubsection{Serving Performance of the Trained Checkpoints}
\label{sec:sdv2-serving}
\label{sec:mtp-deployment-protocol}

We benchmark offline serving performance by comparing standard decoding against each checkpoint's saved native MTP module on 224 requests (128 OR and 96 DocQA) at client concurrencies $N_{\mathrm{conc}}\in\{1,32\}$ across three paired host blocks per checkpoint. Target weights, KV-cache budgets, and hardware topology are matched within each pair (BF16 across 16 logical Ascend NPUs with TP4/DP4/EP16, 131,072-token context, 8,192-token batching budget, decode graphs for the target, and seven eager draft proposals per round at temperature 1, seed 42, and thinking disabled).

\phantomsection\label{sec:mtp-performance}
At $N_{\mathrm{conc}}=1$, enabling native MTP delivers \textbf{$1.583\times$--$1.715\times$} standard token throughput on OR across all twelve checkpoint--host pairs (versus $0.937\times$--$1.066\times$ on DocQA and $1.420\times$--$1.504\times$ overall) and reduces OR end-to-end request latency by $1.659\times$--$1.748\times$, despite a modest prefill overhead reflected in standard/MTP time-to-first-token (TTFT) ratios of $0.821$--$0.830$. At $N_{\mathrm{conc}}=32$, OR throughput ratios remain positive across all twelve pairs ($1.082\times$--$1.717\times$), whereas DocQA spans $0.661\times$--$1.316\times$ as batching contention increases (timing counts retain 201 isolated BOS tokens observed in $N_{\mathrm{conc}}=32$ MTP streams). Because target-only RL also realizes comparable speedups from its inherited MTP weights, these gains reflect enabling native MTP on structured OR code/formulation workloads rather than joint auxiliary training.

\subsubsection{Private-MTP Refinement with a Fixed OR Target}
\label{sec:sdv2-refinement}
\label{sec:mtp-predictive-evidence}

\phantomsection\label{sec:mtp-private-objective}
To improve draft alignment without perturbing the best OR policy, H3 freezes the Feature Detachment (\emph{Trunk}) target---including shared embeddings and the output head---and updates only the private MTP parameters from $\phi_0$ for 20 steps using three-step recursive cross-entropy (Appendix~\ref{app:sdv2}). Teacher-forced three-token prefix agreement on the OR development panel rises from \textbf{48.44\% to 60.16\%}, while 1,627-case solver correctness remains intact ($1{,}138$ versus $1{,}130$ confirmed solves; paired difference $+0.49$ percentage points, 95\% CI $[-0.86,+1.85]$).

\begin{figure}[!htbp]
\centering
\includegraphics[width=0.88\linewidth]{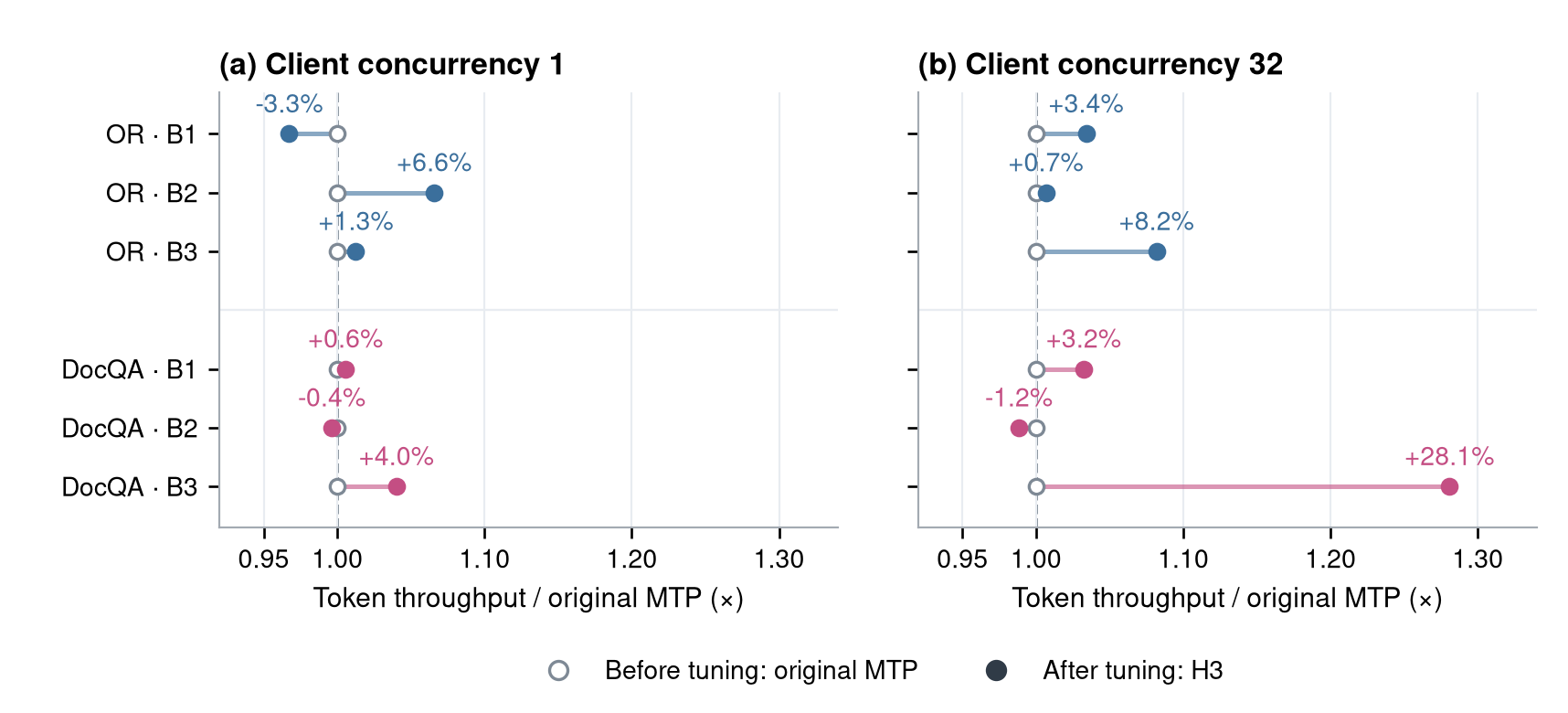}
\caption{Paired token throughput before ($\phi_0$, hollow circles normalized to $1.0$) and after private-MTP refinement (H3, filled circles) on the frozen Trunk target across three host blocks ($B_1$--$B_3$) at client concurrencies 1 and 32.}
\label{fig:sdv2-tuning}
\end{figure}

Figure~\ref{fig:sdv2-tuning} evaluates the incremental serving impact of H3 relative to $\phi_0$ on matched hosts: OR token throughput shifts by $-3.3\%$, $+6.6\%$, and $+1.3\%$ across blocks $B_1$--$B_3$ at $N_{\mathrm{conc}}=1$, and by $+3.4\%$, $+0.7\%$, and $+8.2\%$ at $N_{\mathrm{conc}}=32$, while overall paired $\phi_0$/H3 request-latency ratios improve modestly to $1.028$ $[1.012, 1.044]$ at $N_{\mathrm{conc}}=1$ and $1.019$ $[1.004, 1.035]$ at $N_{\mathrm{conc}}=32$.

\phantomsection\label{sec:mtp-delivery-boundaries}
\paragraph*{Scope and Takeaways.}
Taken together, our experiments show that (i) detaching trunk features during RL followed by frozen-target private MTP refinement avoids the accuracy--acceptance trade-offs observed in archived shared-gradient joint runs while lifting multi-token agreement; and (ii) native MTP substantially accelerates low-concurrency OR decoding ($1.583\times$--$1.715\times$ TPS), though high-concurrency contention and stochastic length variation narrow gains. Integrating lossless MTP verification directly into high-concurrency online RL rollout workers to measure end-to-end training wall-clock savings remains an important next step.

\subsection{Solver-Guided Asynchronous Prefix RL for NPU Kernel Synthesis}
\label{sec:kernel-perf-opt}
% Revised around the six-part outline supplied by the author.
% AscendGYM evidence: source-paper commit c6afcc1.
% WindMatmulV2 evidence: OR_page.pptx, slide 1, including the embedded timing plot.

\subsubsection{Problem Formulation and Cost-Aware Evaluation Criteria}

Low-level hardware kernels form the computational foundation of large-scale LLM training and inference, directly governing device utilization and throughput~\citep{flashattention,slai2026trexpart1}. Automated kernel optimization seeks to transform runnable baseline operators into faster, numerically equivalent implementations tailored to target accelerators~\citep{ouyang2025kernelbench}. While KernelBench~\citep{ouyang2025kernelbench} and CANN Bench~\citep{gao2026cannbench} evaluate kernel synthesis on GPUs and Ascend NPUs, practical deployment poses a sharper two-part question: \emph{can LLM agents reliably improve production-grade NPU kernels, and at what exploration cost?}

To study this trade-off systematically, we introduce \textsc{AscendGYM} for cost-aware iterative kernel optimization, examine solver-guided parameter warm-starting via OR modeling, and develop asynchronous candidate-prefix agentic RL to overcome multi-hour profiling bottlenecks.

\subsubsection{AscendGYM: Cost-Aware Kernel Optimization Evaluation over Real Workloads}
\label{sec:kernel-perf-opt-background}

\textsc{AscendGYM} comprises 48 operator tasks extracted from real LLM training and serving workloads across three difficulty tiers. Each task supplies a runnable baseline kernel, host binding code, workload-derived input shapes, golden reference outputs, data types, and datatype-specific precision checks. Its execution environment, \textsc{AscendForge}, standardizes native compilation, CPU-reference verification, and \texttt{msprof} device timing while logging full multi-turn agent trajectories across external coding harnesses or our built-in \textsc{AuraKernel} harness.

Because iterative optimization often consumes dozens of hardware evaluations, final speedup alone obscures search efficiency. In our empirical diagnosis below, we evaluate both optimization quality---via \textbf{geometric mean (GM) speedup} over baseline latencies and \textbf{task coverage ($\mathrm{fast}_p$)}~\citep{ouyang2025kernelbench}, the fraction of tasks strictly exceeding speedup threshold $p$---and exploration cost, tracked through candidate correctness pass rates, tool invocations, hardware evaluations, token consumption, and wall-clock search duration. (\textsc{AscendGYM} also defines budget-penalized summary extensions such as Speedup--Token Score and Cost-to-Gain for fixed-token leaderboard comparisons.)

\subsubsection{Empirical Diagnosis of Coverage and Exploration Bottlenecks}
\label{sec:kernel-results}

We evaluate representative frontier and open-weight models across three agent harnesses (Claude Code~\citep{anthropic2026claudecode}, Codex~\citep{openai_codex_cli}, and OpenCode) on the 17 production operators of \textsc{AscendGYM} Batch~1 (up to 80 candidate evaluations per task on Ascend~910C). Two capability bottlenecks emerge clearly.

\begin{figure}[!htbp]
    \centering
    \includegraphics[width=0.92\linewidth]{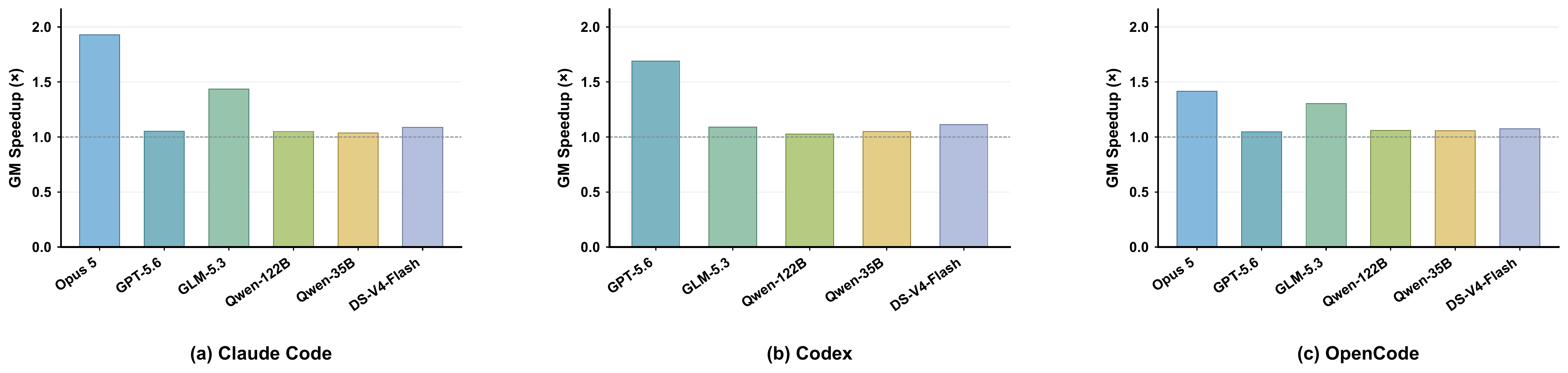}
    \caption{Geometric mean (GM) speedup across the 17 operators of \textsc{AscendGYM} Batch~1 for evaluated model--harness combinations, using the best correctness-verified candidate per task (or $1.0\times$ baseline when no improvement is found).}
    \label{fig:ascendgym-gm-speedup}
\end{figure}

\begin{figure}[!htbp]
    \centering
    \includegraphics[width=0.88\linewidth]{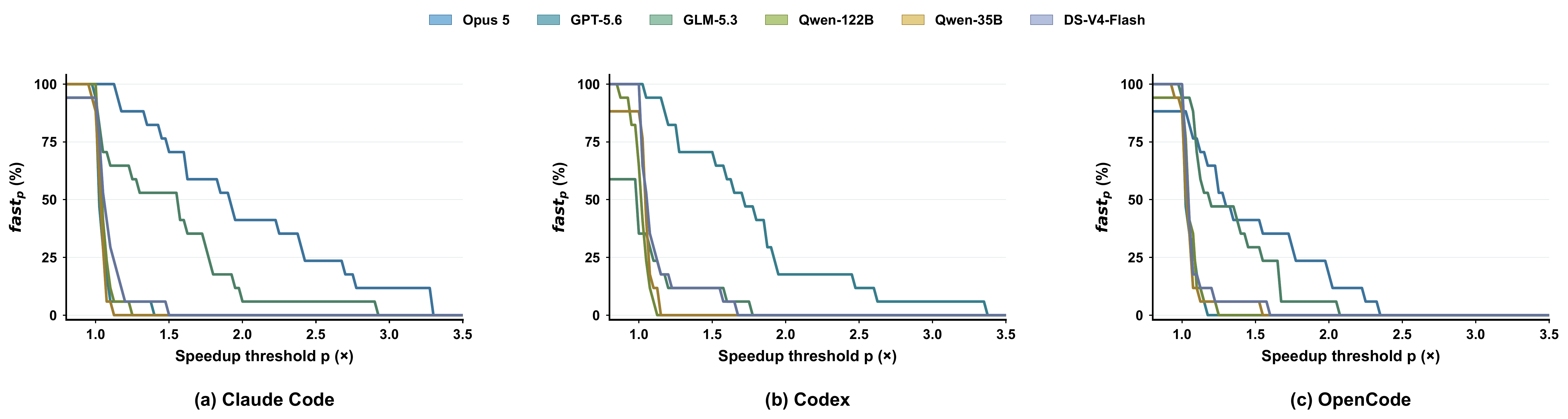}
    \caption{Task coverage ($\mathrm{fast}_p$) across speedup thresholds $p$ on \textsc{AscendGYM} Batch~1. Coverage drops sharply as the target speedup threshold increases.}
    \label{fig:ascendgym-fast-p}
\end{figure}

\paragraph{Substantial speedups remain narrow in task coverage.}
Retaining the fastest correctness-verified implementation discovered during search (or $1.0\times$ when no valid improvement is found), roughly $70\%$ of evaluated model--harness pairs achieve a GM speedup below $1.2\times$ (Figure~\ref{fig:ascendgym-gm-speedup}). Even the strongest configurations---Claude Code with Claude Opus~5~\citep{anthropic2026opus5} ($1.93\times$ GM speedup) and Codex with GPT-5.6 Terra~\citep{openai2026gpt56terra} ($1.69\times$ GM speedup)---exceed a $2\times$ speedup on only 7 and 3 of the 17 kernels, respectively (Figure~\ref{fig:ascendgym-fast-p}).

\begin{figure}[!htbp]
    \centering
    \includegraphics[width=0.94\linewidth, height=0.42\textheight, keepaspectratio]{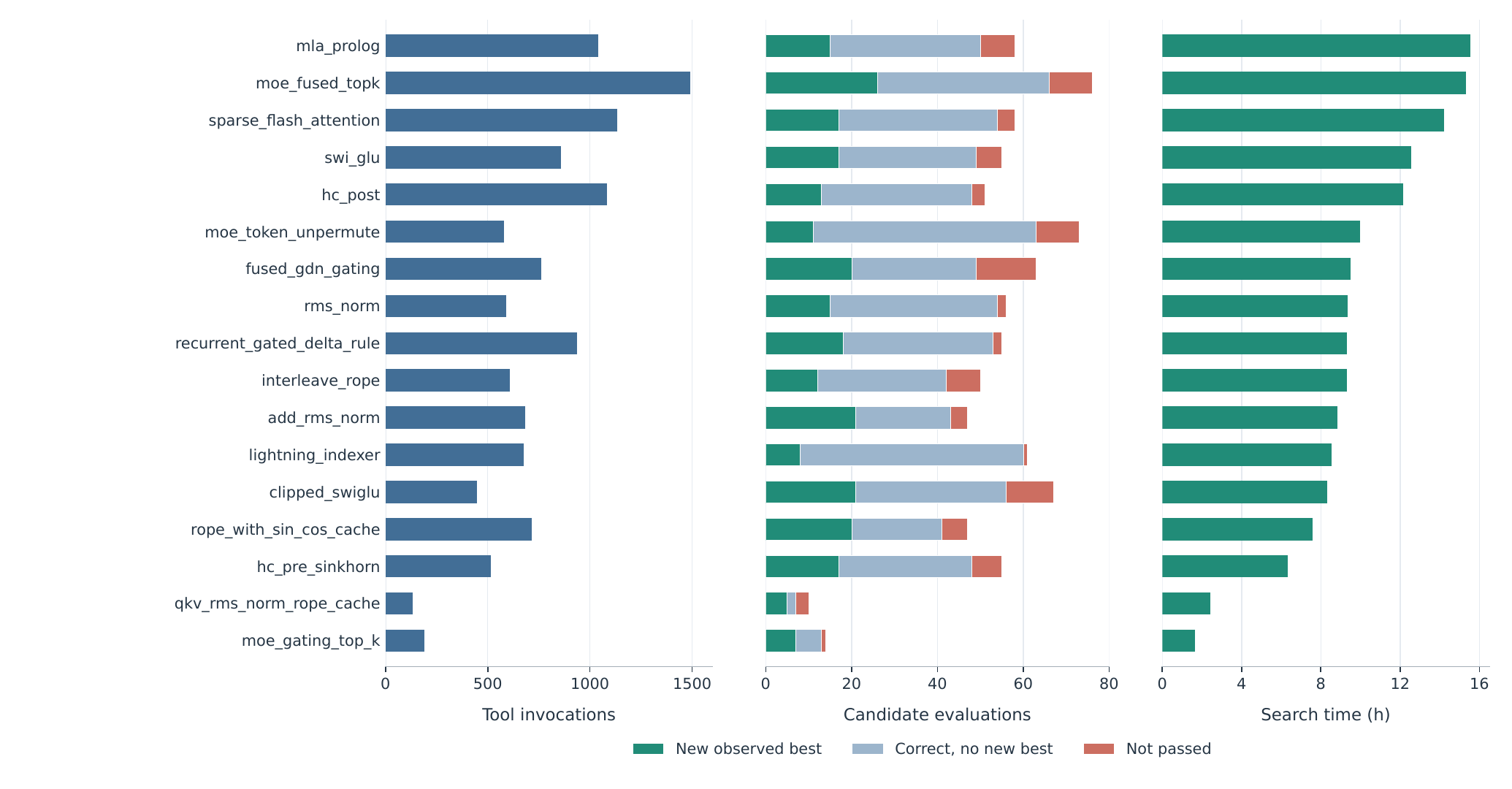}
    \caption{Per-operator exploration cost of Claude Code/Opus~5 across \textsc{AscendGYM} Batch~1: total tool invocations (left), candidate evaluation outcomes distinguishing new bests, non-improving valid kernels, and verification failures (middle), and total wall-clock search time in hours (right).}
    \label{fig:kernel-search-cost}
\end{figure}

\paragraph{High exploration cost and long episode tails.}
Reaching these gains requires an average of 81.75M and 140.4M tokens per task for Claude Code/Opus~5 and Codex/GPT-5.6 Terra, respectively. Trajectory breakdown for Claude Code/Opus~5 (Figure~\ref{fig:kernel-search-cost}) shows that individual tasks routinely span hundreds to over a thousand tool invocations, 40--75 hardware candidate evaluations, and 6--15 hours of wall-clock search---with a large share of profiler runs spent on valid candidates that yield no further speedup or candidates that fail verification.

Mitigating this bottleneck requires both stronger structural priors for coupled hardware parameters and more sample-efficient RL updates that do not stall behind multi-hour episodes. We address these two needs via OR-guided parameter warm-starting and asynchronous candidate-prefix agentic RL.

\subsubsection{Analytical Parameter Warm-Starting via OR Modeling}
\label{sec:kernel-or-modeling}

Before free-form code rewriting, key AscendC scheduling decisions---tile dimensions, inter-core partitioning, and on-chip L1 buffering---interact through analytical memory and compute constraints and can be formulated directly as an OR problem~\citep{slai2026trexpart1}. Solving this formulation supplies a hardware-validated warm start for subsequent agent refinement.

We illustrate this approach on \texttt{WindMatmulV2}, a BF16 matrix multiplication kernel computing $C=AB$ for $(M,K,N)=(16,4096,62080)$. Because $M=16$ severely restricts reuse of the $485\,\mathrm{MiB}$ weight matrix $B$, device profiling identifies HBM-to-on-chip data movement (MTE2 utilization $99.4\%$) as the primary bottleneck. Our OR model jointly selects matrix tile sizes, partitions work across cores along the long $N$ dimension, and configures multi-stage L1 buffering to overlap data prefetching with cube computation under on-chip memory limits. On Ascend hardware, this solver-selected configuration reduces mean kernel latency from $369.13\,\mu\mathrm{s}$ to $348.68\,\mu\mathrm{s}$ ($-5.54\%$), providing a verified analytical starting point without brute-force trial-and-error.

\subsubsection{Asynchronous Candidate-Prefix RL with Shared Lesson Memory}
\label{sec:kernel-agentic-rl}

Beyond parameter selection, general operator optimization requires iterative code synthesis, compiler/profiler diagnosis, and hypothesis revision via agentic RL~\citep{dai2026cudaagent}. Because full multi-turn episodes take hours to terminate, synchronous episode-level RL leaves learner NPUs idle for most of a training run.

\begin{figure}[!htbp]
  \centering
  \includegraphics[width=0.95\linewidth]{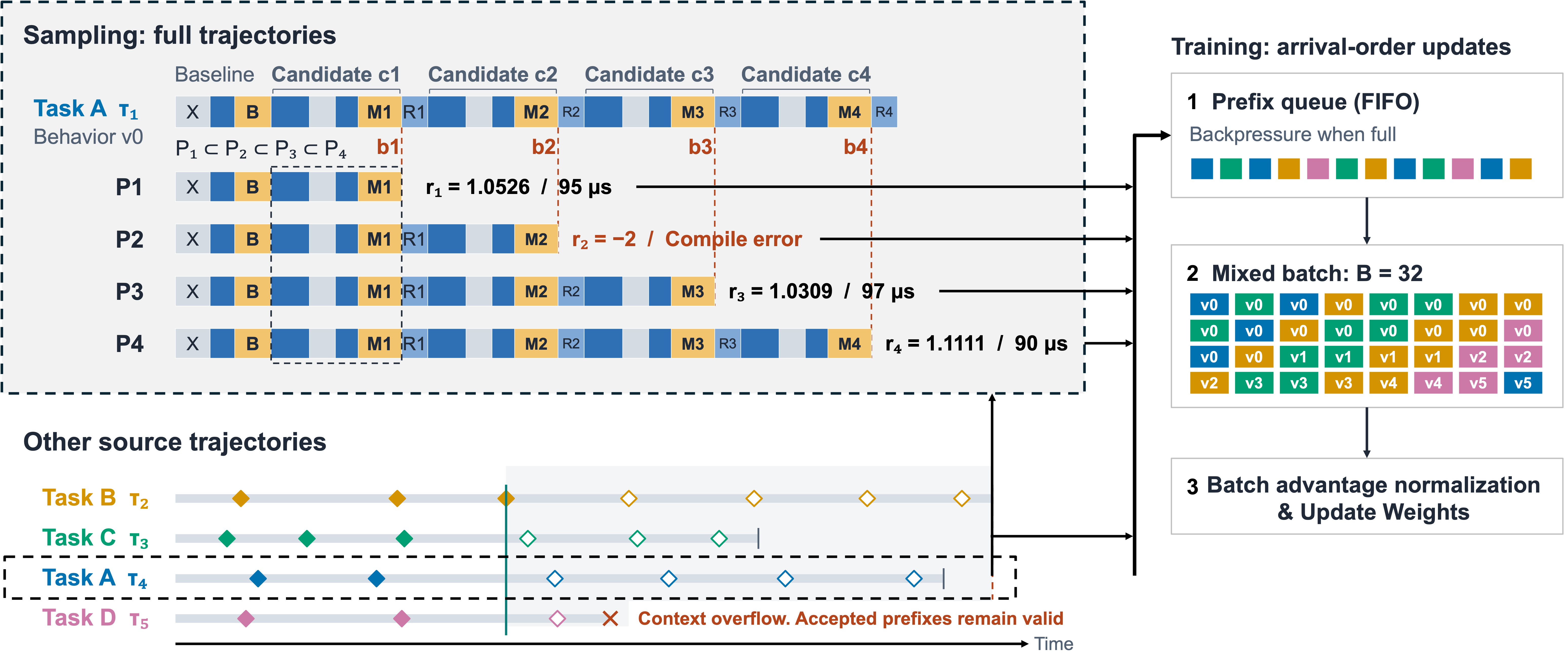}
  \caption{Asynchronous candidate-prefix agentic RL for Ascend kernel optimization. Each completed candidate profiling step emits a history-preserving trajectory prefix scored against the unmodified baseline, feeding a bounded FIFO queue for immediate learner updates without waiting for full episodes to terminate.}
  \label{fig:ops-perf-scheme}
\end{figure}

Notice, however, that every completed candidate profiling step within an episode already yields exact correctness and device-latency feedback. As illustrated in Figure~\ref{fig:ops-perf-scheme}, we convert each completed candidate evaluation boundary into a self-contained \textbf{trajectory prefix} preserving the full preceding interaction history and its baseline-relative speedup reward. Eligible prefixes from concurrent tasks enter an asynchronous FIFO queue to assemble mixed training batches immediately, complemented by an evidence-grounded \textbf{shared lesson memory} that publishes and retrieves verified optimization findings across trajectories. The training study uses 710 cases across 142 \texttt{ops-math} kernels.

\begin{figure}[!htbp]
  \centering
  \includegraphics[trim=24bp 12bp 6bp 8bp, clip, width=0.95\linewidth]{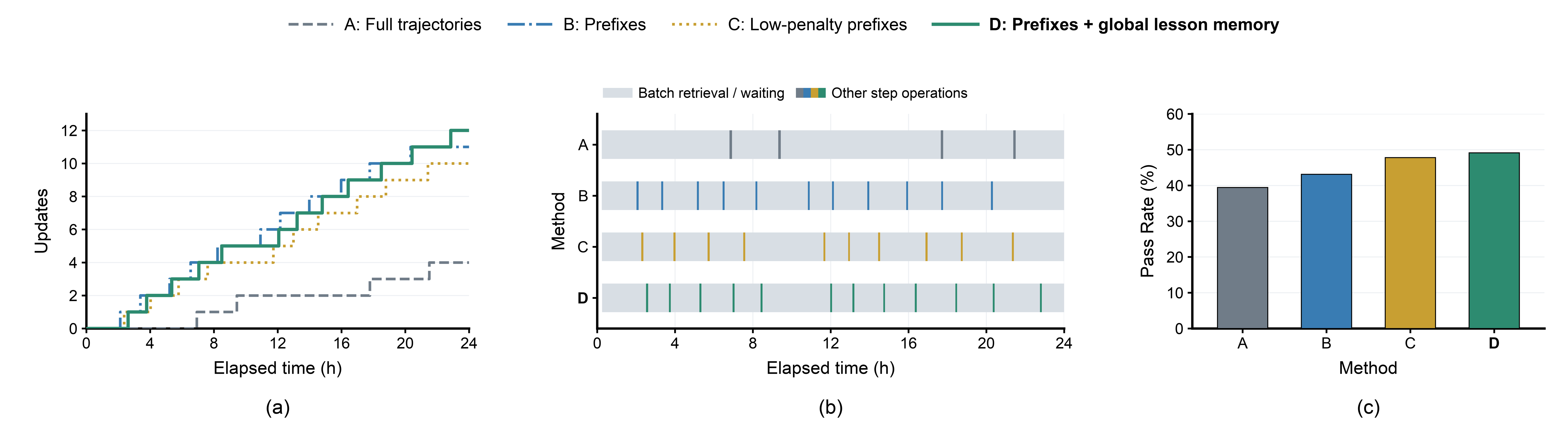}
  \caption{Optimization throughput and candidate validity over 24 hours for methods A--D: (a) cumulative completed parameter updates, (b) learner timelines distinguishing queue waiting (gray) from active gradient steps (colored ticks), and (c) overall candidate correctness pass rate.}
  \label{fig:update-freq}
\end{figure}

Figure~\ref{fig:update-freq} compares four 24-hour DeepSeek-V4-Flash training runs on Ascend: synchronous full-trajectory training (\textbf{A}), asynchronous candidate-prefix training (\textbf{B}), prefix training with reduced failure penalties (\textbf{C}), and prefix training augmented with shared lesson memory (\textbf{D}). Decoupling updates at candidate boundaries triples completed optimizer steps over 24 hours---from \textbf{4 updates in A to 12 updates in D}---while combining prefix learning with reduced failure penalties and shared memory steadily lifts candidate correctness pass rates from \textbf{39.4\% (A) to 49.1\% (D)}.

\begin{figure}[!htbp]
  \centering
  \includegraphics[trim=24bp 12bp 6bp 8bp, clip, width=0.95\linewidth]{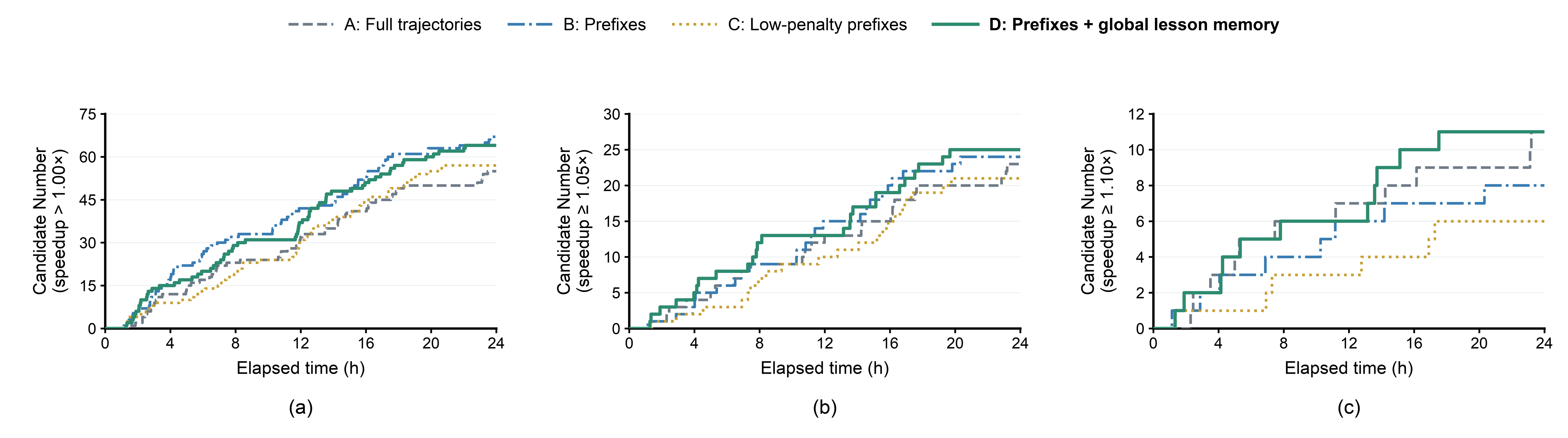}
  \caption{Cumulative discovery of optimized candidates over 24 hours across methods A--D, counting distinct source trajectories that yield a correctness-verified kernel with speedup (a) $>1.00\times$, (b) $\geq 1.05\times$, and (c) $\geq 1.10\times$.}
  \label{fig:opt-candidate-num}
\end{figure}

Figure~\ref{fig:opt-candidate-num} tracks cumulative source trajectories discovering verified speedups across three thresholds during online \texttt{ops-math} search (evaluated via online trajectory discovery; held-out re-testing on \textsc{AscendGYM} is left to future work). On any positive speedup ($>1.00\times$, Panel~a), prefix methods \textbf{B} and \textbf{D} track closely and outpace \textbf{A} and \textbf{C} throughout the 24-hour window. On substantive gains ($\geq 1.05\times$ and $\geq 1.10\times$, Panels~b--c), pure prefix updates alone (\textbf{B}=8, \textbf{C}=6 at $\geq 1.10\times$) lag synchronous full-trajectory \textbf{A} (11) because nested prefixes repeatedly weight early exploratory steps without cross-episode transfer; pairing prefix updates with the evidence-grounded shared lesson memory in \textbf{D} (seeded with historical \texttt{ops-math} optimization findings and updated online) overcomes this local myopia, leading at $\geq 1.05\times$ (25 trajectories) and reaching 11 verified source trajectories at $\geq 1.10\times$ around hour~17.5 ($\approx 6.5$ hours ahead of \textbf{A}).

\subsubsection{System-Level Impact and Outlook}
\label{sec:kernel-or-system-value}

Efficient NPU kernels are essential for high-throughput MoE post-training. In our SuperPOD stack~\citep{slai2026trexpart1}, hybrid parallelism, NPU--CPU overlap, and targeted operator fusion and kernel rewriting jointly increased MFU for DeepSeek-V4-Pro from $27.00\%$ to $34.22\%$, a gain of $7.22$ percentage points. Building on this foundation, this work advances automated NPU kernel optimization to support more efficient large-scale training. Complementing single-turn post-training and RL infrastructure optimization, Section~\ref{sec:or-harness-skills} turns next to test-time agentic OR execution, pairing the base model with a lightweight harness and modular verification skills.

% LaTeX fragment for inclusion in the DeepSeek-V4 technical report.
% Required packages: graphicx, booktabs, tabularx.

\Needspace{10\baselineskip}
\section{Test-Time Agentic Modeling with Modular Domain Skills}
\label{sec:or-harness-skills}

\subsection{Decoupling Minimalist Execution from Domain Modeling Priors}

Whereas Section~\ref{sec:or-bench-post-training} specializes model weights for single-turn formulation and Section~\ref{sec:or-rl-infrastructure} optimizes the underlying execution-in-the-loop RL stack, complex OR problems also benefit at deployment from iterative data inspection, solver execution, and semantic self-verification~\citep{optimus}. We therefore design a lightweight agent harness for OR, inspired by the minimalist philosophy of PI-style coding agents, comprising a compact execution core and a modular OR skill layer. The core manages tool calls, workspace isolation, and program execution; the skills contribute structured domain manuals and deterministic runtime checks.

This separation matches the structure of OR modeling. Because the agent primarily inspects instance data, formulates a mathematical program, writes and executes solver code, and audits the returned solution, a minimal set of file and shell primitives suffices. The central bottleneck is semantic rather than operational: mapping input fields to exact indices and units, encoding every coupling constraint, and distinguishing a faithful formulation from a misspecified program that still terminates with an ``optimal'' solver status. Event-driven skills enforce discipline across these stages without bloating the core action space.

\subsection{Execution-Grounded ReAct Architecture and Solver Verification}

The execution core implements a free-form ReAct loop~\citep{yao2023react}. At each step, the model
reasons about the current state, invokes one or more tools, observes their
outputs, and chooses its next action. Reasoning and tool results remain
interleaved, preserving the link between execution feedback and subsequent
revisions.

\begin{table}[!htbp]
  \centering
  \small
  \caption{Tool interface of the OR harness.}
  \label{tab:or-harness-tools}
  \begin{tabularx}{\linewidth}{@{}p{0.21\linewidth}p{0.32\linewidth}X@{}}
    \toprule
    Category & Tools & Purpose \\
    \midrule
    File operations
      & \texttt{read}, \texttt{write}, \texttt{edit},
        \texttt{glob}, \texttt{grep}
      & Inspect inputs and construct solver artifacts. \\
    Execution
      & \texttt{bash}
      & Run code and solver-backed checks. \\
    Workflow
      & \texttt{todo\_write}, \texttt{finish}
      & Track progress and submit the final artifact. \\
    Extensions
      & \texttt{skill}, \texttt{subagent}
      & Load a skill manual or delegate a bounded subtask. \\
    Interaction
      & \texttt{ask\_question}
      & Request clarification in production; disabled during training. \\
    \bottomrule
  \end{tabularx}
\end{table}

There is no dedicated solver tool. The agent writes \texttt{src/solve.py}
with GurobiPy or Pyomo and executes it through \texttt{bash}. A valid program
prints \texttt{ANSWER: <objective\_value>}, allowing the evaluator to rerun the
same artifact. The runtime enforces a time limit and, when a local Gurobi
license is too restrictive, can export the model to a HiGHS fallback.

Each rollout receives an independent, path-confined workspace, and the shell
blocks selected unsafe operations. These controls improve isolation and
reproducibility but do not constitute a complete operating-system sandbox.

\subsection{Declarative Domain Skills and Runtime Validation Hooks}

Frontier models are strong at coding and pure mathematics, yet OR requires the
two abilities to be combined with precise semantic modeling. Reliable OR
reasoning therefore requires a disciplined workflow that connects data
interpretation, mathematical formulation, solver implementation, and solution
validation.

Skills are self-contained modules within the harness. Each pairs a declarative \texttt{SKILL.md}
domain manual with deterministic Python runtime hooks that attach structured diagnostics to tool outputs or block invalid actions and premature submissions before final artifact handoff.

\begin{table}[!htbp]
  \centering
  \footnotesize
  \caption{The eight modular OR skills evaluated in this section: two benchmark-oriented prior cards (top) and six benchmark-independent workflow verification skills (bottom).}
  \label{tab:or-skills}
  \begin{tabularx}{\linewidth}{@{}p{0.23\linewidth}X@{}}
    \toprule
    Skill & Domain guidance and deterministic runtime checks \\
    \midrule
    \texttt{scene\_cards}
      & Supplies benchmark-tailored scenario patterns, domain conventions, and recurring semantic pitfalls. \\
    \texttt{technique\_cards}
      & Provides benchmark-oriented mathematical formulation templates and associated failure diagnostics. \\
    \midrule
    \texttt{numeric\_probes}
      & Runs solver-independent feasibility, bound, unit-scale, and conservation probes on candidate solutions beyond solver termination status. \\
    \texttt{field\_roles}
      & Audits data-to-model schema mapping from raw input fields to sets, indices, parameters, units, and aggregation rules. \\
    \texttt{code\_checks}
      & Applies static and runtime structural checks on solver code (e.g., variable domains, unreferenced constraints, and objective sense). \\
    \texttt{modeling\_spec}
      & Enforces a written specification of index sets, decision variables, objective, and constraints before \texttt{src/solve.py} is created. \\
    \texttt{gurobi\_landing}
      & Checks fidelity across the formulation-to-GurobiPy boundary, including indexing alignment, big-$M$ scaling, and output formatting. \\
    \texttt{referee}
      & Performs pre-submission cross-verification across the problem statement, mathematical specification, solver script, and numerical output. \\
    \bottomrule
  \end{tabularx}
\end{table}

The six skills other than \texttt{scene\_cards} and
\texttt{technique\_cards} implement reusable stages of a solver-backed OR
workflow: formulation and data semantics, solver implementation and code
validation, and numerical and end-to-end verification. They encode procedures
that recur whenever a problem description is converted into an executable
mathematical program, rather than templates tied to a particular benchmark.

Skills are activated by execution events rather than a rigid workflow. For
example, \texttt{modeling\_spec} requires a formulation before
\texttt{solve.py} is written; \texttt{numeric\_probes} and
\texttt{code\_checks} run after solver execution; and \texttt{referee}
participates in final submission checks. The harness thus combines a small
action space with stage-specific OR guidance and verification.

\subsection{Multi-Rollout Empirical Evaluation and Skill Ablations}

We evaluate the Base model (DeepSeek-V4-Flash-0731) on \textsc{SLAI-OR-Bench-Hard} by drawing five independent agent rollouts for each of the 93 problems and reporting solve accuracy across all 465 rollouts (note that our Section~\ref{sec:or-bench-post-training} post-trained checkpoint targets single-turn formulation generation rather than multi-turn tool-use schemas; bridging the two routes is addressed in Section~\ref{sec:skill-internalization}). Each rollout is bounded by 30 agent steps, 7,200 seconds of wall-clock execution/reasoning time, and a 64K-token context window, with timeouts counted as failures. External CLI coding harnesses (Claude Code and Codex) execute in their native tool loops under single-run evaluation on the 93 Hard tasks within the same time and context caps, without our internal Gurobi-to-HiGHS export fallback or OR skill hooks.

\subsubsection{End-to-End Impact of Harness Execution and Domain Skills}

\begin{figure}[!htbp]
  \centering
  \includegraphics[width=0.86\linewidth]{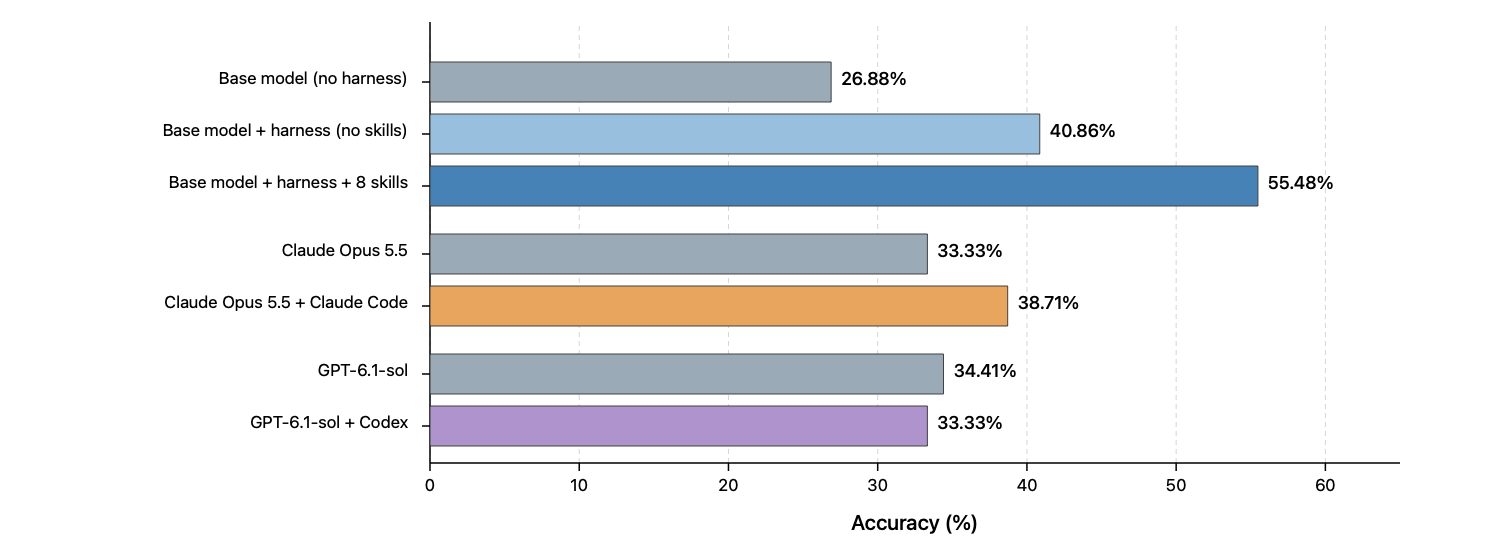}
  \caption{Effects of harness configuration on accuracy across the 93 tasks of \textsc{SLAI-OR-Bench-Hard} (five rollouts per task). Upper bars compare single-turn Base inference, the lightweight harness alone, and the harness with all eight evaluated OR skills; lower bars compare two external coding harnesses on their respective frontier models.}
  \label{fig:harness-effects}
\end{figure}

As shown in Figure~\ref{fig:harness-effects}, equipping the Base model (26.88\% single-turn accuracy) with the lightweight execution harness alone raises accuracy to \textbf{40.86\%} ($+13.98$ percentage points); enabling the six benchmark-independent workflow skills reaches \textbf{49.46\%} ($+22.58$ points; Figure~\ref{fig:or-skill-comparison}), and adding the two benchmark-tailored prior cards reaches \textbf{55.48\%} ($+28.60$ points). Keeping the core action space minimal while encapsulating domain discipline in event-driven skills thus converts multi-turn interaction directly into modeling reliability.

By contrast, general-purpose coding harnesses yield mixed results: Claude Code~\citep{anthropic2026claudecode} lifts Claude Opus~5.5 from 33.33\% to 38.71\% ($+5.38$ points), whereas Codex~\citep{openai_codex_cli} slightly lowers GPT-6.1-Sol from 34.41\% to 33.33\% ($-1.08$ points). Without OR-specific semantic checks, generic shell iteration can fix syntax errors while leaving silent formulation mismatches intact---and additional generic scaffolding can introduce context overhead when misaligned with the underlying model.

\subsubsection{Individual Skills, Generalizable Priors, and Test-Time Scaling}

\begin{figure}[!htbp]
  \centering
  \includegraphics[width=0.84\linewidth]{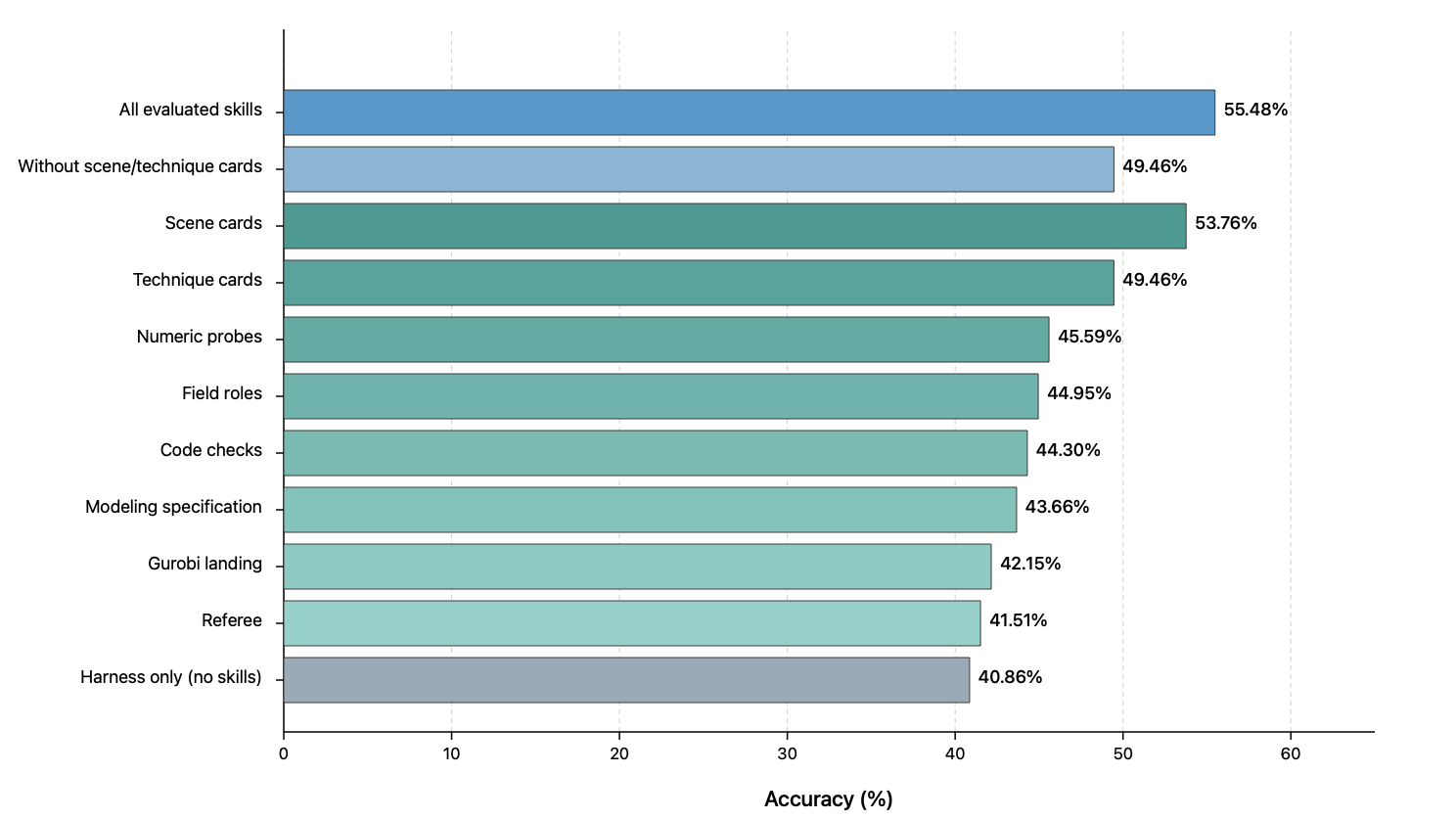}
  \caption{Single-skill and combined-skill ablations across the 93 tasks of \textsc{SLAI-OR-Bench-Hard} (five rollouts per task). ``Without scene/technique cards'' enables the six reusable benchmark-independent workflow skills (49.46\%), whereas ``All evaluated skills'' enables all eight skills (55.48\%).}
  \label{fig:or-skill-comparison}
\end{figure}

Figure~\ref{fig:or-skill-comparison} ablates individual and combined skills over the 40.86\% harness-only baseline. Every evaluated skill maintains or improves accuracy over harness-only---from \texttt{referee} at 41.51\% ($+0.65$ points, neutral-to-slightly positive within rollout variance) up to \texttt{numeric\_probes} at 45.59\% ($+4.73$ points) among standalone workflow checks, and 49.46\% (\texttt{technique\_cards}) and 53.76\% (\texttt{scene\_cards}) with benchmark-oriented priors---confirming that gains accumulate across complementary verification stages rather than hinging on a single rule.

Crucially, because \texttt{scene\_cards} and \texttt{technique\_cards} encode scenario and formulation priors tailored to patterns represented in this benchmark to probe attainable upper-bound performance, cross-benchmark generalization is best measured by removing both cards. Enabling only the \textbf{six benchmark-independent workflow skills} achieves \textbf{49.46\%} accuracy ($+22.58$ points over single-turn Base and $+8.60$ points over harness-only), outperforming Claude Opus~5.5 with Claude Code (38.71\%) by $+10.75$ points. Viewed through the lens of test-time scaling, single-submission multi-turn verification with these six general-purpose skills (49.46\%) surpasses eight-sample single-turn Base coverage ($\mathrm{pass@8}=46.24\%$) and approaches post-trained SFT+RL $\mathrm{pass@8}$ (51.61\%) \emph{without} requiring an external ground-truth oracle to select among candidate solutions.

\subsection{Toward Parameter Internalization of Skill-Guided Trajectories}
\label{sec:skill-internalization}

While Section~\ref{sec:or-bench-post-training} demonstrates parameter specialization for single-turn solving and this section establishes verified multi-turn execution on the Base model, our next milestone unifies both tracks. Specifically, we plan to distill verified, skill-assisted multi-turn trajectories back into model parameters via agentic SFT followed by execution-in-the-loop RL---leveraging the trajectory-reuse, numerical-alignment, and speculative-decoding infrastructure of Section~\ref{sec:or-rl-infrastructure}---so that \model internalizes disciplined formulation and self-verification habits while retaining lightweight deterministic runtime checks at deployment.

\section{Conclusion and Future Directions}
\label{sec:conclusion}

This report presented \model, a full-stack operations research post-training and agentic reasoning platform built on DeepSeek-V4-Flash and Huawei Ascend SuperPOD. Starting from an eight-sample diagnosis showing that 53.76\% of \textsc{SLAI-OR-Bench-Hard} tasks remain unsolved by the base model, we connected a four-layer OR taxonomy to one-hop solver-verified data synthesis, solve-rate routing, cold-start SFT, and sequential Chat/Thinking GRPO with graded failure rewards and resource-limit outcome masking. On our expert-curated \textsc{SLAI-OR-Bench} (131 tasks) and \textsc{SLAI-OR-Bench-Hard} (93 tasks) benchmarks, post-training raised single-turn accuracy from 38.26\% to \textbf{50.86\%} ($+5.82$ and $+6.59$ points over GPT-6.1-Sol and Claude Opus~5.5) and from 26.88\% to \textbf{35.89\%} (slightly surpassing both single-draw frontier baselines), while expanding eight-sample solve coverage from 46.24\% to \textbf{51.61\%}. Along our complementary test-time agentic route, pairing the base model with a minimalist ReAct harness raised \textsc{SLAI-OR-Bench-Hard} accuracy to \textbf{40.86\%}, reaching \textbf{49.46\%} with six benchmark-independent general-purpose OR skills ($+22.58$ percentage points over single-turn base inference) and \textbf{55.48\%} ($+28.60$ points) when adding two benchmark-tailored prior cards.

On the systems side, four controlled investigations on Ascend NPUs establish practical design principles for execution-in-the-loop MoE reinforcement learning. First, expensive solver-verified rollouts and profiler-measured kernel episodes can be exploited far more efficiently without stalling behind fresh full-episode generation: bounded same-batch trajectory reuse with recalibrated advantages ($R_{\max}=4$) cuts newly generated response tokens by \textbf{82.5\%} over 80 updates while preserving task accuracy, and asynchronous candidate-prefix RL with shared lesson memory triples completed parameter updates ($12$ versus $4$ in 24 hours), lifts candidate pass rates from \textbf{39.4\% to 49.1\%}, and accelerates the discovery of $\geq 1.05\times$ and $\geq 1.10\times$ NPU kernel speedups. Second, progressive layer-wise cosine scanning and recursive same-input replay across 64 Ascend 910C NPUs isolate operator-level discrepancies (SwiGLU clipping and MoE router-bias precision mismatches) from downstream routing amplification, preventing RL training collapse. Third, our OR--MTP co-adaptation studies show that detaching shared trunk features during RL followed by frozen-target private MTP refinement raises three-token OR development agreement from \textbf{48.44\% to 60.16\%} without degrading solver accuracy, while enabling saved native MTP modules delivers \textbf{$1.583\times$--$1.715\times$} offline OR token throughput at client concurrency one.

Three directions define our immediate roadmap: (i) extending recalibrated trajectory reuse over multi-hundred-step horizons with adaptive distribution-drift controls; (ii) integrating lossless native MTP verification directly into high-concurrency online RL rollout workers to translate decoding speedups into end-to-end training wall-clock reductions; and (iii) closing the loop between our two capability routes by distilling verified, skill-assisted multi-turn trajectories into \model via agentic SFT and RL while retaining lightweight deterministic verification hooks at inference time.

\FloatBarrier
\vspace{0.8em}
\phantomsection
\markboth{References}{References}
\bibliographystyle{slai}
\bibliography{custom}

\clearpage
\appendix
\section{Author List}
\label{app:author-list}

\noindent\textit{Authors are grouped by contribution role into Core Contributors, Contributors, and Project Leaders, and ordered alphabetically by surname within Core Contributors and Contributors.}

\subsection*{Core Contributors}
Xuhui~Chen, Sihang~Chen, Tianxiang~Fang, Wenlong~Jin, Weiyan~Kong, Jiajun~Liu, Zhengxuan~Lu, Zhuohan~Wang, Jinmin~Xiang, Yucheng~Xie, Mingzhe~Yang, Zirong~Zeng, Bingwei~Zhang, Ziwei~Zhu

\subsection*{Contributors}
Xin~Chen, Yuntian~Fan, Yingmeng~Gao, Dongfang~Li, Jing~Li, Liangyou~Li, Xiangyu~Li, Yizhou~Liang, Xiaodong~Luo, Linyuan~Qiu, Ruoyu~Sun, Weijian~Sun, Yiting~Wang, Chenwei~Wu, Zhengjun~Yue

\subsection*{Project Leaders}
Baotian~Hu, Min~Zhang, Haizhou~Li, Zhiquan~Luo

\Needspace{12\baselineskip}
\section{Provenance and Curation of Public AI4OR Benchmarks}
\label{app:existing-benchmarks}

\subsection{Evaluated Public Benchmark Suite}
\label{app:public-benchmark-suite}

Our evaluation includes
NL4OPT~\citep{ramamonjison2022nl4opt},
OptiBench~\citep{yang2025optibench},
Bench4Opt~\citep{wang2025orgeval},
ComplexOR~\citep{xiao2024chainofexperts},
IndustryOR~\citep{huang2025orlm},
MAMO-ComplexLP and MAMO-EasyLP~\citep{huang2025mathematical},
and NLP4LP~\citep{optimus}.
We additionally include problems from the ICML 2024 Automated
Optimization competition, described by
Nguyen and Ko~\citep{nguyen2024icmlreport}.

\subsection{Release Provenance and Instance Selection Criteria}
\label{app:dataset-selection}

We use curated releases associated with
SAGE~\citep{zhao2026sage} for NL4OPT, OptiBench,
MAMO-EasyLP, and MAMO-ComplexLP.
Table~\ref{tab:dataset-versions} distinguishes the original
dataset sizes from the numbers of instances used in our
evaluation.

Within our source collection, we select the parseable candidate release with the most stringent community cleaning (and fewest retained records when versions differ), preferring explicitly audited splits on ties. Following the cleaning taxonomy of \citet{xiao2025survey}---which manually inspected benchmark statements with 11 domain experts and cross-validated every flagged logical inconsistency, missing parameter, or erroneous ground-truth objective across at least three experts---and subsequent refinements in StepORLM~\citep{zhou2025steporlm}, we adopt these upstream curated evaluation splits directly (recording exact file checksums in our evaluation harness; direct cross-paper comparisons should match both the evaluated split and solver tolerance).

\begin{table}[!htbp]
\centering
\small
\renewcommand{\arraystretch}{1.10}
\caption{Original and evaluated dataset sizes.
Original sizes follow
\citet{ramamonjison2022nl4opt},
\citet{yang2025optibench}, and
\citet{huang2025mathematical}.
For NL4OPT, the evaluated set excludes 11 instances from
the 213-instance curated release.}
\label{tab:dataset-versions}
\begin{tabular*}{\linewidth}{@{\extracolsep{\fill}}lrr@{}}
\toprule
\textbf{Dataset} & \textbf{Original} & \textbf{Evaluated} \\
\midrule
NL4OPT         & 289 & 202 \\
OptiBench      & 605 & 403 \\
MAMO-EasyLP    & 652 & 545 \\
MAMO-ComplexLP & 211 & 111 \\
\bottomrule
\end{tabular*}
\end{table}

\Needspace{14\baselineskip}
\section{Qualitative Formulation Case Study on SLAI-OR-Bench-Hard}
\label{app:representative_examples}

We present the optical-network routing and resource optimization
problem (\textsc{SLAI-OR-Bench-Hard} \#17). The following blocks reproduce the
problem description supplied to the LLM and the reference
mathematical formulation.

\Needspace{10\baselineskip}
\subsection*{Original problem description}

The wording and numerical data below are unchanged; table markup
is converted to plain text for readability.

\begin{lstlisting}[style=benchmarkexample, basicstyle=\ttfamily\scriptsize, aboveskip=5pt, belowskip=6pt]
A telecom carrier operates an optical fiber network with 6 nodes and 9 bidirectional links. Each fiber supports 4 wavelength channels (W1--W4), each with 100 Gbps capacity. 8 traffic demands must each be routed on a single (path, wavelength) lightpath. Multiple demands may share a wavelength on a link if total used bandwidth <= 100 Gbps (traffic grooming). The lightpath uses the same wavelength on all its hops (wavelength continuity, no regeneration).

Network links (distance in km):

Link    Endpoints    Distance
L1      1--2         200
L2      1--3         350
L3      2--3         180
L4      2--4         250
L5      3--4         300
L6      3--5         220
L7      4--5         160
L8      4--6         290
L9      5--6         210

Demands and pre-computed candidate paths (up to 3 paths per demand, max 4 hops):

Demand  s -> t  Gbps  Path 1 (km)     Path 2 (km)       Path 3 (km)
d1      1 -> 6   40   1-3-5-6 (780)  1-2-4-6 (740)    1-3-4-6 (940)
d2      1 -> 5   80   1-3-5 (570)    1-2-3-5 (600)    1-2-4-5 (610)
d3      2 -> 6   60   2-4-6 (540)    2-3-5-6 (610)    2-3-4-6 (770)
d4      2 -> 5   30   2-3-5 (400)    2-4-5 (410)      2-3-4-5 (640)
d5      3 -> 6   50   3-5-6 (430)    3-4-6 (590)      3-4-5-6 (670)
d6      1 -> 4   90   1-2-4 (450)    1-3-4 (650)      1-2-3-4 (680)
d7      3 -> 4   70   3-4 (300)      3-5-4 (380)      3-2-4 (430)
d8      2 -> 4   40   2-4 (250)      2-3-4 (480)      2-3-5-4 (560)

Each demand is routed on exactly one (path, wavelength) combination. For each link l and wavelength w, the sum of bandwidth requirements of all demands whose chosen path uses link l and whose chosen wavelength is w must not exceed 100 Gbps. The objective is to minimise total wavelength-kilometres consumed.
\end{lstlisting}

\Needspace{12\baselineskip}
\subsection*{Reference mathematical formulation}

\begin{tcolorbox}[colback=black!2, colframe=black!25, boxrule=0.5pt, arc=2pt, left=8pt, right=8pt, top=6pt, bottom=6pt, breakable]
\small
Let $D$ be the set of demands, $P_d$ the set of pre-computed candidate paths for demand $d \in D$, $W$ the set of wavelengths, and $L$ the set of network links. Let $b_d$ denote the bandwidth requirement of demand $d$, $c_l$ the length of link $l$, and let $\delta_{dpl}=1$ if candidate path $p \in P_d$ uses link $l$, and $0$ otherwise. Decision variables are defined as follows:
\[
x_{dpw} \in \{0,1\} \qquad \forall d\in D,\ p\in P_d,\ w\in W,
\]
where $x_{dpw}=1$ if demand $d$ is assigned to candidate path $p \in P_d$ and wavelength $w$, and
\[
u_{lw} \in \{0,1\} \qquad \forall l\in L,\ w\in W,
\]
where $u_{lw}=1$ if wavelength $w$ is used on link $l$ by at least one selected demand path.

\medskip\noindent
\textbf{Objective:} Minimize total wavelength-kilometres consumed across all activated link--wavelength pairs:
\[
\min \quad \sum_{l\in L}\sum_{w\in W} c_l u_{lw}.
\]

\noindent
\textbf{Constraints:} Each demand must select exactly one candidate path and one wavelength:
\[
\sum_{p\in P_d}\sum_{w\in W} x_{dpw}=1 \qquad \forall d\in D.
\]
The aggregate bandwidth assigned to each link--wavelength pair cannot exceed the 100\,Gbps channel capacity:
\[
\sum_{d\in D}\sum_{p\in P_d} b_d\,\delta_{dpl}\,x_{dpw}\le 100 \qquad \forall l\in L,\ w\in W.
\]
A link--wavelength pair is activated whenever any selected demand path using that link is assigned the wavelength:
\[
u_{lw}\ge x_{dpw} \qquad \forall d\in D,\ p\in P_d,\ l\in L,\ w\in W\text{ such that }\delta_{dpl}=1.
\]
\end{tcolorbox}

\Needspace{6\baselineskip}
\subsection*{Modeling challenge}
The task requires distinguishing the routes assigned to individual
demands from the shared network resources those routes occupy.
Choosing one path--wavelength combination per demand ensures
wavelength continuity, while bandwidth constraints determine
whether multiple demands can share a resource. The objective
requires a further distinction: occupied wavelength-kilometres
measure the link--wavelength resources used, rather than the sum
of individual demand path lengths.

For example, demands d1 and d3 require 40 and 60 Gbps, respectively.
If their selected routes both traverse the 290-km link L8 using
the same wavelength, their combined bandwidth fits within its
100-Gbps capacity. That shared link--wavelength resource contributes
290 wavelength-km to the reference objective, irrespective of the
two demands using it. Summing the distance separately for each
demand would instead contribute 580. The activation variable
$u_{lw}$ captures this distinction.

\Needspace{6\baselineskip}
\subsection*{Observed model results}
DeepSeek-V4-Flash-0731 and Claude Opus 5 both introduced resource
activation variables and matched the reference optimum of
2,820 wavelength-km. GPT-5.6-Sol encoded demand assignment,
wavelength continuity, and bandwidth capacity, but minimized the
sum of individual demand path lengths. Its actual objective
implementation was:

\begin{lstlisting}[
  style=benchmarkexample,
  language=Python
]
model.setObjective(
    gp.quicksum(
        path_length[d, p] * x[d, p, w]
        for d in demands
        for p in paths[d]
        for w in wavelengths
    ),
    GRB.MINIMIZE,
)
\end{lstlisting}

The program solved to optimality and returned 3,680 under this
different objective, counting shared resources separately for
each demand. This was an objective-formulation mismatch, rather
than a code execution error or an unfinished solve. The two
objective values correspond to different metrics and therefore
should not be interpreted as an optimality gap under the reference
objective. The example illustrates the domain interpretation
required to translate ``wavelength-kilometres consumed'' into
shared-resource accounting.

\Needspace{10\baselineskip}
\section{Formulation and Diagnostics of Zero-Advantage GRPO Variants}
\label{app:grpo-variants}

This appendix provides the detailed formulations, mechanism diagrams, training dynamics, and diagnostic analyses for the Base-initialized Chat-GRPO zero-advantage interventions summarized in Section~\ref{sec:pt-variants}.

\subsection{Budget Reallocation, Off-Policy Replay, and Virtual Rewards}
\begin{figure}[!htbp]
 \centering
 \includegraphics[width=0.88\linewidth]{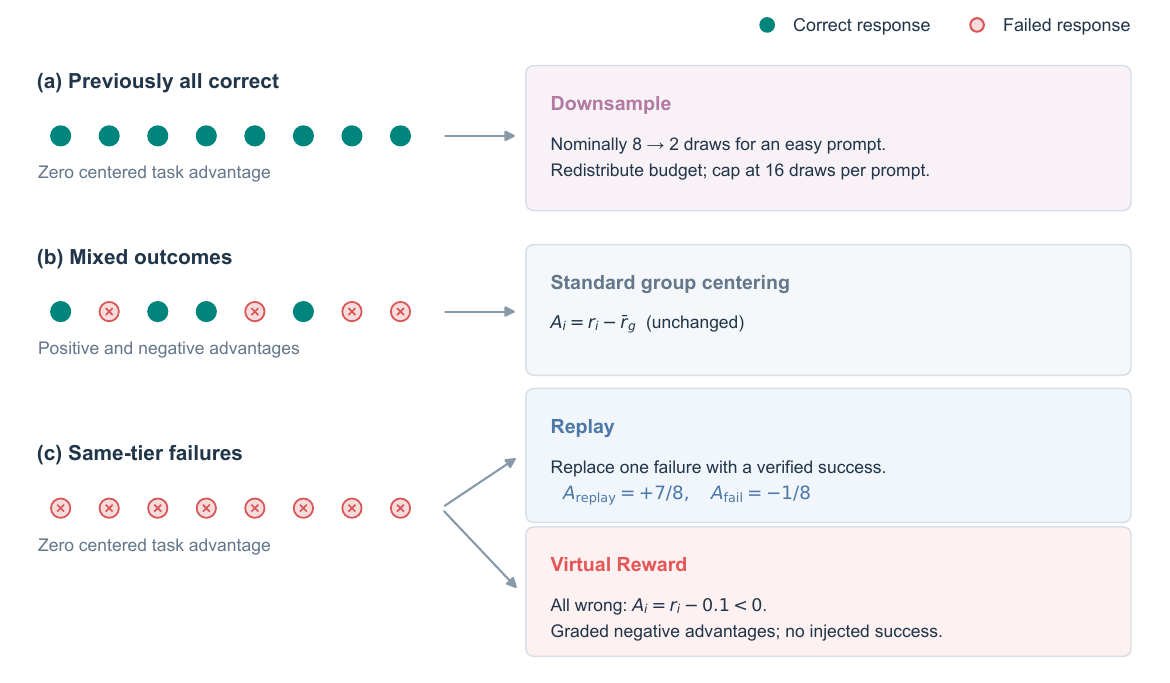}
 \caption{Mechanisms for handling zero-advantage response groups. (a) Downsample reduces the nominal allocation for prompts that were all correct at the previous visit. (b) Mixed correct/failed groups retain ordinary mean-centered advantages. (c) Replay introduces a verified success, whereas Virtual Reward changes the all-wrong-group baseline. Each illustration contains eight valid responses; Replay values assume one reward-1 response and seven reward-0 failures.}
 \label{fig:grpo-variants-design}
\end{figure}

All four runs in this study initialize directly from DeepSeek-V4-Flash-0731 (Base), without cold-start SFT, and run Chat generation for 200 updates on the same training prompt pool. Each rollout batch contains 32 prompts and a nominal budget of eight responses per prompt (256 responses in total). The runs share solver-based Failure Reward Tiers and Outcome Masking (Table~\ref{tab:pt-reward-mask}), learning rate $10^{-6}$, one optimizer update per rollout batch, PPO clipping at $0.2$, and a separate reference-policy KL loss with coefficient $0.001$; advantages are not divided by the within-group standard deviation~\citep{v4_liu2025r1zero}. Greedy validation is evaluated every ten updates on the 177 seed problems represented through two modeling interfaces (354 records). Figure~\ref{fig:grpo-variants-design} illustrates the three mechanisms relative to Equation~\eqref{eq:grpo-variants-baseline}, and Table~\ref{tab:grpo-variants-design} positions them alongside related off-policy and budget-allocation approaches.

\begin{table}[!htbp]
 \centering\small
 \setlength{\tabcolsep}{5pt}
 \renewcommand{\arraystretch}{1.12}
 \caption{Exploratory GRPO interventions and their relation to nearby approaches.}
 \label{tab:grpo-variants-design}
 \begin{tabularx}{\linewidth}{@{}L{0.13\linewidth}L{0.31\linewidth}Y@{}}
 \toprule
 \textbf{Variant} & \textbf{Target and intervention} & \textbf{Relation to nearby approaches}\\
 \midrule
 Downsample & Previously all-correct prompts: reduce the nominal allocation from eight responses to two and redistribute the saved budget, with a per-prompt cap of 16. & DAPO and GRESO filter or skip uninformative groups~\citep{dapo,zheng2025greso}; Knapsack RL optimizes rollout allocation~\citep{li2025knapsackrl}. This heuristic uses the previous visit, preserves the batch budget, and does not estimate variance.\\[4pt]
 Replay & All-wrong groups: replace one failed response with a verified successful trajectory for the same prompt. & LUFFY uses off-policy guidance~\citep{yan2025luffy}; RLEP and ExGRPO mix stored experience~\citep{zhang2025rlep,v4_zhan2026exgrpo}. This variant injects one stored success specifically into an all-wrong group, without a behavior-policy importance correction.\\[4pt]
 Virtual Reward & All-wrong groups: replace the group-mean baseline with $c=0.1$, making every failure advantage negative. & GRPO and Dr.~GRPO retain a group-mean baseline~\citep{shao2024deepseekmath,v4_liu2025r1zero}. Unlike negative reinforcement applied to failed samples generally~\citep{zhu2025nsr}, this intervention acts only on all-wrong groups and preserves graded failure severity.\\
 \bottomrule
 \end{tabularx}
\end{table}

\paragraph*{Downsample.}
For a 32-prompt batch $P_t$ with previously all-correct prompts $E_t$, Downsample nominally retains two verification draws per prompt in $E_t$ and distributes saved slots round-robin across remaining prompts up to a cap of 16:
\begin{equation}
 \sum_{p\in P_t}n_p=256,\qquad 2\leq n_p\leq16,
 \qquad n_p^{\mathrm{nominal}}=2\quad(p\in E_t),
 \label{eq:grpo-variants-downsample}
\end{equation}
adjusting $n_p$ when non-easy prompts hit the cap so the batch total remains 256.

\paragraph*{Replay.}
A trajectory library $\mathcal C$ stores one verified correct trajectory per training prompt, updating the entry whenever a newly sampled correct trajectory achieves higher length-normalized log-likelihood under the current policy. When a group produces no correct response, Replay identifies the most frequent failure tier and replaces its longest trajectory $y_{i^\star}$ with the stored success $y_g^{\mathcal C}$ (assigned score $1$), yielding
\begin{equation}
 \begin{aligned}
 \bar r'_g&=\frac{1}{G_g}\left(1+\sum_{\substack{j\in I_g\\j\ne i^\star}}r_j\right),\\
 A_{i^\star}&=1-\bar r'_g,\qquad
 A_j=r_j-\bar r'_g\quad(j\ne i^\star).
 \end{aligned}
 \label{eq:grpo-variants-replay}
\end{equation}
Because the injected sequence's reference log-probability is recomputed under the current policy, its initial PPO ratio is $1$ without behavior-policy importance correction, distinguishing Replay from regularized off-policy guidance~\citep{yan2025luffy} and from the same-batch recalibration in Section~\ref{sec:v4-reuse}.

\paragraph*{Virtual Reward.}
Virtual Reward shifts the baseline exclusively on groups without a correct response:
\begin{equation}
 b_g=
 \begin{cases}
  \bar r_g,&\max_{j\in I_g}r_j>0,\\
  c,&\max_{j\in I_g}r_j\leq0,
 \end{cases}
 \qquad c=0.1,
 \label{eq:grpo-variants-virtual}
\end{equation}
so failure tiers $0$, $-0.1$, and $-0.15$ receive strictly negative advantages $-0.1$, $-0.2$, and $-0.25$. In the idealized special case of fixed group size $G$, independent on-policy draws, and binary rewards with success probability $p=p_\theta$, the extra all-wrong-group policy-gradient term satisfies
\begin{equation}
 \begin{aligned}
 &\mathbb E\!\left[-c\,\mathbf 1\{\text{all }G\text{ fail}\}
       \sum_{i=1}^{G}\nabla_\theta\log\pi_\theta(y_i\mid x_g)\right]\\
 &\qquad=-c\,\nabla_\theta(1-p)^G
   =cG(1-p)^{G-1}\nabla_\theta p
   =c\,\nabla_\theta\mathrm{pass@}G,
 \end{aligned}
 \label{eq:grpo-variants-passg}
\end{equation}
which encourages early sampling coverage when $p$ is small, though variable masks, PPO clipping, token averaging, and a non-decaying constant $c$ break exact equivalence to $\mathrm{pass@}G$ maximization.

\subsection{Training Dynamics and Process-Level Diagnostics}

\begin{figure}[!htbp]
 \centering
 \includegraphics[width=0.88\linewidth]{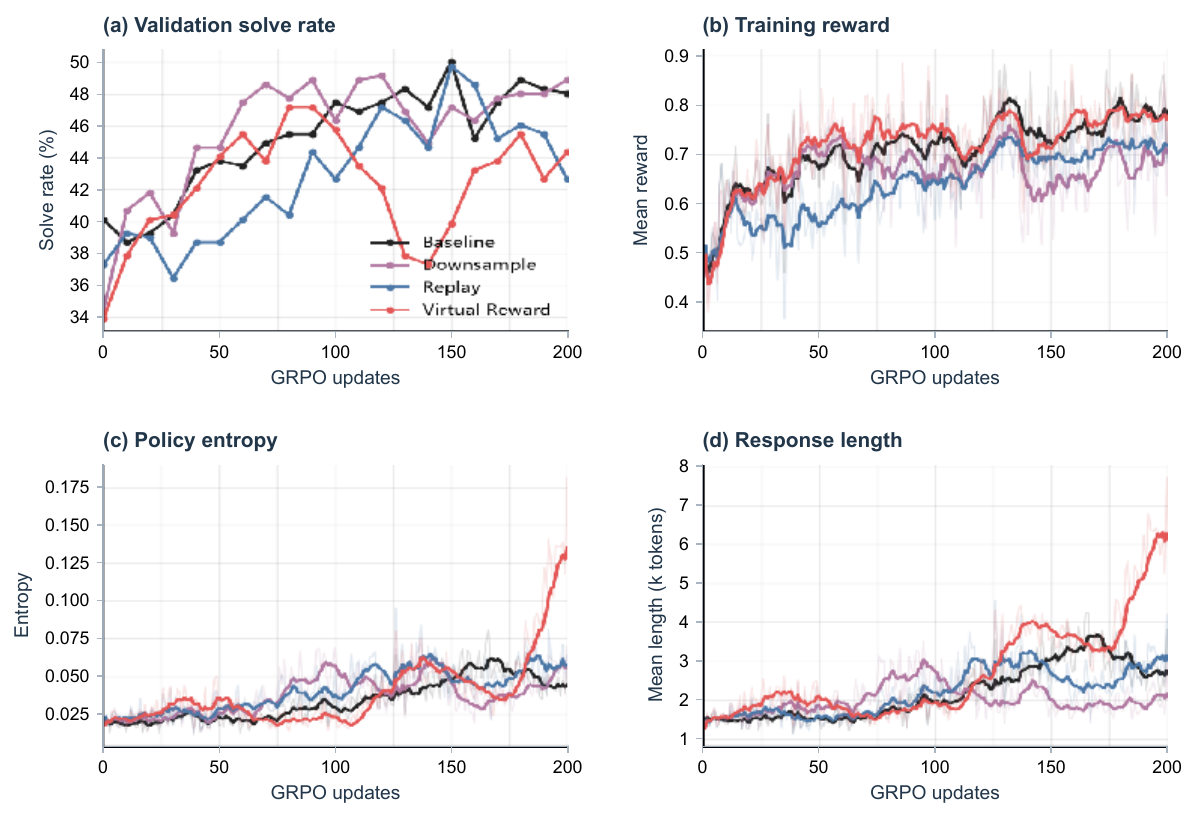}
 \caption{Training dynamics of the Base-initialized Chat-GRPO variants. (a) Greedy validation solve rate at ten-update intervals. (b--d) Training reward, policy entropy, and response length (faint lines: raw values; solid lines: centered seven-update moving averages).}
 \label{fig:grpo-variants-outcomes}
\end{figure}

\begin{figure}[!htbp]
 \centering
 \includegraphics[width=0.88\linewidth]{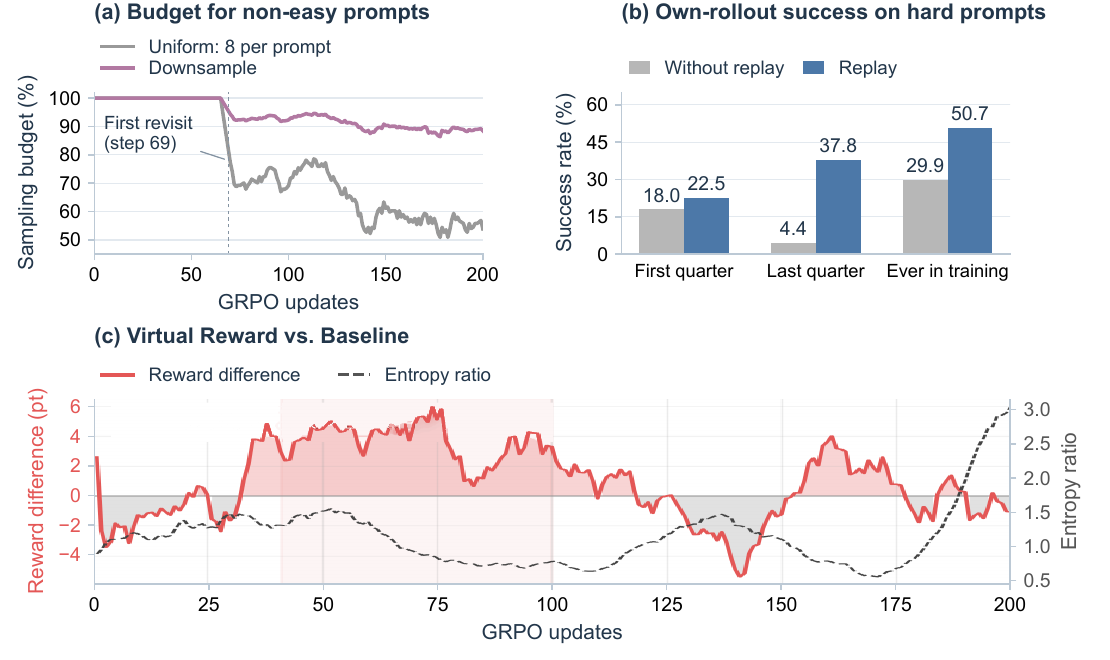}
 \caption{Process-level effects of the GRPO variants. (a) Non-easy-prompt budget share under Downsample versus uniform allocation. (b) Autonomous rollout success on hard training prompts under Replay versus Baseline (excluding injected trajectories). (c) Virtual Reward minus Baseline training reward ($\times 100$, left axis) and entropy ratio (right axis).}
 \label{fig:grpo-variants-mechanisms}
\end{figure}

Figure~\ref{fig:grpo-variants-outcomes} plots validation solve rates, training rewards, entropy, and response lengths across the 200 updates, while Figure~\ref{fig:grpo-variants-mechanisms} verifies that each mechanism alters training dynamics as designed:
\begin{itemize}[leftmargin=1.8em, itemsep=3pt]
 \item \textbf{Budget reallocation under Downsample:} Easy-prompt labels emerge at update 69 as prompts are revisited, averaging 11.5 prompts per batch over updates 69--200 (14.5 in the last ten updates). Non-easy prompts receive 90.9\% of active rollout slots versus 64.2\% under uniform allocation, with a mean per-batch maximum of 12.1 draws.
 \item \textbf{Autonomous hard-prompt gains under Replay:} Replay injects a stored trajectory into 450 of 453 encountered all-wrong groups (2.25 per update). Excluding injected responses, the policy's own rollout success rate on a preselected subset of hard training prompts reaches 37.8\% in the final quarter (versus 4.4\% for Baseline), solving 50.7\% at least once during training (versus 29.9\%).
 \item \textbf{Early process lift under Virtual Reward:} Mean training reward over updates 41--100 reaches 0.739 versus 0.703 for Baseline, and validation reaches 47.18\% at updates 80--90 (versus 45.48\% for Baseline) before late-stage entropy triples and mean output length grows to 6,247 tokens.
\end{itemize}

\paragraph*{Why process gains do not yield superior held-out validation.}
Five factors account for the gap between training-process shifts and peak greedy validation accuracy (Table~\ref{tab:grpo-variants-results}):
(1)~\emph{Evaluation granularity and distributed runtime variance:} the 354 validation records span 177 distinct problems ($\approx 0.28$ percentage points per record), and independent distributed launches exhibit Update-0 greedy evaluation variance across runs (33.90\%--40.11\% on the same initial checkpoint due to non-deterministic parallel reductions and discrete MoE Top-6 routing boundary flips prior to the Section~\ref{sec:train-infer-consistency} consistency repairs), limiting sensitivity to sub-point peak differences;
(2)~\emph{Bounded intervention scope:} Replay acts on an average of 2.25 of 32 prompts per update, while Downsample activates only after prompt revisits begin at update 69 (and its lower mean training reward of 0.701 reflects fewer draws on easy prompts rather than policy degradation);
(3)~\emph{Limited out-of-prompt transfer in Replay:} boosting likelihood on uncorrected historical trajectories improves same-prompt autonomous success during training but risks late-stage overfitting (dropping from 49.72\% at update 150 to 42.66\% at update 200);
(4)~\emph{Unannealed negative drift and token-averaged length bias in Virtual Reward:} keeping $c=0.1$ active without a positive target pushes probability mass away from sampled failures indiscriminately once early gains saturate, while dividing sequence-level negative advantages by response length $L_i$ dilutes per-token penalties on longer completions~\citep{v4_liu2025r1zero}, driving entropy and $2.3\times$ output-length inflation (6,247 versus 2,729 tokens); and
(5)~\emph{Sampling-versus-greedy divergence:} Virtual Reward's late training reward (0.775 at temperature 1) rises even as greedy validation declines, reflecting shifts across failure tiers and multi-sample coverage rather than greedy single-attempt precision.
\FloatBarrier
\Needspace{10\baselineskip}
\section{Extended Ablations and Controls for OR--MTP Co-Adaptation}
\label{app:sdv2}

\subsection{Archived Endpoint Trajectory and Motivation}
\begin{table}[!htbp]
\centering\small
\renewcommand{\arraystretch}{1.10}
\caption{Archived endpoint comparison supporting the training motivation. OR raw/H1 counts report the earlier audited public1627 scores, with one unresolved license case per trained endpoint. Acceptance is pooled over the separate historical 1,344-request concurrency-1 greedy panel. Table~\ref{tab:sdv2-quality} reports a separate response batch for four trained targets and excludes Full Joint CE.}
\label{tab:sdv2-history}
\begin{tabular*}{\linewidth}{@{\extracolsep{\fill}}lrrr@{}}
\toprule
\textbf{Historical endpoint} & \textbf{Raw / 1,627} & \textbf{H1 / 1,627} & \textbf{Pooled AR (\%)} \\
\midrule
CPT ref. & 969 & 978 & 24.27 \\
RL only & 1072 & 1114 & 23.03 \\
Joint CE & 989 & 1012 & 24.44 \\
Feature detach & 1111 & 1111 & 22.36 \\
G2 & 966 & 1059 & 23.03 \\
G4 & 1023 & 1061 & 23.61 \\
\bottomrule
\end{tabular*}

\end{table}

The historical CPT reference serves as the initialization for this earlier training series rather than the 0731 base model studied in Section~\ref{sec:or-bench-post-training}, and training recipes as well as recorded seed provenance differ across the archived endpoints. The small gains of Full Joint CE over the CPT reference in raw score and acceptance do not establish a stable joint-training advantage over target-only RL, nor do either the archived endpoints or the fresh four-target comparison isolate the causal effect of the gating variants.

In these joint runs, the auxiliary cross-entropy (CE) coefficient is $0.3$: policy-gradient and KL terms use valid-token means, whereas auxiliary CE uses a per-response mean followed by a batch mean. Because global gradient clipping and Adam moment updates operate across parameters, training dynamics can remain coupled even when an auxiliary activation path is detached; preliminary partial-parameter gradient diagnostics do not reflect full data-parallel-averaged optimizer updates and are therefore omitted from causal attribution.

\subsection{Paired Quality Contrasts and Protocol Bridges}
\begin{table}[!htbp]
\centering\small
\renewcommand{\arraystretch}{1.10}
\caption{Raw paired contrasts on the 1,627-case regression evaluation. Each row comprises 1,626 known pairs; gains and losses count cases solved exclusively by the first and second listed conditions, respectively, and differences report first minus second. Cluster intervals and point estimates are computed over known pairs.}
\label{tab:sdv2-paired}
\begin{tabular*}{\linewidth}{@{\extracolsep{\fill}}lrrr@{}}
\toprule
\textbf{Comparison} & \textbf{Gains / losses} & $\bm{\Delta}$ \textbf{(pp)} & \textbf{95\% cluster CI (pp)} \\
\midrule
RL only: MTP vs STD & 74 / 68 & $+0.37$ & $[-1.16, +1.85]$ \\
Detach: MTP vs STD & 68 / 62 & $+0.37$ & $[-1.04, +1.79]$ \\
G2: MTP vs STD & 128 / 115 & $+0.80$ & $[-1.11, +2.66]$ \\
G4: MTP vs STD & 126 / 120 & $+0.37$ & $[-1.55, +2.31]$ \\
Detach: H3 vs original MTP & 69 / 61 & $+0.49$ & $[-0.86, +1.85]$ \\
\bottomrule
\end{tabular*}

\end{table}

On the complete Trunk temperature-1 bridge, standard decoding, original MTP, and H3 yield 1,122, 1,115, and 1,118 raw confirmed successes out of 1,627, respectively (with 2, 1, and 1 unknown cases). Because both the 8,192-token generation budget and sampling temperature differ jointly from the main quality protocol, this bridge evaluates combined serving-configuration sensitivity rather than isolating temperature alone.

A separate historical evaluation on the 93-problem COPT Q93 regression set yields standard/MTP success counts of 17/16, 13/15, 11/9, and 11/10 out of 93 for target-only RL, Feature Detachment, G2, and G4, respectively (all 744 scores known), under temperature $1.0$, seed 42, concurrency 8, COPT~8.0.6, and objective tolerance $10^{-4}+10^{-6}|z^\star|$ (including 14 instances with prior public exposure). This historical regression protocol differs in both answer batch and scoring setup from public1627 and from the eight-run 0731 evaluation in Section~\ref{sec:evaluation}. A subsequent native-interface sensitivity bridge generates an additional 93 responses per interface from the fixed Trunk standard target, yielding 13 confirmed successes (0 unknown) under COPT and 20 confirmed successes (15 unknown) under Gurobi; because both modeling interface and solver backend change simultaneously, this comparison reflects combined interface/backend sensitivity rather than a pure API effect.

\subsection{Frozen-Trunk Private MTP Alignment Controls}
The private-only alignment control freezes the Feature Detachment target together with its input embeddings and output head, training on 640 prefixes (491 OR and 149 document prefixes; batch size 32, 20 updates) and evaluating on a reused 256-prefix development panel (128 examples per domain, disjoint problem/document hashes from training). Three-step recursive CE reuses MTP hidden states under teacher-token inputs with position weights $(1,0.6,0.36)/1.96$. At the higher tested learning rate ($3\times10^{-6}$), OR three-token agreement rises from $62/128$ ($48.44\%$) to $75/77/79$ across three data-order seeds (mean $60.16\%$), while document agreement moves from $50/128$ ($39.06\%$) to $52/52/51$ (mean $40.36\%$).

Across the four tested configurations---initial module, single-position CE at $10^{-6}$, recursive CE at $10^{-6}$, and recursive CE at $3\times10^{-6}$---overall three-token matches out of 256 are $112$, $113$, $112/112/112$, and $127/129/130$, respectively (triplets denote data-order seeds 42--44 on the same development set; the teacher implementation matches the saved serving first token on $252/256$ prefixes). Note that teacher-token prefix agreement on this 256-prefix development panel, verifier acceptance, serving token throughput, and end-to-end OR solve accuracy measure distinct quantities. Finally, in an export screen comparing the higher-rate versus lower-rate recursive modules across three physical node pairs, median TPOT drops by $4.97\%$ and $4.41\%$ on two pairs while showing negligible median change and a $6\%$ p95 regression on the third; whereas that screen compares two adapted modules, Figure~\ref{fig:sdv2-tuning} compares the original module against H3 on the same fixed target.

\FloatBarrier
\end{document}